\documentclass[final, onefignum,onetabnum]{siamonline220329}

\usepackage{lipsum}
\usepackage{amsfonts}
\usepackage{graphicx}
\usepackage{epstopdf}
\usepackage{algorithmic}

\usepackage{microtype}
\usepackage{graphicx}
\usepackage{caption}
\usepackage{subcaption}

\usepackage{colortbl}
\definecolor{bgcolor}{rgb}{0.66,0.88,1.00}

\usepackage{makecell}
\usepackage{multirow}

\usepackage{xspace}

\xspaceaddexceptions{[]\{\}}

\usepackage{tablefootnote}
\usepackage{threeparttable}

\newcommand{\cofig}{{\sf\footnotesize COFIG}\xspace}
\newcommand{\cofigNORMAL}{{\sf COFIG}\xspace}
\newcommand{\frecon}{{\sf\footnotesize FRECON}\xspace}
\newcommand{\freconNORMAL}{{\sf FRECON}\xspace}
\newcommand{\algname}[1]{{\sf\footnotesize#1}\xspace}

\newcommand{\dataset}[1]{{\tt #1}\xspace}

\usepackage{hyperref}

\usepackage{thm-restate}

\usepackage{todonotes}

\usepackage{amsmath,amsfonts,bm}

\def\1{\bm{1}}

\def\va{{\bm{a}}}

\makeatletter
\newcommand{\ve}{\@ifnextchar\bgroup{\velong}{{\bm{e}}}}
\newcommand{\velong}[1]{{\bm{#1}}}
\makeatother

\def\vg{{\bm{g}}}
\def\vh{{\bm{h}}}

\def\vq{{\bm{q}}}

\def\vu{{\bm{u}}}
\def\vv{{\bm{v}}}
\def\vw{{\bm{w}}}
\def\vx{{\bm{x}}}
\def\vy{{\bm{y}}}

\DeclareMathAlphabet{\mathsfit}{\encodingdefault}{\sfdefault}{m}{sl}
\SetMathAlphabet{\mathsfit}{bold}{\encodingdefault}{\sfdefault}{bx}{n}

\def\gC{{\mathcal{C}}}
\def\gD{{\mathcal{D}}}

\def\gS{{\mathcal{S}}}

\newcommand{\E}{\mathbb{E}}
\newcommand{\R}{\mathbb{R}}

\newcommand{\eqdef}{\overset{\mathrm{def}}{=\joinrel=}}

\newcommand{\dotp}[2]{\left<#1, #2\right>}

\newcommand{\norm}[1]{\left\| #1 \right\|}

\def\eqref#1{(\ref{#1})}

\ifpdf
\DeclareGraphicsExtensions{.eps,.pdf,.png,.jpg}
\else
\DeclareGraphicsExtensions{.eps}
\fi

\usepackage{enumitem}
\setlist[enumerate]{leftmargin=.5in}
\setlist[itemize]{leftmargin=.5in}

\newsiamremark{remark}{Remark}
\newsiamremark{hypothesis}{Hypothesis}
\crefname{hypothesis}{Hypothesis}{Hypotheses}
\newsiamthm{claim}{Claim}
\newsiamthm{assumption}{Assumption}

\headers{Compressed FL with Client-Variance Reduction}{H. Zhao, K. Burlachenko, Z. Li, P. Richt{\'a}rik}

\title{Faster Rates for Compressed Federated Learning with Client-Variance Reduction}

\author{Haoyu Zhao\thanks{Princeton University, NJ 
		(\email{haoyu@princeton.edu}).}
	\and Konstantin Burlachenko\thanks{KAUST, Saudi Arabia 
		(\email{konstantin.burlachenko@kaust.edu.sa}, \email{peter.richtarik@kaust.edu.sa}).}
	\and Zhize Li\thanks{Carnegie Mellon University, PA
	(\email{zhizeli@cmu.edu}); Corresponding author.}
	\and Peter Richt{\'a}rik\footnotemark[2]}

\usepackage{amsopn}

\ifpdf
\hypersetup{
	pdftitle={Faster Rates for Compressed Federated Learning with Client-Variance Reduction},
	pdfauthor={H. Zhao, K. Burlachenko, Z. Li, P. Richt{\'a}rik}
}
\fi

\begin{document}

\maketitle

\begin{abstract}
Due to the communication bottleneck in distributed and federated learning applications, algorithms using communication compression have attracted significant attention and are widely used in practice. Moreover, the huge number, high heterogeneity and limited availability of clients result in high client-variance. This paper addresses these two issues together by proposing compressed and client-variance reduced methods \cofig and \frecon. We prove an $O(\frac{(1+\omega)^{3/2}\sqrt{N}}{S\epsilon^2}+\frac{(1+\omega)N^{2/3}}{S\epsilon^2})$ bound on the number of communication rounds of \cofig in the nonconvex setting, where $N$ is the total number of clients, $S$ is the number of clients participating in each round, $\epsilon$ is the convergence error, and $\omega$ is the variance parameter associated with the compression operator. In case of \frecon, we prove an $O(\frac{(1+\omega)\sqrt{N}}{S\epsilon^2})$ bound on the number of communication rounds. In the convex setting, \cofig converges within $O(\frac{(1+\omega)\sqrt{N}}{S\epsilon})$ communication rounds, which, to the best of our knowledge, is also the first convergence result for compression schemes that do not communicate with all the clients in each round. We stress that neither \cofig nor \frecon needs to communicate with all the clients, and they enjoy the first or faster convergence results for convex and nonconvex federated learning in the regimes considered. Experimental results point to an empirical superiority of \cofig and \frecon over existing baselines.
\end{abstract}



\section{Introduction}

With the proliferation of mobile and edge devices, federated learning~\cite{konevcny2016federatedlearning, mcmahan2017communication} has emerged as a framework for training machine learning models over geographically distributed and heterogeneous devices. In federated learning, there is a central server and there are plenty of clients. The central server communicates with the clients to train a machine learning model using the local data stored on the clients. Federated learning is often modeled as a distributed optimization problem~\cite{konevcny2016federatedoptimization, konevcny2016federatedlearning, mcmahan2017communication, kairouz2019advances, wang2021field}.
Let $\gD$ be the entire dataset distributed across all $N$ clients/devices/workers/machines, where each client $i$ has a local dataset $\gD_i$. The goal is to solve the following optimization problem without sharing the local datasets with the central server:
\begin{align} \label{eq:prob}
\min_{\vx\in\R^d}\left\{ f(\vx; \gD) := \frac{1}{N}\sum_{i=1}^N f(\vx; \gD_i)\right\},
\end{align}
where $f(\vx; \gD_i) := \E_{d_i\sim\gD_i} f(\vx; d_i)$. Here, $\vx \in \mathbb{R}^d$ is the model, $f(\vx; \gD)$, $f(\vx; \gD_i)$, and $f(\vx; d_{i})$ denotes the loss function of the model $\vx$ on the entire dataset $\gD$, the local dataset $\gD_i$, and a random data sample $d_{i}$, respectively.

\subsection{Communication efficiency}

Communication between the central server and the clients is a main bottleneck for the federated learning applications, especially when the number of clients is large, the clients such as mobile phones have limited bandwidth, and/or the machine learning model has a lot of parameters. Thus, it is crucial to design federated learning algorithms to reduce the overall communication cost, which takes into account both \emph{the number of communication rounds} and \emph{the communication cost per communication round} for reaching the desired accuracy. Based on these two quantities, there are two lines of approaches for communication-efficient federated learning:
1) \emph{local methods}: in each communication round, clients run multiple local update steps before communicating with the central server, hoping to reduce the number of communication rounds;
2) \emph{compressed methods}: clients send compressed communication messages to the central server, in the hope of reducing the communication cost per communication round.

Both categories have gained significant attention. For local methods, a small sample of examples include \algname{FedAvg}~\cite{mcmahan2017communication}, \algname{Local-SVRG}~\cite{gorbunov2020local}, \algname{SCAFFOLD}~\cite{karimireddy2020scaffold} and \algname{FedPAGE}~\cite{zhao2021fedpage}.
On the other hand, many compressed methods are proposed, including but not limited to \cite{alistarh2017qsgd, khirirat2018distributed, mishchenko2019distributed, horvath2019stochastic, li2020acceleration, li2020unified, gorbunov2021marina, condat2021murana, li2021canita, richtarik20223pc, li2022soteriafl, zhao2022beer}. 
In particular, the current state-of-the-art guarantees in convex and nonconvex settings are provided by \algname{CANITA}~\cite{li2021canita} and \algname{MARINA}~\cite{gorbunov2021marina}, respectively.

In this paper, we focus on compression to enhance communication efficiency. See \Cref{sec:compression} for more discussions.

\begin{table*}[!t]
	\renewcommand{\arraystretch}{1}
	\centering
	\scriptsize
	\caption{Communication rounds for different federated learning algorithms with compression in the \emph{convex} and \emph{nonconvex} settings. \emph{PP} denotes the algorithms support \emph{partial participation}, and \emph{FP} denotes the algorithms require \emph{full clients participation}. $S$ denotes the number of communicated clients in each round in the \emph{PP} situation, and $N$ denotes the total number of clients.}
	\label{tab:results}
	\vspace{2mm}
	\begin{threeparttable}
		\begin{tabular}{|c|c|ccc|}
			\hline
			 & \bf Algorithms & \bf Convex & \bf Nonconvex & \bf \makecell{Additional\\ Conditions} \\
			\hline
			\multirow{6}{*}{PP\tnote{(2)}}\ \  & \makecell{{\sf FedGLOMO} \cite{das2020improved}} & --- & $ \frac{1}{\epsilon^3} + \frac{1+\omega}{\sqrt{S}\epsilon^3}$ &
			\makecell{$f_i$ bounded below}\\
			& \makecell{{\sf PP-MARINA}\cite{gorbunov2021marina} } & --- & $\frac{1}{\epsilon^2}+\frac{(1+\omega)\sqrt{N}}{S\epsilon^2}$ & \makecell{Periodically FP using \\ uncompressed info.\tnote{(4)}} \\ 
			& \makecell{{\sf EF21-PP} \cite{fatkhullin2021ef21}} & --- & $\frac{(1+\omega)N}{S\epsilon^2}$ & ---  \\ 
			& \cellcolor{bgcolor}{\sf COFIG} (Alg~\ref{alg:cofig})\tnote{(1)} & \cellcolor{bgcolor} $\frac{L}{\epsilon} + \frac{(1+\omega)\sqrt{N} L}{S\epsilon}$ & \cellcolor{bgcolor} $\frac{1}{\epsilon^2} + \frac{(1+\omega)}{S\epsilon^2}(N^{2/3}+\sqrt{\omega N})$ & \cellcolor{bgcolor} --- \\ 
			& \cellcolor{bgcolor} {\sf FRECON} (Alg~\ref{alg:frecon}) \tnote{(1)} & \cellcolor{bgcolor} --- & \cellcolor{bgcolor} $\frac{1}{\epsilon^2} + \frac{(1+\omega)\sqrt{N}}{S\epsilon^2}$ & \cellcolor{bgcolor} --- \\
			\hline
			\hline
			\multirow{9}{*}{FP\tnote{(3)}}\ \  & \makecell{{\sf EC/EF} \cite{stich2020error,karimireddy2019error} \tnote{(1)} } & $\frac{(1+\omega)L}{\epsilon}$ & $\frac{1+\omega}{\epsilon^2}$ & Single node only \\ 
			& \makecell{{\sf DC-GD}\cite{khaled2020better} \tnote{(1)}} 
			& --- & $\frac{1}{\epsilon^2} + \frac{1+\omega}{N\epsilon^2}$
			& Bounded heterogeneity \\
			& \makecell{{\sf FedCOMGATE} \cite{haddadpour2021federated}\tnote{(1)} } & $\frac{(1+\omega)L}{\epsilon}\log\frac{1}{\epsilon}$ & $\frac{1+\omega}{\epsilon^2}$ & \makecell{Strong assumption \\ for compression\tnote{(5)}} \\ 
			& \makecell{{\sf DIANA}  \cite{mishchenko2019distributed,khaled2020unified}}\tnote{(1)} & \makecell{$\big(1+\frac{\omega}{N}\big)\frac{L}{\epsilon}+ \frac{\omega^2 + \omega}{\omega+N}\frac{1}{\epsilon}$} & $\frac{1+(1+\omega)\sqrt{\omega / N}}{\epsilon^2}$ & --- \\ 
			& \makecell{{\sf EF21} \cite{richtarik2021ef21} \tnote{(1)} } & --- & $\frac{1+\omega}{\epsilon^2}$ & --- \\ 
			& \makecell{{\sf MARINA} \cite{gorbunov2021marina} } & --- & $\frac{1+\omega/\sqrt{N}}{\epsilon^2}$ & \makecell{Periodically FP using \\ uncompressed info.} \\
			& \makecell{{\sf CANITA} \cite{li2021canita} } & \makecell{$\sqrt{\Big(1+\sqrt{\frac{\omega^3}{N}}\Big)\frac{L}{\epsilon}}+ \omega\big(\frac{1}{\epsilon}\big)^{\frac{1}{3}}$} & --- & --- \\
			& \cellcolor{bgcolor} {\sf COFIG} (Alg~\ref{alg:cofig}) \tnote{(1)} & \cellcolor{bgcolor} $\frac{L}{\epsilon} + \frac{(1+\omega)L}{\sqrt{N}\epsilon}$ & \cellcolor{bgcolor} $\frac{1}{\epsilon^2} + \frac{1+\omega}{N^{1/3}\epsilon^2} + \frac{(1+\omega)^{3/2}}{\sqrt{N}\epsilon^2}$ & \cellcolor{bgcolor} --- \\ 
			& \cellcolor{bgcolor} {\sf FRECON} (Alg~\ref{alg:frecon}) \tnote{(1)}\ \ \   & \cellcolor{bgcolor} --- & \cellcolor{bgcolor} $\frac{1}{\epsilon^2} + \frac{1+\omega}{\sqrt{N}\epsilon^2}$ & \cellcolor{bgcolor} --- \\
			\hline
		\end{tabular}
		\vspace{1.5mm}
		\begin{tablenotes}\small
			\item[(1)] Algorithms support that local clients use stochastic gradients with bounded local-variance (Assumption~\ref{ass:bounded-variance}) and do not require the local full gradients $\nabla f_i(\vx)$s.
			\item[(2)] The total communication cost in the partial participation case is computed as $ST$, where $T$ is the communication rounds (listed in the third column of Table~\ref{tab:results}) and $S$ is the number of communicated clients in each round. 
			\item[(3)] The communication cost in the full participation case is $NT$, where $T$ is the communication rounds (listed in the third column of Table~\ref{tab:results}) and $N$ is the total number of clients.
			\item[(4)] \algname{PP-MARINA} needs to periodically communicate with all clients (full clients participation) using uncompressed information.
			\item[(5)] \algname{FedCOMGATE} uses the assumption that for all vectors $\vx_1,\dots,\vx_n\in\R^n$, the compression operator $\gC$ satisfies $\E\left[\norm{\frac{1}{n}\sum_{i=1}^n \gC(\vx_i)}^2 - \norm{\gC\left(\frac{1}{n}\sum_{i=1}^n \vx_i\right)}^2\right] \le G$ for some $G \ge 0$, which does not hold for classical compression operators like RandK and $\ell_2$-quantization on $\R^d$.
		\end{tablenotes}
	\end{threeparttable}
\end{table*}

\subsection{Methods with variance reduction}
Variance reduction, or \emph{data-variance} reduction is a well-studied technique in standard/centralized finite-sum optimization, which aims to achieve a faster convergence rate. There are many optimization algorithms using the variance reduction technique, including \algname{SVRG}~\cite{johnson2013accelerating,reddi2016stochastic}, \algname{SAGA}~\cite{defazio2014saga}, \algname{SCSG}~\cite{lei2017less}, \algname{SARAH}~\cite{nguyen2017sarah}, \algname{SPIDER}~\cite{fang2018spider}, 
\algname{SSRGD}~\cite{li2019ssrgd,li2022simple},
and \algname{PAGE}~\cite{li2021page,li2021short}. 
These algorithms usually compute the stochastic gradients using a subset of data samples (obtain better convergence results) and periodically compute the full gradient using all data samples (reduce data-variance).
For example, the convergence result for standard gradient descent (\algname{GD}) (which always computes full gradients) is $O(N/\epsilon^2)$ in the nonconvex setting, while the variance-reduced methods \algname{SARAH/SPIDER/PAGE} obtain much better result $O(\sqrt{N}/\epsilon^2)$.

In the federated learning setting, we decompose the \emph{data-variance} into two parts: the \emph{client-variance} and the clients' \emph{local-variance}. The \emph{client-variance} captures the dissimilarity of data stored on different heterogeneous clients, i.e., the variance induced by different datasets $\gD_i$ (see problem \eqref{eq:prob}), which is the variance faced by the central server. The \emph{local-variance} or \emph{local-data-variance} is induced by the local data samples stored on each client, and it is the data-variance faced by the local clients if they compute local \emph{stochastic gradients} not local \emph{full gradients}. 
We point out that the \emph{client-variance} is the main issue for federated learning because of the data heterogeneity issue, and even if one can compute the local \emph{full gradients} using the whole local dataset $\gD_i$ on each client (i.e., remove the \emph{local-variance}), the \emph{client-variance} faced by the central server still exists. 

Moreover, directly applying the variance reduction algorithms like \algname{SVRG/PAGE} in the federated learning has the following drawbacks: (1) These variance-reduced methods usually need to periodically compute the full gradient using all data samples (reduce data-variance), which leads to communication with all clients periodically in the federated learning setting. However, federated learning applications typically involve a lot of clients, for example, Google Keyboard (Gboard)~\cite{hard2018federated,ramaswamy2019federated} involves billions of mobile phones, and thus it is impractical to communicate with all the clients in a communication round to reduce the \emph{client-variance}. (2) These variance-reduced algorithms do not use communication compression to further reduce the communication cost.
On the other hand, the existing communication compression algorithms usually need to communicate with all the clients in each communication round (see Tables~\ref{tab:results}). Among the compressed algorithms that support partial participation (PP) (do not need to communicate with all clients in each round), previous methods either need to communicate with all the clients (FP) using uncompressed messages periodically (\algname{PP-MARINA}~\cite{gorbunov2021marina}), or do not enjoy the faster convergence rate of variance reduction (\algname{FedGLOMO}~\cite{das2020improved}, \algname{EF21-PP}~\cite{fatkhullin2021ef21}). 
Thus, we consider the following natural and important question:

\begin{quote}
	\centering
    \emph{Can we incorporate client-variance reduction with communication compression\\ in the federated learning setting?}
\end{quote}

\subsection{Our contributions}

In this paper, we design \cofig and \frecon algorithms for federated learning. Both algorithms support communication compression and fully partial participation, and achieve the first/faster convergence using client-variance reduction technique (see \Cref{tab:results}) under \emph{any heterogeneity}. 
Concretely, we summarize our main contributions as follows:
\begin{enumerate}
    \item We design \cofig algorithm, which to our best knowledge, 
    is the \emph{first} algorithm that supports \emph{partial participation} and communication compression for the \emph{general} non-strongly convex federated learning setting (see the convex column in \Cref{tab:results}). 
    For the nonconvex setting, \cofig successfully supports communication compression and fully partial participation (PP) (unlike \algname{PP-MARINA}) and enjoys the convergence speedup (unlike \algname{FedGLOMO} and \algname{EF21-PP}) by using client-variance reduction (see the nonconvex column in \Cref{tab:results}).
    Besides, \cofig also supports the client's local-variance (\Cref{ass:bounded-variance}) and PP simultaneously, which is more practical than other compressed algorithms such as \algname{FedGLOMO} and \algname{PP-MARINA} in federated learning.
    \item For the nonconvex setting, we further design a new \frecon algorithm based on \cofig, which can converge much faster than \cofig (see the nonconvex column in \Cref{tab:results}). 
    In particular, \frecon converges within $O((1+\omega)\sqrt{N}/(S\epsilon^2))$ communication rounds, matches the convergence result of \algname{PP-MARINA}. However, \frecon supports fully partial participation with a subset clients and thus is more practical than \algname{PP-MARINA} which requires periodical full participation (FP) with uncompressed messages.
    Note that periodic FP needs to periodically synchronize all clients/devices in federated learning, which is usually impossible or unaffordable.
\end{enumerate}

\section{Preliminaries}

Let $[n]$ denote the set $\{1,2,\cdots,n\}$ and $\norm{\cdot}$ denote the Euclidean norm for a vector.
Let $\dotp{\vu}{\vv}$ denote the standard Euclidean inner product of two vectors $\vu$ and $\vv$. In addition, we use the standard order notation $O(\cdot)$ to hide absolute constants.
For simplicity, we use $f(\vx)$ and $f_i(\vx)$ to denote $f(\vx; D)$ and $f(\vx; D_i)$ respectively. Then, the federated optimization problem \eqref{eq:prob} becomes

\begin{equation*}
\min_{\vx\in\R^d}\left\{ f(\vx) := \frac{1}{N}\sum_{i=1}^N f_i(\vx)\right\},
\end{equation*}

where $f_i(x):=f(\vx; \gD_i) := \E_{d_i\sim\gD_i} f(\vx; d_i)$. We assume that we have access to the local stochastic gradient oracle $\tilde\nabla f_i(\vx) := \nabla f_i(\vx; d_i)$ where the sample $d_i$ is randomly drawn from the local dataset $\gD_i$. We define $\vx^*$ to be the solution of \eqref{eq:prob}, and $f^* := f(\vx^*)$. For simplicity, for a nonconvex function $f$, we assume that $f^* > -\infty$, and for a convex function $f$, we assume $\norm{\vx^*} < \infty$.

\subsection{Assumptions about the functions}

In this paper, we consider two general settings: the nonconvex setting and the convex setting. 

In the nonconvex setting, the functions $\{f_{i}\}_{i\in [N]}$ are arbitrary functions that satisfy the following smoothness assumption~\cite{johnson2013accelerating,defazio2014saga,nguyen2017sarah,zhou2018stochastic,fang2018spider,li2021page}, and we assume that the unbiased local stochastic gradient oracle $\tilde\nabla f_i(\vx)$ has bounded local variance~\cite{mcmahan2017communication,karimireddy2020scaffold,zhao2021fedpage}, which is also standard in the federated learning literature. We use $\tilde\nabla_b f_i(\vx)$ to denote the stochastic gradient oracle that uses a minibatch size $b$, which is the average of $b$ independent unbiased stochastic gradients $\tilde\nabla f_i(\vx)$.

\begin{assumption}[Smoothness]\label{ass:smooth}
	The functions $\{f_i\}_{i\in [N]}$ are $L$-smooth if there exists $L \ge 0$ such that for any client $i\in [N]$,
	$\norm{ \nabla f_i(\vx_1) - \nabla f_i(\vx_2)} \le L\norm{\vx_1 - \vx_2}$, for all $\vx_1, \vx_2\in \R^d$.
\end{assumption}

\begin{assumption}[Bounded local-variance]\label{ass:bounded-variance}
	There exists a constant $\sigma \ge 0$ such that for any client $i\in [N]$ and any $\vx\in\R^d$,
	$\E\norm{\tilde\nabla f_i(\vx) - \nabla f_i(\vx)}^2 \le \sigma^2$.
	Then for a stochastic gradient oracle with minibatch size $b$, we have
	$\E\norm{\tilde\nabla_b f_i(\vx) - \nabla f_i(\vx)}^2 \le \frac{\sigma^2}{b}$.
\end{assumption}

For the convex setting, besides the smoothness and bounded local-variance assumptions, we also need the function $f$ to be convex. Formally, we use the following assumption.

\begin{assumption}[Convexity]\label{ass:convex}
	A function $f$ is convex if $\forall\vx,\vy\in\R^d,f(\vy) \ge f(\vx) + \dotp{\nabla f(\vx)}{\vy-\vx}$.
\end{assumption}

\subsection{Compression operators}
\label{sec:compression}
We now introduce the notion of a randomized {\em compression operator} which we use to reduce the total communication bits. 
We adopt the standard class of unbiased compressors (see \Cref{def:comp}) that was widely used in previous distributed/federated learning literature \cite{alistarh2017qsgd, khirirat2018distributed, mishchenko2019distributed, horvath2019stochastic, li2020acceleration, gorbunov2021marina, li2021canita}. 

\begin{definition}[Compression operator]\label{def:comp}
	A randomized map $\gC: \R^d\mapsto \R^d$ is an $\omega$-compression operator  if for all $\vx\in \R^d$, it satisfies
	\begin{equation}\label{eq:comp}
	\E[\gC(\vx)]=\vx, \qquad \E\left[\norm{\gC(\vx)-\vx}^2\right]\leq \omega\norm{\vx}^2.
	\end{equation}
	In particular, no compression ($\gC(\vx)\equiv \vx$) implies $\omega=0$.
\end{definition}

\section{\cofigNORMAL: Fast Compressed Algorithm with Variance Reduction}\label{sec:cofig}

In this section, we present our \cofig algorithm (\Cref{alg:cofig}), which supports gradient compression, fully partial participation, and has better convergence results. In \Cref{sec:cofig-alg}, we introduce our \cofig algorithm, and in \Cref{sec:cofig-converge}, we present the convergence rate of \cofig in both convex and nonconvex settings.

\subsection{\cofigNORMAL algorithm}\label{sec:cofig-alg}

\begin{algorithm}[!t]
	\caption{COmpressed Fast Incremental Gradient Algorithm (\cofigNORMAL)} \label{alg:cofig}
	\begin{algorithmic}[1]
		\REQUIRE initial point $\vx^0$, $\vw^0_i = \vx^0$ for all $i\in [N]$, $\vh^0_i$ for all $i\in [N]$, $\vh^0 = \frac{1}{N}\sum_{i=1}^N \vh^0_i$, step size $\eta$, shift step size $\alpha$, minibatch size $b$
		\FOR{$t=0,1,2,\dots$}
		\STATE Randomly sample clients $\gS^t, \tilde\gS^t$ with size $S$, and server broadcast $\vx^{t}$ to clients $\gS^t, \tilde\gS^t$
		\STATE \textbf{for all clients} $i \in \gS^t$
		\STATE \hspace{\algorithmicindent} compute $\vu_i^{t} = \gC(\tilde\nabla_b f_{i} (\vx^{t}) - \vh^t_{i})$ {and send $\vu^t_i$ to the server} \label{line:cofig-active-update-mess}
		\STATE \hspace{\algorithmicindent} $\vh^{t+1}_{i} = \vh^t_{i} + \alpha \vu^t_i$, $\vw_i^{t+1} = \vx^t$ \label{line:cofig-active-update}
		\STATE \textbf{end all clients} $i \in \gS^t$
		\STATE \textbf{for all clients} $j\notin \gS^t$
		\STATE \hspace{\algorithmicindent} $\vh^{t+1}_j = \vh^t_j$, $\vw_i^{t+1} = \vw_i^t$ \label{line:cofig-inactive}
		\STATE \textbf{end all clients} $j\notin \gS^t$
		\STATE \textbf{for all clients} $i \in \tilde\gS^t$
		\STATE \hspace{\algorithmicindent} compute $\vv_i^{t} = \gC(\tilde\nabla_b f_{i} (\vx^{t}) - \vh^t_{i})$ {and send $\vv^t_i$ to the server}
		\STATE \textbf{end all clients} $i \in \tilde\gS^t$
		\STATE {// Server update:}
		\STATE $\vg^{t} = \frac{1}{S}\sum_{i\in \tilde\gS^{t}}\vv^{t}_i + \vh^t$ \label{line:cofig-unbiased}
		\STATE $\vx^{t+1} = \vx^t - \eta \vg^t$
		\STATE $\vh^{t+1} = \vh^t + \frac{\alpha}{N}\sum_{i\in\gS^t} \vu^t_i$ \label{line:cofig-global-shift}
		\ENDFOR
	\end{algorithmic}
\end{algorithm}

In our \cofig algorithm, each client $i$ maintains the local shift sequence $\{\vh_i^t\}$ and a virtual sequence $\{\vw_i^t\}$\footnote{Here, the virtual sequence $\{\vw_i^t\}$ is only used for algorithmic illustration and analysis purpose. In real execution, the clients do not need to maintain $\{\vw_i^t\}$.}, and the server maintains the global shift $\{\vh^t\}$ and the model $\vx^t$ in each round. At a very high-level point of view, the virtual variable $\vw_i^t$ records the model $\vx$ in client $i$'s last active communication round, and the shifts $\{\vh_i^t\}$ are used for reducing the variance.
 
Now, we describe these ingredients in more detail.
\cofig can be divided into three steps: (1) clients compute and compress the local gradient information using the current model $\vx^t$ and local shifts $\{\vh_i^t\}$; (2) clients update the local shift $\{\vh_i^{t+1}\}$ and the virtual sequence $\{\vw_i^{t+1}\}$; and (3) the server updates the model to $\vx^{t+1}$ and the global shift $\{\vh^{t+1}\}$ using the compressed gradient information aggregated from the clients.

In each round $t$, the server first sample two sets of clients $\gS^t$ and $\tilde\gS^t$. For clients $i\in\gS^t$, they updates their control sequence $\{\vh_i^t\}_{i\in [N]}$ and the virtual sequence $\{\vw_i^t\}_{i\in [N]}$, and send the compressed message to the server (Line \ref{line:cofig-active-update-mess}, \ref{line:cofig-active-update} of \Cref{alg:cofig}). Note that when we update the local shift, there is a shift step size $\alpha$, which makes the shift $\vh_i^{t+1}$ a biased estimator of $\nabla f_i(\vw_i^{t+1})$, since
\begin{align*}
    \E[\vh_i^{t+1}] = & \E[\vh_i^t + \alpha\gC(\tilde\nabla_b f_i(\vx^t) - \vh_i^t)] = (1-\alpha)\vh_i^t + \alpha \nabla f_i(\vx^t).
\end{align*}
This biased shift is useful for controlling the variance.

Then for the clients $j\in\tilde\gS^t$, they compute the estimators and send to the server. Note the compressed message $\vv_j^t$ computed by clients $j\in\tilde\gS^t$ is nearly the same as the compressed message $\vu_i^t$ computed by clients $i\in \gS^t$ except the randomness of compression operator, and we need to compute $\vv_j^t$ to decouple the randomness in the nonconvex setting.

Finally, the server updates the model $\vx^t$ and the global shift $\vh^t$. From Line \ref{line:cofig-global-shift} of Algorithm \Cref{alg:cofig}, it is easy to prove that he global shift $\vh^t$ is an average of the local shifts $\{\vh_i^t\}$. To update the model, the server first computes a gradient estimator $\vg^t$ using the compressed information aggregated from the clients. The estimator $\vg^t$ is an unbiased estimator for $\nabla f(\vx^t)$, since
\begin{align*}
    \E [\vg^t] = & \E \left[\frac{1}{S}\sum_{i\in \tilde\gS^{t}}\gC(\tilde\nabla_b f_{i} (\vx^{t}) - \vh^t_{i}) + \vh^t\right]
    =  \E \left[\frac{1}{S}\sum_{i\in \tilde\gS^{t}}(\nabla f_{i} (\vx^{t}) - \vh^t_{i}) + \vh^t\right] = \nabla f(\vx^t).
\end{align*}

Note that in real execution, each client only needs to maintain vector $\vh_i$ and the server only need to maintain $\vh = \frac{1}{N}\sum_{i=1}^N \vh_i$. Thus, Line \ref{line:cofig-inactive} does not need to be executed by clients $j\notin\gS^t$. Besides, the clients do not need to maintain the sequence $\vw_i$. 
We explicitly write out the superscripts and the sequence $\vw$ for better illustration and analysis.

Now we point out \cofig in the degenerated cases: 1) the case with no compression ($\omega=0$) and no clients' local-variance ($\tilde\nabla_b f_i(\vx^t) = \nabla f_i(\vx^t)$), 
our \cofig algorithm reduces to the \algname{SAGA} algorithm~\cite{defazio2014saga} for finite-sum optimization;
2) the case with full participation ($\gS^t = \tilde\gS^t = [N]$), our algorithm reduces to the \algname{DIANA} algorithm~\cite{mishchenko2019distributed,horvath2019stochastic}. 
As a result, our \cofig enhances the \algname{SAGA} with communication compression, and generalizes the \algname{DIANA} without requiring full clients participation.

\subsection{Convergence results for \cofigNORMAL}\label{sec:cofig-converge}

Now we state the convergence results of \cofig in convex setting (\Cref{thm:cofig-convex}) and nonconvex setting (\Cref{thm:cofig-nonconvex}). These two theorems show that, by incorporating client-variance reduction with communication compression, \cofig obtains faster convergence results (enjoying the speedup from variance reduction techniques) and supports partial participation (thus more practical for federated learning).

\begin{theorem}[\cofig in convex case]\label{thm:cofig-convex}
	Suppose that \Cref{ass:smooth}, \Cref{ass:bounded-variance}, and \Cref{ass:convex} hold. By choosing $\alpha = \frac{1}{1+\omega}$, and
    $\eta = \min\Big\{\frac{1}{L(2 + 8(1+\omega)/S)}, \frac{S}{(1+\omega)\sqrt{N}}, \frac{bS\epsilon}{2(1+\omega)\sigma^2}\Big\}$,
    \cofig~(\Cref{alg:cofig}) will find a point $x$ such that $\E f(\vx) - f^* \le \epsilon$ in the following number of communication rounds
    \[T = O\left(\frac{L}{\epsilon} + \frac{(1+\omega)\sqrt{N}L}{S\epsilon} + \frac{(1+\omega)\sigma^2}{bS\epsilon^2}\right).\]
\end{theorem}

\begin{theorem}[\cofig in nonconvex case]\label{thm:cofig-nonconvex}
    Suppose that \Cref{ass:smooth} and \Cref{ass:bounded-variance} hold. By choosing $\alpha = \frac{1}{1+\omega}$, and
    $\eta = \min\Big\{\frac{1}{2L}, \frac{S}{5L(1+\omega)N^{2/3}}, \frac{S}{5L(1+\omega)^{3/2}\sqrt{N}},\frac{bS\epsilon^2}{20(1+\omega)\sigma^2}\Big\}$,
    \cofig~(\Cref{alg:cofig}) will find a point $x$ such that $\E\norm{\nabla f(\vx)} \le \epsilon$ in the following number of communication rounds
    \[O\left(\frac{L}{\epsilon^2} + \frac{L(1+\omega)N^{2/3}}{S\epsilon^2} + \frac{L(1+\omega)^{3/2}\sqrt{N}}{S\epsilon^2} + \frac{(1+\omega)\sigma^2}{bS\epsilon^4}\right).\]
\end{theorem}

Now we interpret the convergence results in \Cref{thm:cofig-convex} and \Cref{thm:cofig-nonconvex}. First in the convex setting (\Cref{thm:cofig-convex}), \cofig converges to an $\epsilon$-solution with the following total communication cost $O(\frac{S}{\epsilon}+\frac{(1+\omega)\sqrt{N}}{\epsilon} + \frac{(1+\omega)\sigma^2}{b\epsilon^2})$,
by multiplying the communication rounds $T$ by $S$. 
The first two terms in the communication rounds $T$ come from the client-variance reduction, and the last term comes from the clients' local variance. 
Note that our \cofig algorithm provides the \emph{first} convergence result which supports partial participation and communication compression in the convex setting (see \Cref{tab:results}). 
Moreover, our algorithm support clients' local-variance (\Cref{ass:bounded-variance}), which has more advantages compared with other compressed gradient methods listed in \Cref{tab:results} and \Cref{tab:results}.

Next in the nonconvex case (\Cref{thm:cofig-nonconvex}), \cofig converges to an $\epsilon$-approximate stationary point with the following total communication cost $O(\frac{S}{\epsilon^2}+\frac{(1+\omega)N^{2/3}}{\epsilon^2}+\frac{(1+\omega)^{3/2}\sqrt{N}}{\epsilon^2}+\frac{(1+\omega)\sigma^2}{b\epsilon^4})$. Similar to the convex case, the first two terms come from the client-variance reduction, and the last term comes from the clients' local variance. The third term in the convergence rate comes from the analysis of the compression methods, and the similar term also shows up in the convergence rate of \algname{DIANA}, which is the full participation version of \cofig.
When there is no local variance ($\sigma^2 = 0$), note that except \algname{FedGLOMO} and \algname{PP-MARINA}, all other methods in \Cref{tab:results} need $O(N/\epsilon^2)$ total communication cost, and our \cofig converges in $O(N^{2/3}/\epsilon^2)$. Although \algname{FedGLOMO} converges in $\sqrt{S}$ and \algname{PP-MARINA} converges in $O(\sqrt{N})$, \algname{FedGLOMO} has a worse dependency on $\epsilon$ and needs to assume that the individual functions $f_i$ are lower bounded, and \algname{PP-MARINA} needs to periodically communicate with all the clients with uncompressed message. Moreover, both \algname{FedGLOMO} and \algname{PP-MARINA} do not support clients' local-variance, which makes \cofig more flexible/practical for the federated learning setting. In the next section, we introduce another algorithm \frecon (\Cref{alg:frecon}) that matches the convergence rate of \algname{PP-MARINA} in the nonconvex setting, but it does not need to periodically communicate with all of the clients using uncompressed messages unlike \algname{PP-MARINA}.

Note that by choosing a proper minibatch size $b$, we further improve the convergence rate in terms of the communication rounds when $\sigma$ is large. Formally, we have the following corollary.

\begin{corollary}\label{cor:cofig}
    Suppose that \Cref{ass:smooth} and \Cref{ass:bounded-variance} hold. If we choose $\eta$ properly and $b \ge \frac{\sigma^2}{\epsilon^2 N^{2/3}}$,
    \cofig~(\Cref{alg:cofig}) will find a point $x$ such that $\E\norm{\nabla f(\vx)} \le \epsilon$ in the following number of communication rounds
    $T = O\left(\frac{L}{\epsilon^2} + \frac{L(1+\omega)N^{2/3}}{S\epsilon^2} + \frac{L(1+\omega)^{3/2}\sqrt{N}}{S\epsilon^2}\right)$.
    If \Cref{ass:convex} also holds, i.e. $f$ is convex, then if we choose $\eta$ properly and $b \ge \frac{\sigma^2}{\epsilon^2 \sqrt{N}}$,
    \cofig~(\Cref{alg:cofig}) will find a point $x$ such that $\E f(\vx) - f^* \le \epsilon$ in the following number of communication rounds
    $T = O\left(\frac{L}{\epsilon} + \frac{(1+\omega)\sqrt{N}L}{S\epsilon}\right)$.
\end{corollary}

\subsection{Proof sketch and technical novelty}\label{sec:cofig-novelty}
In this part, we highlight our proof sketch and technical novelty, and we focus on the nonconvex setting. Although our \cofig algorithm can be seen as a generalization of both \algname{SAGA} and \algname{DIANA} and our convergence results recover the results of both algorithms, the proof of \cofig is totally different and requires new techniques. Previous compressed algorithms, e.g. \algname{EF21-PP}, \algname{EF21}, \algname{DIANA}, directly bound the following term $\E\sum_i \norm{\vh_i^t - \nabla f_i(\vx^t)}^2$, which describes the difference between the control shift $\vh_i^t$ and the local full gradient $\nabla f_i(\vx^t)$ and usually has the following form
\begin{align}
    & \E\sum_i \norm{\vh_i^{t+1} - \nabla f_i(\vx^{t+1})}^2 \nonumber \le (1-\beta)\E\sum_i \norm{\vh_i^t - \nabla f_i(\vx^t)}^2 + A,\label{eq:cofig-simple-bound}
\end{align}
where $\beta>0$ is a constant related to $S, N, \omega$ and $A$ denotes another term.
However, if we directly bound the term $\E\sum_i \norm{\vh_i^t - \nabla f_i(\vx^t)}^2$, we are not able to get the faster convergence rate $N^{2/3}/(S\epsilon^2)$ in terms of the communication rounds and we can only get $N/(S\epsilon^2)$, which is exactly the convergence rate of \algname{EF21-PP}. To get the better convergence rate, we need to introduce the virtual sequence $\{\vw_i^t\}$ and split $\E\sum_i \norm{\vh_i^{t+1} - \nabla f_i(\vx^{t+1})}^2$ into $\E\sum_i \norm{\vh_i^{t+1} - \nabla f_i(\vw^{t+1})}^2$ and $\E\sum_i \norm{\nabla f_i(\vw^{t+1}) - \nabla f_i(\vx^{t+1})}^2$, and using tighter inequalities to bound these two terms. More specifically, we first apply Young's inequality to get bounds like
\begin{align}
    & \sum_i \norm{\vh_i^{t+1} - \nabla f_i(\vx^{t+1})}^2 \nonumber \\
    & \quad \le (1+\beta_1)\sum_i \norm{\vh_i^{t+1} - \nabla f_i(\vw^{t+1})}^2 + (1+\beta_1^{-1})\sum_i \norm{\nabla f_i(\vw^{t+1}) - \nabla f_i(\vx^{t+1})}^2 \label{eq:cofig-split-term}
\end{align}
for all $t$, where $\beta_1>0$ is a constant. Then, we bound the terms separately and get the bounds like
\begin{align}
    & \E\sum_i \norm{\vh_i^{t+1} - \nabla f_i(\vw^{t+1})}^2 \le (1-\beta_2)\E\sum_i \norm{\vh_i^t - \nabla f_i(\vw^t)}^2 +A_2, \label{eq:cofig-term-1} \\
    & \E\sum_i \norm{\nabla f_i(\vw^{t+1}) - \nabla f_i(\vx^{t+1})}^2 \le (1-\beta_3)\E\sum_i \norm{\nabla f_i(\vw^t) - \nabla f_i(\vx^t)}^2 +A_3, \label{eq:cofig-term-2}
\end{align}
where $\beta_2,\beta_3>0$ are constants and $A_2,A_3$ are other terms. Combining (\ref{eq:cofig-split-term}), (\ref{eq:cofig-term-1}) and (\ref{eq:cofig-term-2}) together can also provide bounds for the term $\E\sum_i \norm{\vh_i^{t+1} - \nabla f_i(\vx^{t+1})}^2$, but the resulting inequality is tighter than (\ref{eq:cofig-simple-bound}), and can lead to better $N^{2/3}/(S\epsilon^2)$ convergence rate.

\section{\freconNORMAL: Faster Algorithm in the Nonconvex Setting}\label{sec:frecon}

In this section, we present another algorithm \frecon described in \Cref{alg:frecon}, which can further achieve a faster convergence rate than \cofig in the nonconvex setting when the clients' local-variance is small. We present our \frecon algorithm in \Cref{sec:frecon-alg} and show the convergence results in \Cref{sec:frecon-converge}.

\subsection{\freconNORMAL algorithm}\label{sec:frecon-alg}

\begin{algorithm}[!t]
	\caption{Faster Recursive gradiEnt COmpression for Nonconvex Federated Learning (\freconNORMAL)} \label{alg:frecon}
	\begin{algorithmic}[1]
		\REQUIRE initial point $\vx^0$, $\vh_i^{0}$, $\vh^{0} = \frac{1}{N}\sum_{i=1}^N \vh^{0}_i$, $\vg^0 = 0$ step size $\eta$, shift step size $\alpha$, parameter $\lambda$, minibatch size $b$
		\FOR{$t=0,1,2,\dots$}
		\STATE $\vx^{t+1} = \vx^t - \eta \vg^t$
		\STATE Randomly sample clients $\gS^t$ with size $S$, and server broadcast $\vx^{t+1}$, $\vx^{t}$ to clients $\gS^t$
		\STATE \textbf{for all clients} $i\in \gS^t$
		\STATE \hspace{\algorithmicindent} compute $\vq^t_i = \gC(\tilde\nabla_b f_i(\vx^{t+1}) - \tilde\nabla_b f_i(\vx^{t}))$ and send $\vq^t_i$ to the server \label{line:frecon-compress-diff}
		\STATE \hspace{\algorithmicindent} compute $\vu^t_i = \gC(\tilde\nabla_b f_i(\vx^{t}) - \vh^{t}_i)$ and send $\vu^t_i$ to the server \label{line:frecon-compress-shift}
		\STATE \hspace{\algorithmicindent} $\vh^{t+1}_i = \vh^t_i + \alpha \vu^t_i$
		\STATE \textbf{end all clients} $i\in \gS^t$
		\STATE \textbf{for all clients} $j\notin \gS^t$
		\STATE \hspace{\algorithmicindent} $\vh^{t+1}_i = \vh^t_i$
		\STATE \textbf{end all clients} $j\notin \gS^t$
		\STATE $\vg^{t+1} = \frac{1}{S}\sum_{i\in\gS^t}\vq^t_i + (1-\lambda)\vg^t + \lambda\left(\frac{1}{S}\sum_{i\in \gS^t}\vu^t_i + \vh^t\right)$ \label{line:frecon-estimator}
		\STATE $\vh^{t+1} = \vh^t + \frac{\alpha}{N}\sum_{i\in\gS^t}\vu^t_i$
		\ENDFOR
	\end{algorithmic}
\end{algorithm}

To achieve $O(\sqrt{N}/\epsilon^2)$ total communication cost in the nonconvex setting, most algorithms use the recursive gradient estimator. Among these algorithms, most of them need to periodically compute full gradients to reduce the variance of the gradient estimator, e.g. \algname{SARAH}~\cite{nguyen2017sarah}, \algname{PAGE}~\cite{li2021page}, \algname{PP-MARINA}~\cite{gorbunov2021marina}, \algname{FedPAGE}~\cite{zhao2021fedpage}, except \algname{ZeroSARAH}\cite{li2021zerosarah} which never needs to compute the full gradient. The idea of \algname{ZeroSARAH} is to maintain another unbiased approximate gradient estimator for the full gradient, and update both the biased recursive and unbiased approximate estimators in each round.

In the federated learning setting with communication compression, recall that in the previous section (\Cref{sec:cofig}), our \cofig (\Cref{alg:cofig}) can also generate approximate full gradient estimator $\vg^t$ (Line \ref{line:cofig-unbiased} of \Cref{alg:cofig}) without communicating with all the clients, the idea of \frecon (Algorithm~\Cref{alg:frecon}) is to incorporate recursive gradient estimator together with \cofig (see Line~\ref{line:frecon-estimator} of \Cref{alg:frecon}).

In each round, the server sample a set of clients $\gS^t$ to communicate with, and the clients $i\in \gS^t$ compute two compressed messages, the first $\vq_i^t$ is the compressed message for the recursive gradient estimator (Line \ref{line:frecon-compress-diff} of \Cref{alg:frecon}), and the second $\vu_i^t$ is the compressed message to update the local shift $\vh_i$ and the unbiased full gradient estimator (Line~\ref{line:frecon-compress-shift} of \Cref{alg:frecon}). Then, the server computes the gradient estimator (Line~\ref{line:frecon-estimator} of \Cref{alg:frecon}) by
\[\vg^{t+1} = \frac{1}{S}\sum_{i\in\gS^t}\vq^t_i + (1-\lambda)\vg^t + \lambda\left(\frac{1}{S}\sum_{i\in\gS^t}\vu^t_i + \vh^t\right).\]
Here, the term $\frac{1}{S}\sum_{i\in\gS^t}\vu^t_i + \vh^t$ is exactly the gradient estimator used in \cofig, and the gradient estimator $\vg^t$ is a biased estimator. 
Besides updating the gradient estimator, the server and the clients also need to maintain the global and local shifts $\vh^t$ and $\{\vh_i^t\}$ similar to \cofig. 
Similarly, \frecon with no compression ($\omega=0$) and no clients' local-variance reduces to the \algname{ZeroSARAH} algorithm~\cite{li2021zerosarah}.

\subsection{Convergence results for \freconNORMAL}\label{sec:frecon-converge}

Now we state the convergence result for \frecon in the nonconvex setting.

\begin{theorem}[\frecon in nonconvex case]\label{thm:frecon-nonconvex}
Suppose that \Cref{ass:smooth} and \Cref{ass:bounded-variance} hold. By choosing $\alpha = \frac{1}{1+\omega}$, $\eta \le \frac{1}{L(1+\sqrt{10 (1+\omega)^2 N/S^2})}$, and the minibatch size $b \ge O(\frac{(1+\omega)^2 N\sigma^2}{S^2\epsilon^2})$, \frecon (\Cref{alg:frecon}) will find a point $x$ such that $\E\norm{\nabla f(\vx)} \le \epsilon$ in the following number of communication rounds
\[T = O\left(\frac{L}{\epsilon^2} + \frac{(1+\omega)\sqrt{N}}{S\epsilon^2}\right).\]
\end{theorem}

From \Cref{thm:frecon-nonconvex}, \frecon enjoys the faster convergence result by the client-variance reduction.
The total communication cost of \frecon is $O(\frac{S}{\epsilon^2}+\frac{(1+\omega)\sqrt{N}}{\epsilon^2})$, which matches the current state-of-the-art result of \algname{PP-MARINA} in the nonconvex setting.  
However, \algname{PP-MARINA} needs to periodically communicate with all of the clients (FP) with exact information (the uncompressed information), which is impractical in the federated learning setting, while our \frecon only needs to communicate with a subset of clients in every communication round (PP) using compressed messages.

We note that \frecon is not strictly better than \cofig.
If there exists clients' local-variance, the total communication cost of \frecon is $O(\frac{S}{\epsilon^2}+\frac{(1+\omega)\sqrt{N}}{\epsilon^2})$, while \cofig is $O(\frac{S}{\epsilon^2}+\frac{(1+\omega)N^{2/3}}{\epsilon^2}+\frac{(1+\omega)^{3/2}\sqrt{N}}{\epsilon^2}+\frac{(1+\omega)\sigma^2}{b\epsilon^4})$. 
However,\frecon also needs to sample minibatch stochastic gradients with minibatch size $b \ge O(\frac{(1+\omega)^2 N\sigma^2}{S^2\epsilon^2})$ for the communicated clients in each round, while  \cofig can adopt minibatch stochastic gradients with size $b=1$ (but also can use minibatch $b> 1$ to decrease the total communication, see \Cref{cor:cofig}).
As a result, the stochastic gradient complexity for \frecon is $O(\frac{(1+\omega)\sqrt{N}}{\epsilon^2} + \frac{(1+\omega)\sqrt{N}\sigma^2}{\epsilon^4})$ by optimizing $S$ to be $(1+\omega)\sqrt{N}$, and for \cofig is $O(\frac{(1+\omega)N^{2/3}}{\epsilon^2}+\frac{(1+\omega)^{3/2}\sqrt{N}}{\epsilon^2}+\frac{(1+\omega)\sigma^2}{\epsilon^4})$ by choosing $S \le (1+\omega)N^{2/3}$.

\section{Numerical Experiments}

In this section, we show our numerical experiment results.

\subsection{General setup}
We run experiments on two problems: logistic regression with a nonconvex regularizer~\cite{wang2018spiderboost, zhao2021fedpage} using the \dataset{a9a} dataset~\cite{chang2011libsvm} and training ResNet-18 neural network~\cite{he2016deep} using the \dataset{CIFAR10} dataset \cite{krizhevsky2009learning}.

For logistic regression with a nonconvex regularizer problem, following~\cite{wang2018spiderboost, zhao2021fedpage}, the objective function for a sample $(\va,b)$ is defined as
\[f(\vx; (\va,b)) =   \log\left(1+\exp(-b \va^\top\vx )\right) + \alpha\sum \limits_{j=1}^d\frac{x_j^2}{1+x_j^2},\]
where the last term is the nonconvex regularizer and we choose $\alpha=0$ in the convex experiments and $\alpha=0.1$ in the nonconvex experiments.

In the convex experiments (logistic regression), we compare \cofig with \algname{DIANA} and \algname{EF21-PP}. We include \algname{EF21-PP} since we want to compare with some algorithms with partial participation, although it does not have convex analysis. In the logistic regression with nonconvex regularizer, we compare \cofig, \frecon with \algname{DIANA}, \algname{EF21-PP}, and \algname{PP-MARINA}. In the previous experiments, we compute the local full gradient ($\sigma = 0$), use the theoretical learning rate, and use \texttt{Natural} compression operator~\cite{horvath2019natural} for different algorithms. Besides, we also sort the dataset to simulate the data heterogeneous setting. In the deep learning experiments, we run \algname{EF21} instead of \algname{DIANA},\footnote{We can get the hyperparameters for \algname{EF21}, e.g., the learning rates, from the original paper.} and we estimate the local gradient using a minibatch ($\sigma > 0$). Besides, we tune the learning rate to get the best convergence performance and use random sparsification compressor~\cite{wangni2018gradient} for all algorithms in the deep learning experiment.

\subsection{Binary classification with convex/nonconvex loss}
In this experiment, we compare different algorithms under the theoretical step size. We consider both the convex and nonconvex problems.

\paragraph{Problem formulation} For logistic regression with a nonconvex regularizer problem, following~\cite{wang2018spiderboost, zhao2021fedpage}, the objective function for a sample $(\va,b)$ is defined as
\[f(\vx; (\va,b)) =   \log\left(1+\exp(-b \va^\top\vx )\right) + \alpha\sum \limits_{j=1}^d\frac{x_j^2}{1+x_j^2},\]
where the last term is the nonconvex regularizer. We choose $\alpha=0$ in the convex experiments and the problem becomes the classical logistic regression, and $\alpha=0.1$ in the nonconvex experiments.

\begin{figure*}[!ht]
	\centering
	\captionsetup[sub]{font=scriptsize,labelfont={}}	
	
	\begin{subfigure}[ht]{0.75\textwidth}
		\includegraphics[width=\textwidth]{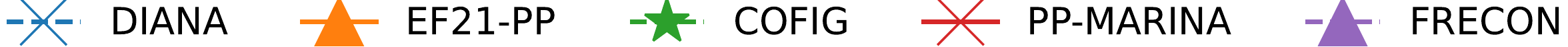}  \caption*{}
	\end{subfigure}
	
	\begin{subfigure}[ht]{0.32\textwidth}
		\includegraphics[width=\textwidth]{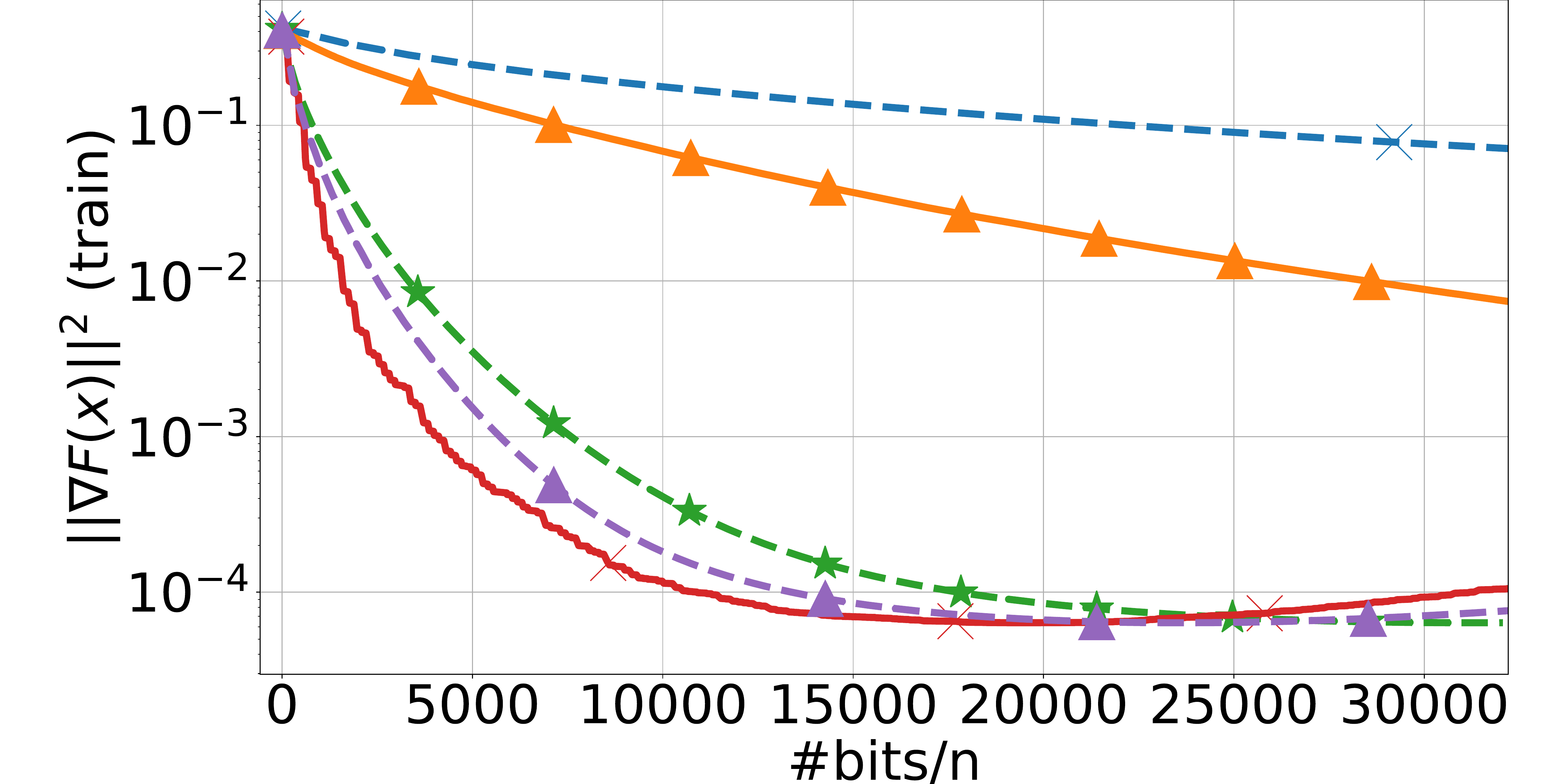}  \caption*{\hspace{20pt}$(a_1)$\,\texttt{mushroom}}
	\end{subfigure}
	\begin{subfigure}[ht]{0.32\textwidth}
		\includegraphics[width=\textwidth]{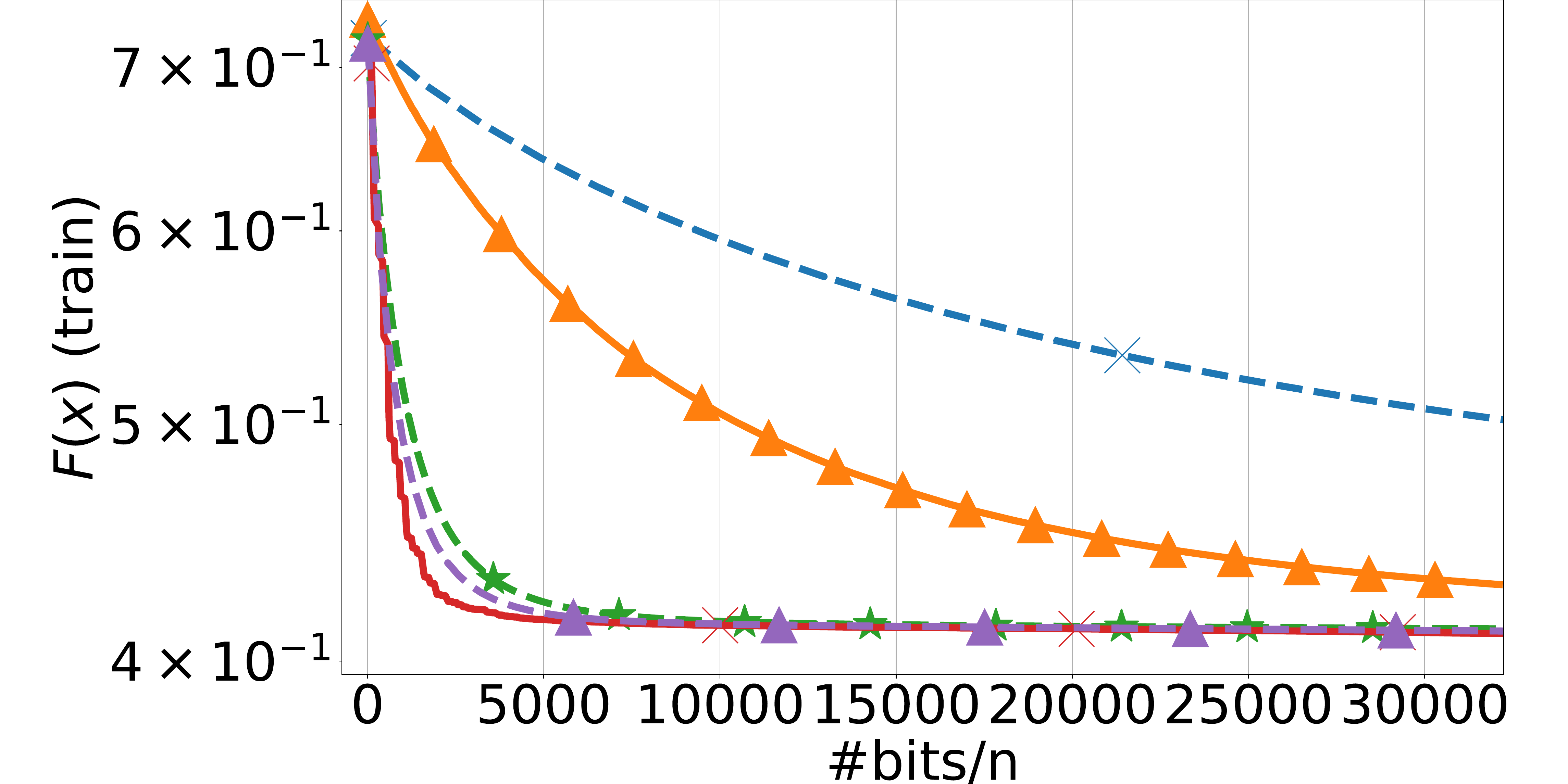}  \caption*{\hspace{20pt}$(a_2)$\, \texttt{mushroom}}
	\end{subfigure}
	\begin{subfigure}[ht]{0.32\textwidth}
		\includegraphics[width=\textwidth]{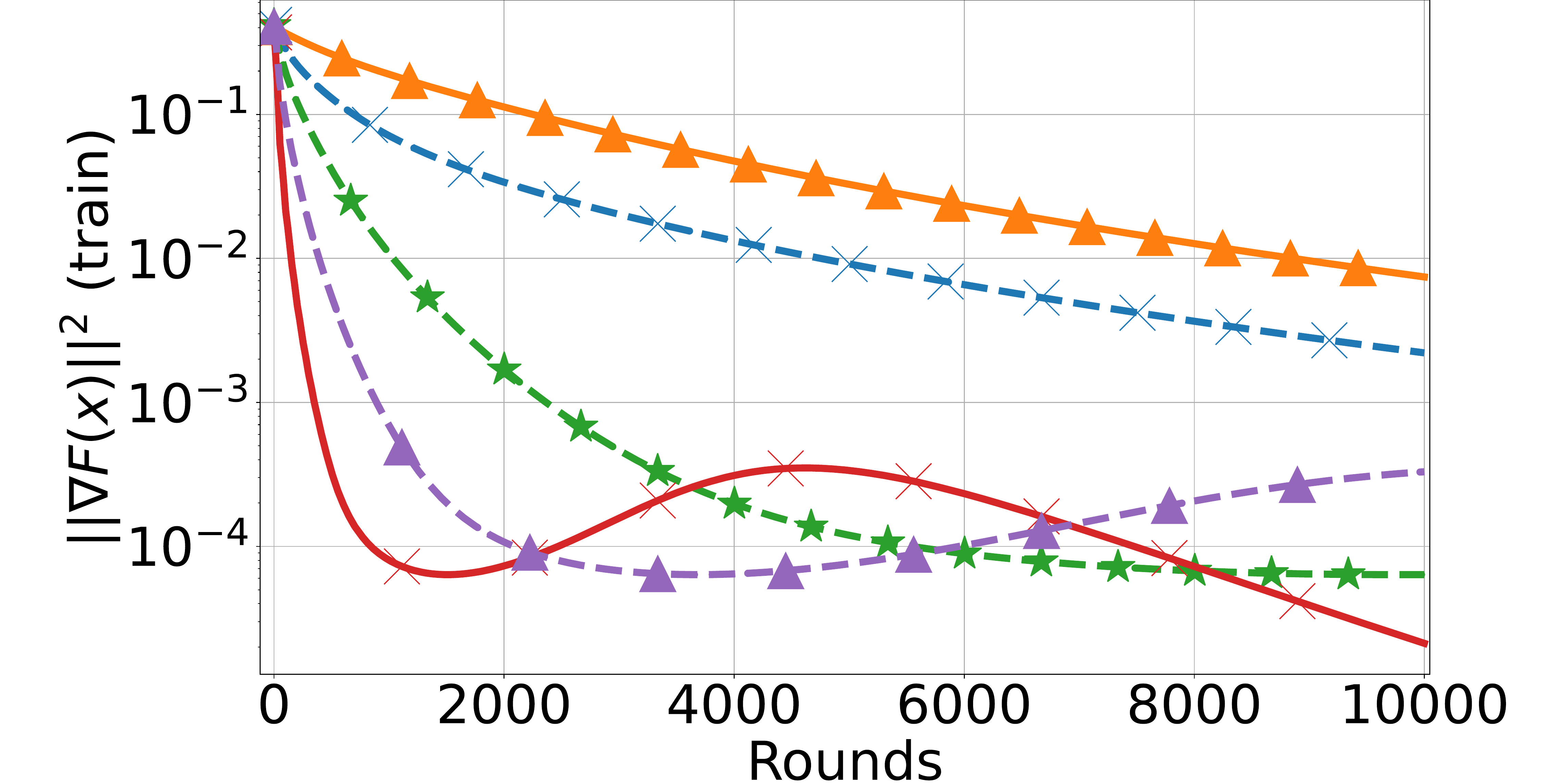}  \caption*{\hspace{20pt}$(a_3)$\,\texttt{mushroom}}
	\end{subfigure}
	
	\begin{subfigure}[ht]{0.32\textwidth}
		\includegraphics[width=\textwidth]{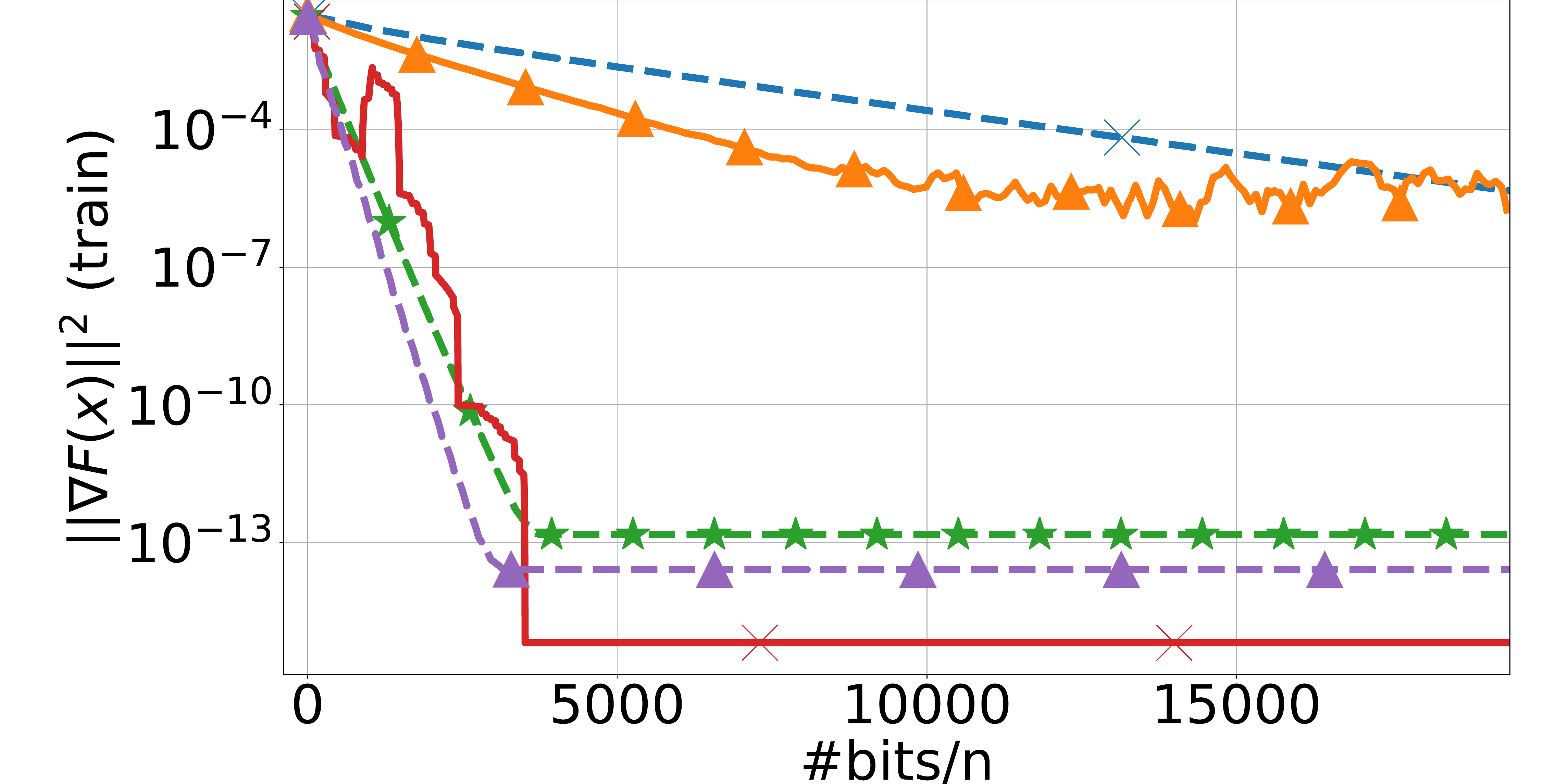}  \caption*{\hspace{20pt}$(b_1)$\,\texttt{phishing}}
	\end{subfigure}
	\begin{subfigure}[ht]{0.32\textwidth}
		\includegraphics[width=\textwidth]{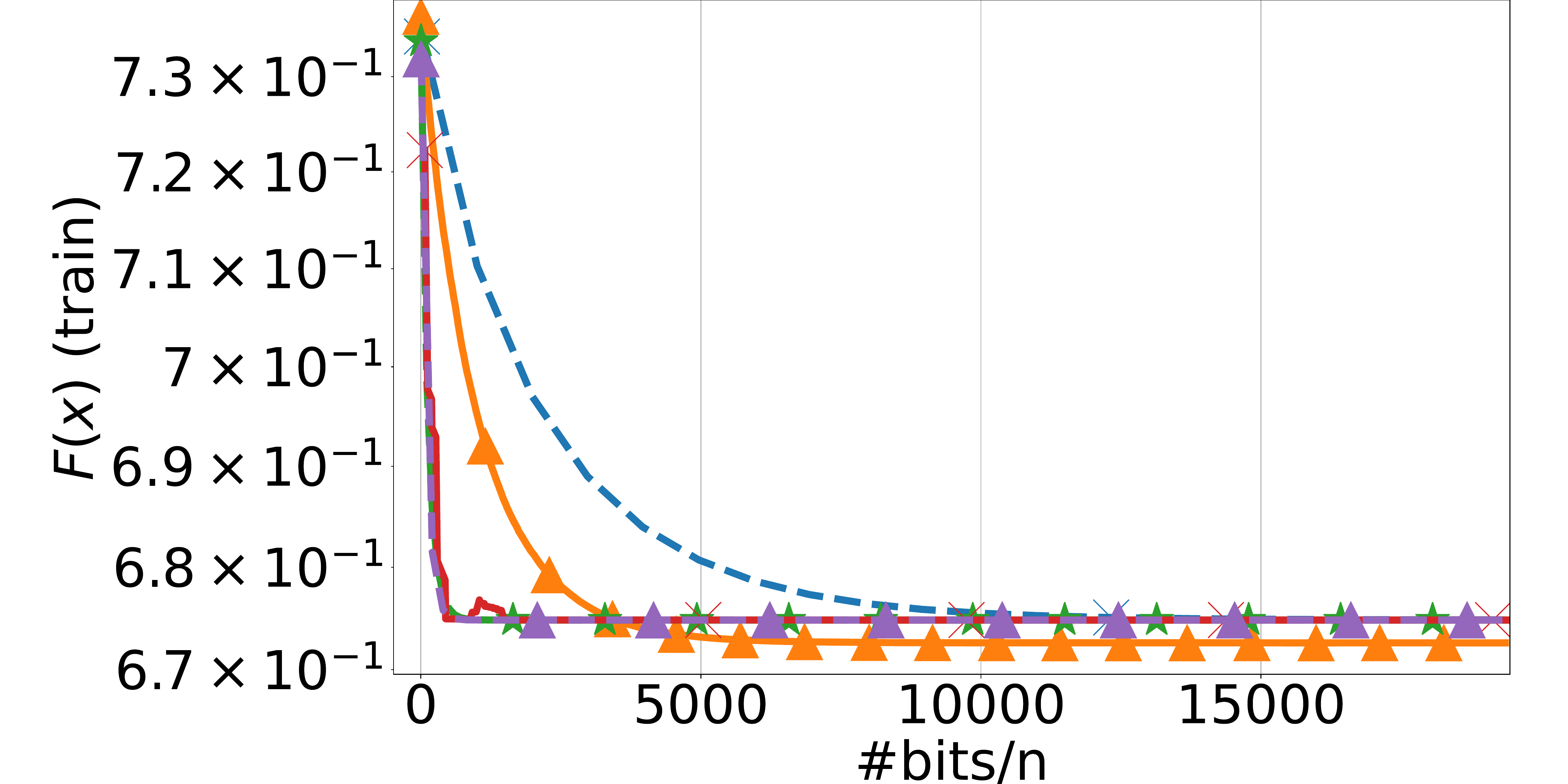}  \caption*{\hspace{20pt}$(b_2)$\,\texttt{phishing}}
	\end{subfigure}
	\begin{subfigure}[ht]{0.32\textwidth}
		\includegraphics[width=\textwidth]{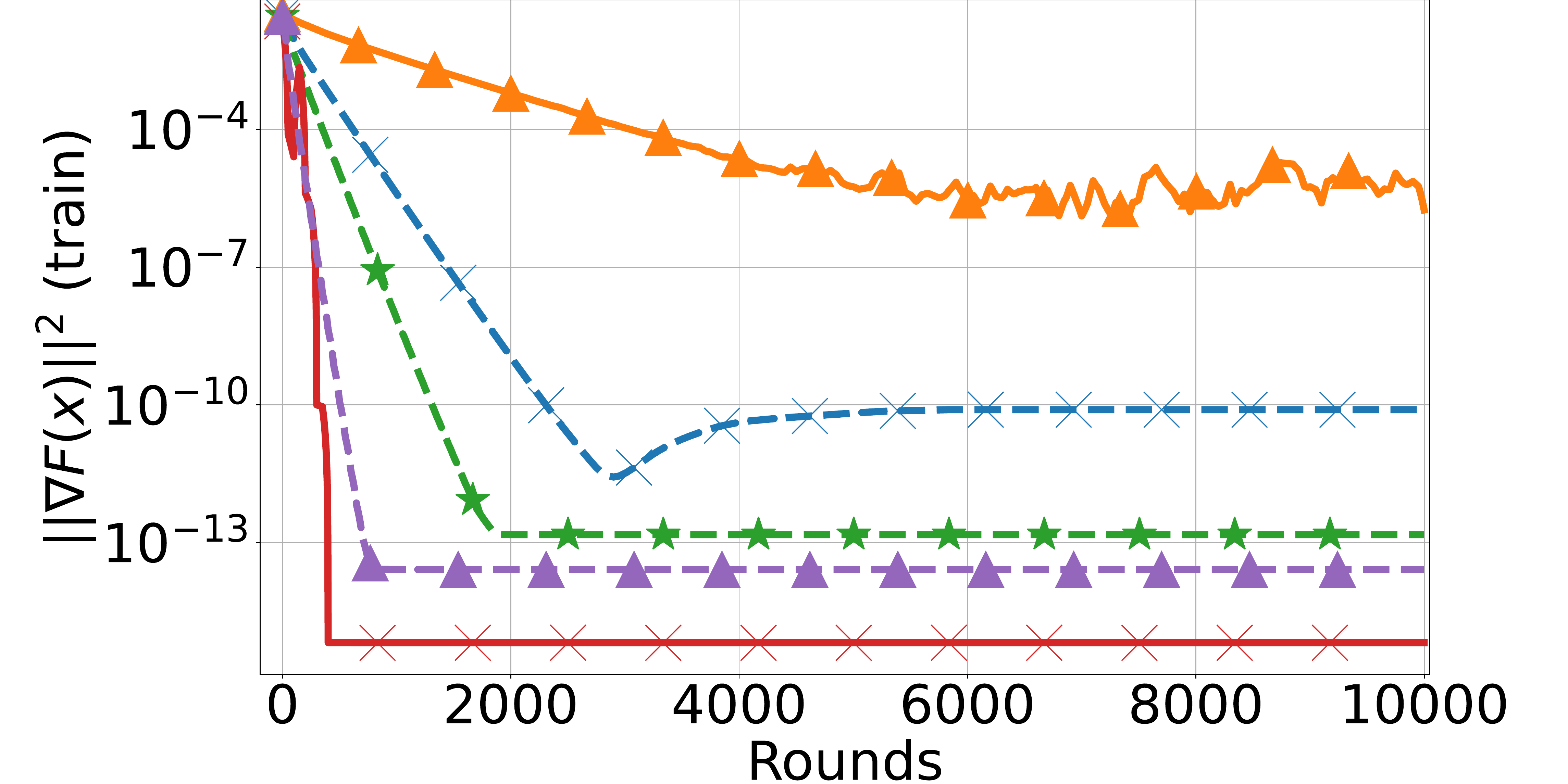}  \caption*{\hspace{20pt}$(b_3)$\,\texttt{phishing}}
	\end{subfigure}
	
	\begin{subfigure}[ht]{0.32\textwidth}
		\includegraphics[width=\textwidth]{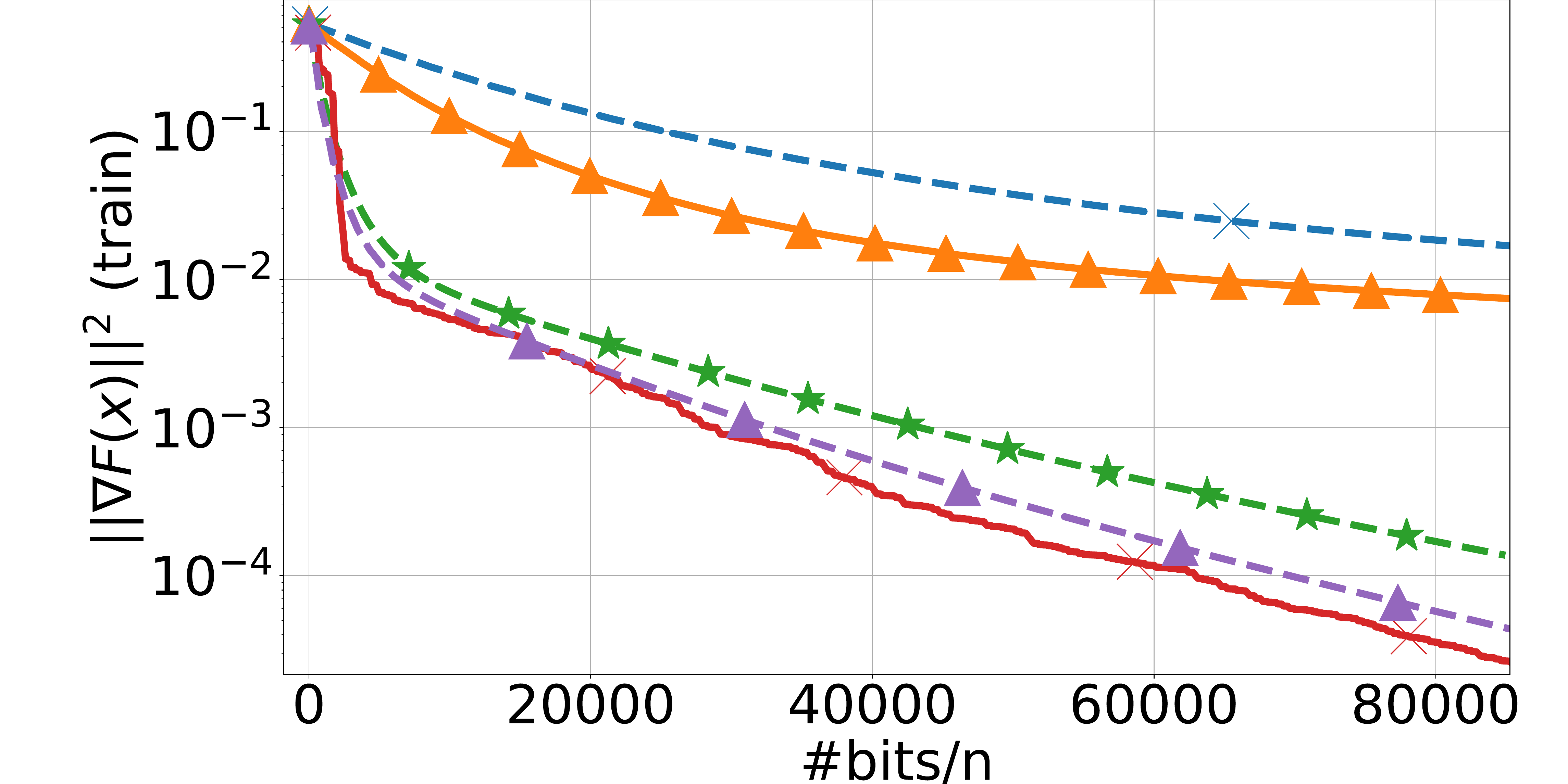}  \caption*{\hspace{20pt}$(c_1)$\, \texttt{w8a}}
	\end{subfigure}
	\begin{subfigure}[ht]{0.32\textwidth}
		\includegraphics[width=\textwidth]{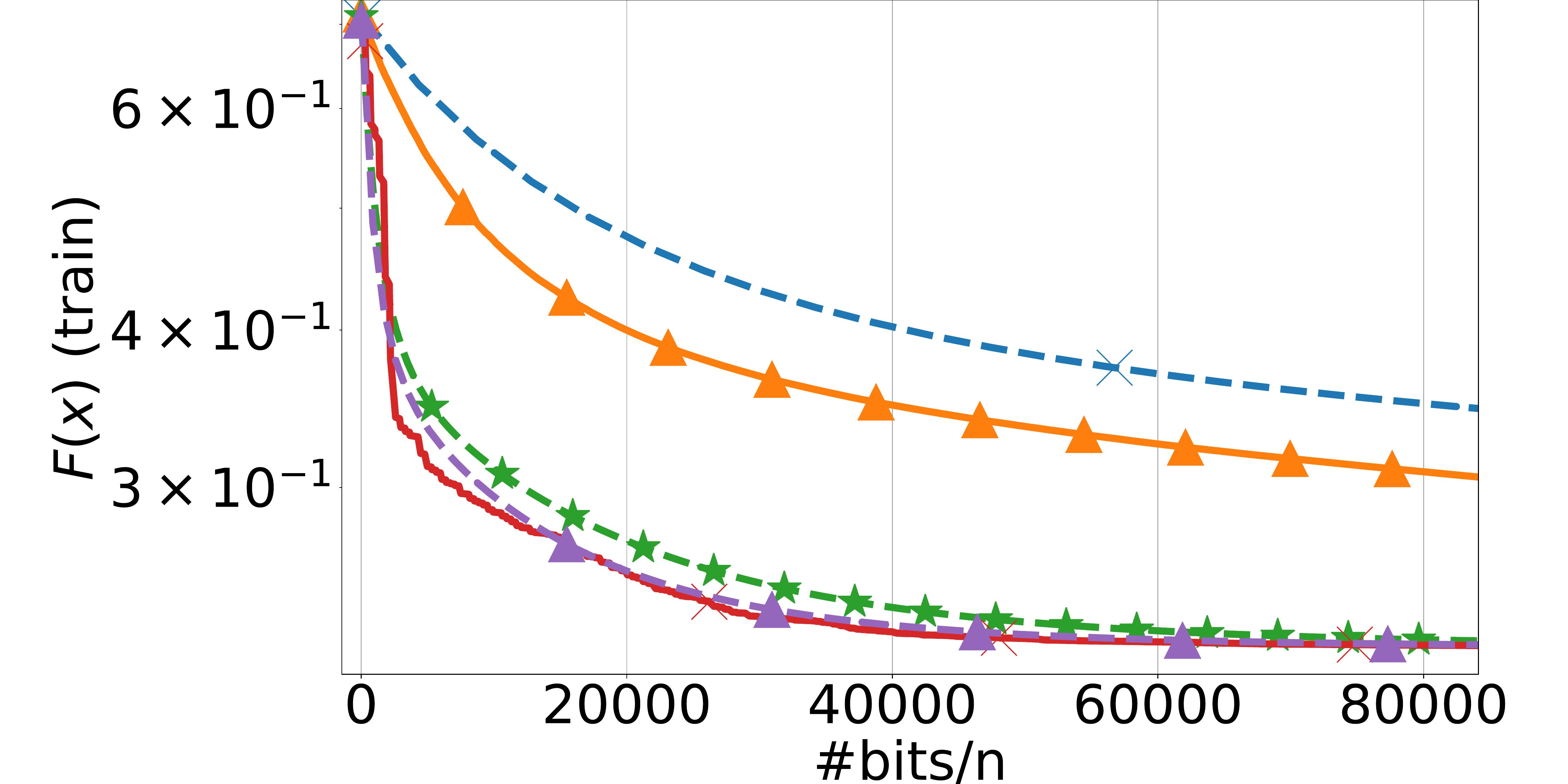}  \caption*{\hspace{20pt}$(c_2)$\, \texttt{w8a}}
	\end{subfigure}
	\begin{subfigure}[ht]{0.32\textwidth}
		\includegraphics[width=\textwidth]{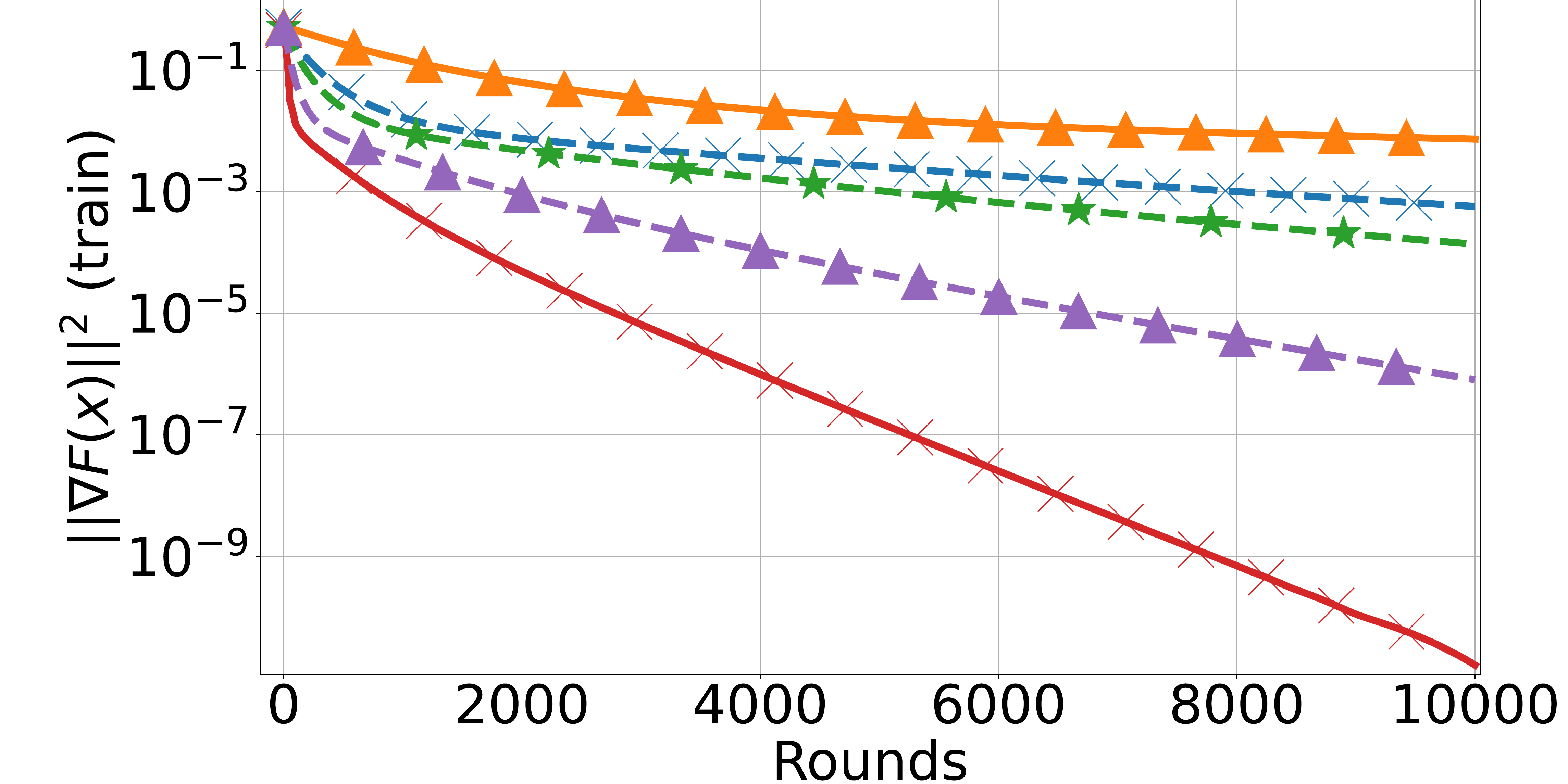}  \caption*{\hspace{20pt}$(c_3)$\, \texttt{w8a}}
	\end{subfigure}
	
	\begin{subfigure}[ht]{0.32\textwidth}
		\includegraphics[width=\textwidth]{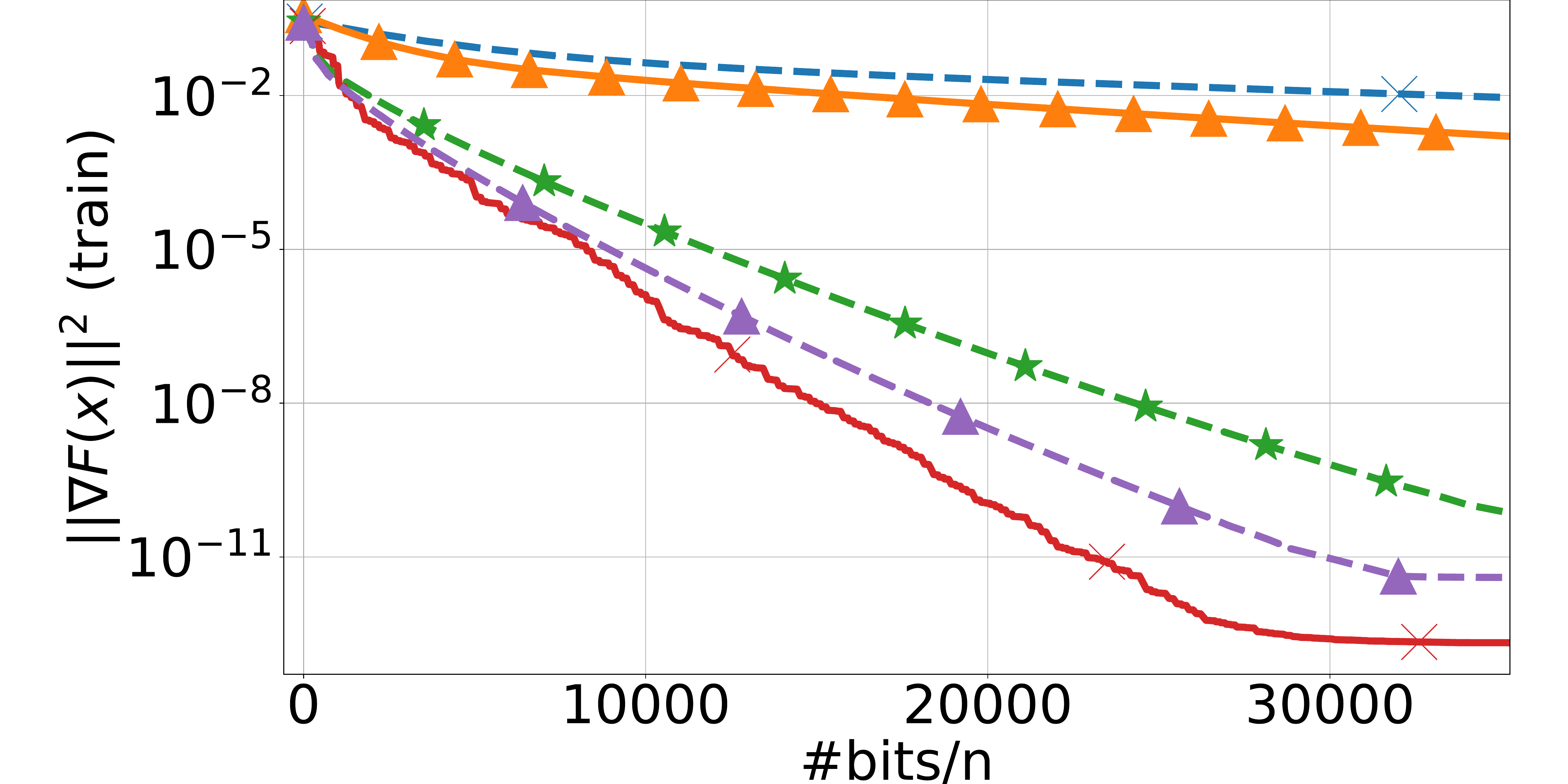}  \caption*{\hspace{20pt}$(d_1)$\, \texttt{a9a}}
	\end{subfigure}
	\begin{subfigure}[ht]{0.32\textwidth}
		\includegraphics[width=\textwidth]{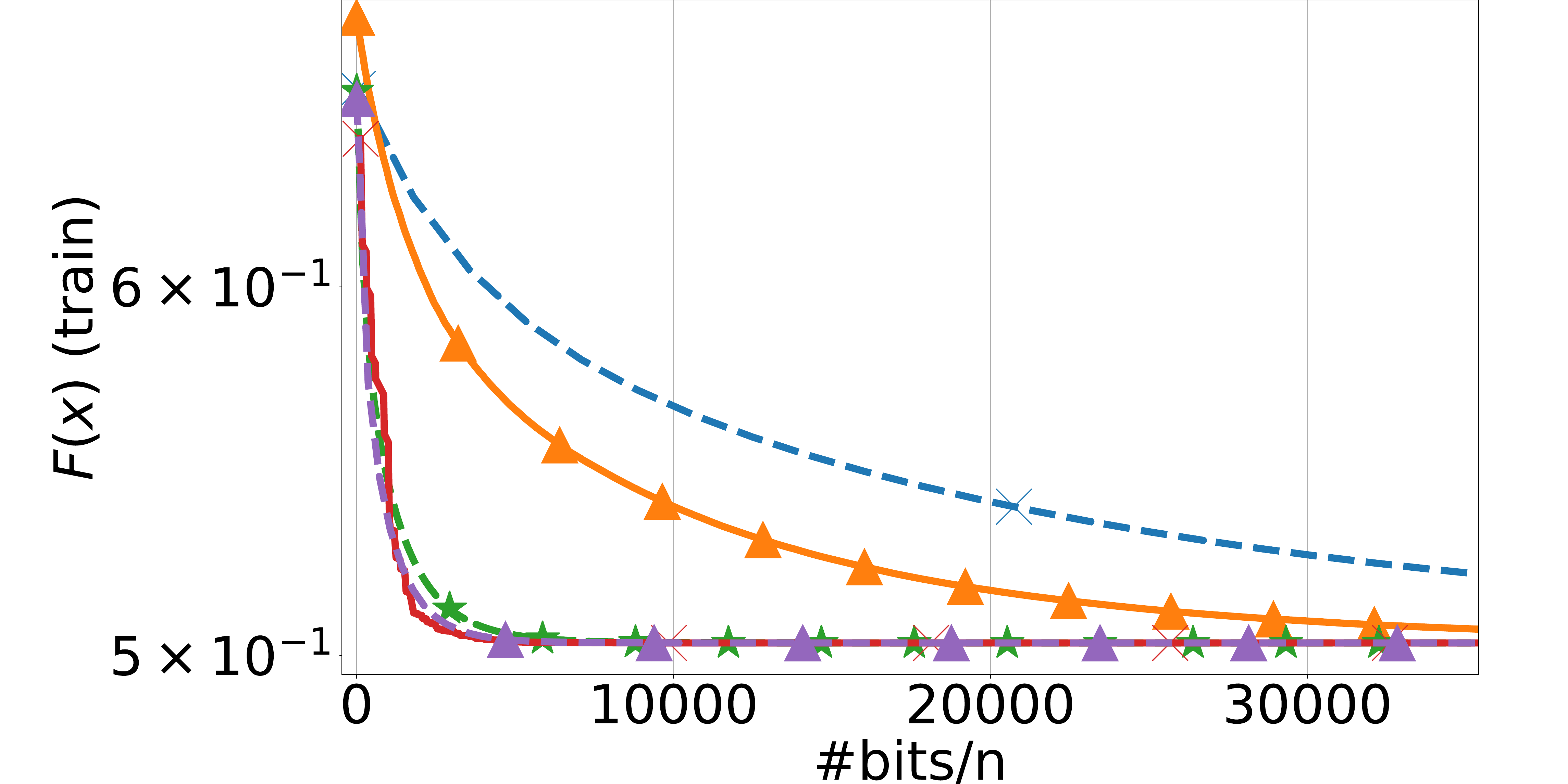}  \caption*{\hspace{20pt}$(d_2)$\, \texttt{a9a}}
	\end{subfigure}
	\begin{subfigure}[ht]{0.32\textwidth}
		\includegraphics[width=\textwidth]{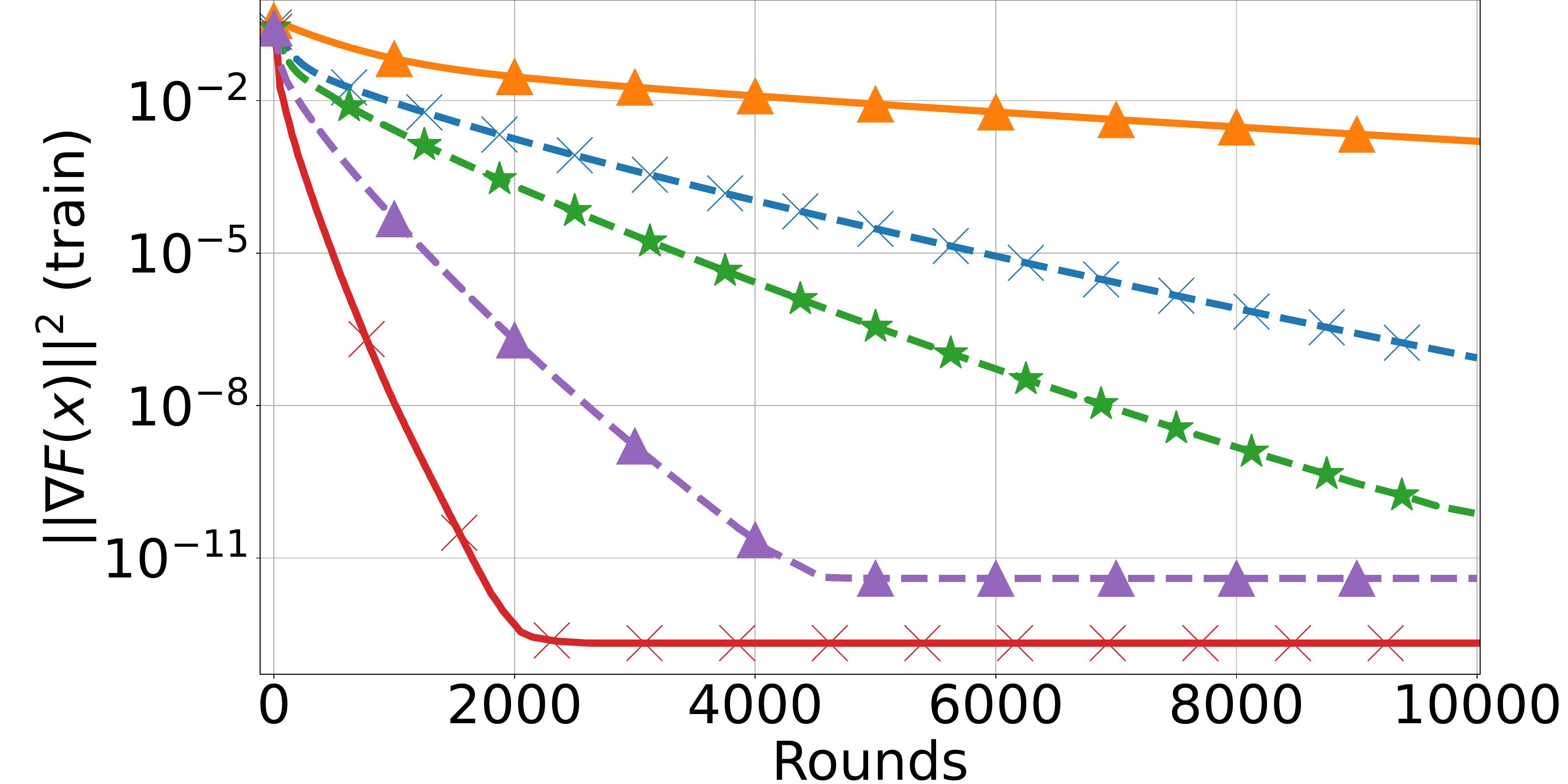}  \caption*{\hspace{20pt}$(d_3)$\, \texttt{a9a}}
	\end{subfigure}
	
	\vspace{-3pt}
	\caption{\small{Results for logistic regression with nonconvex regularizer on several \texttt{LIBSVM} datasets with $100$ homogeneous clients. We use \texttt{Natural} compressor on the client-side. The step sizes we choose are the most aggressive step sizes according to the non-convex theory of different algorithms.}}
	\label{fig:training_logreg_non_convex_rr}
\end{figure*}

\begin{figure*}[!ht]
	\centering
	\captionsetup[sub]{font=scriptsize,labelfont={}}	
	\begin{subfigure}[ht]{0.75\textwidth}
		\includegraphics[width=\textwidth]{./imgs/logregexp1/legend.pdf}  \caption*{}
	\end{subfigure}
	
	\begin{subfigure}[ht]{0.32\textwidth}
		\includegraphics[width=\textwidth]{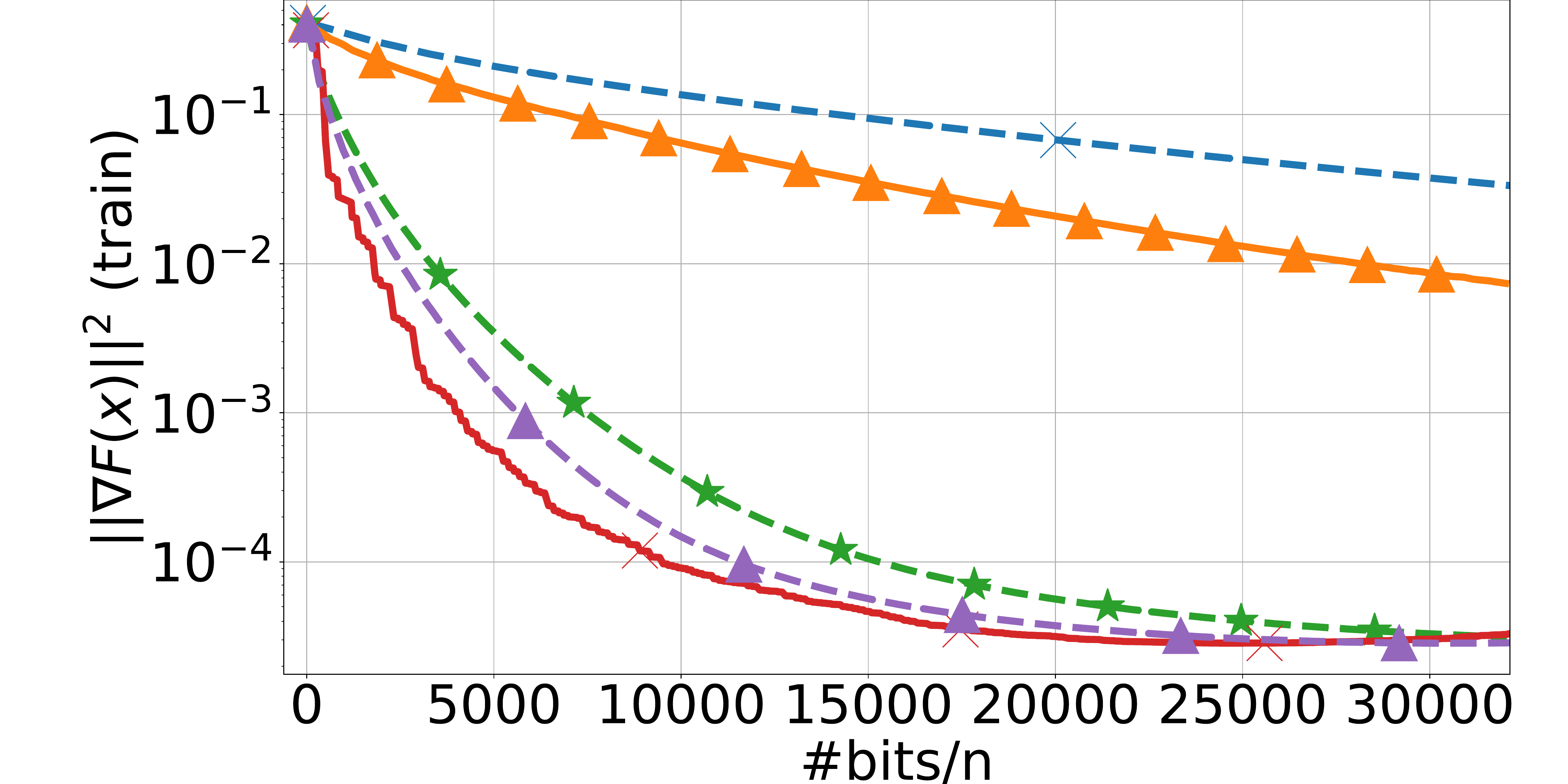}  \caption*{\hspace{20pt}$(a_1)$\, \texttt{mushroom}}
	\end{subfigure}
	\begin{subfigure}[ht]{0.32\textwidth}
		\includegraphics[width=\textwidth]{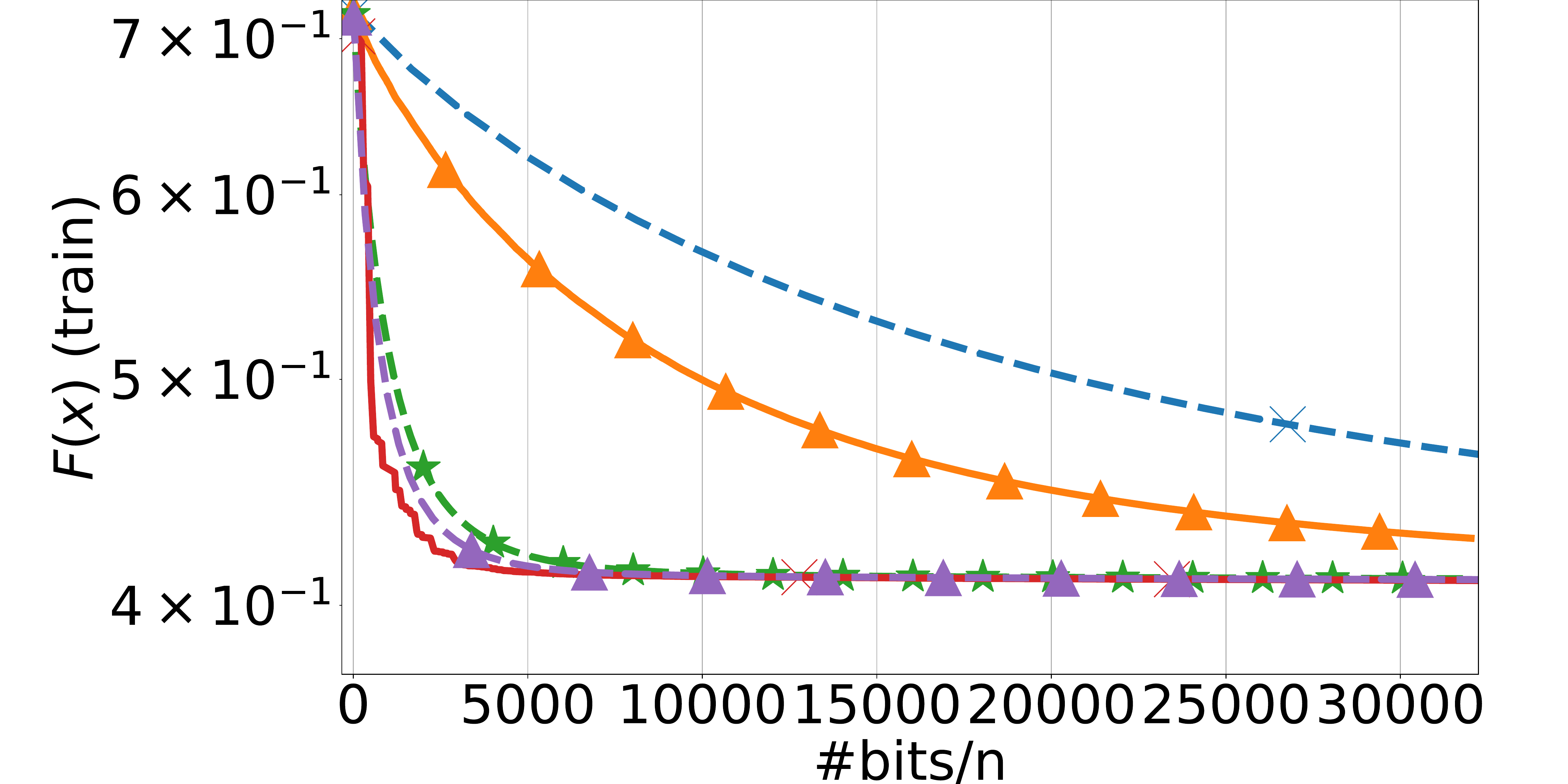}  \caption*{\hspace{20pt}$(a_2)$\, \texttt{mushroom}}
	\end{subfigure}
	\begin{subfigure}[ht]{0.32\textwidth}
		\includegraphics[width=\textwidth]{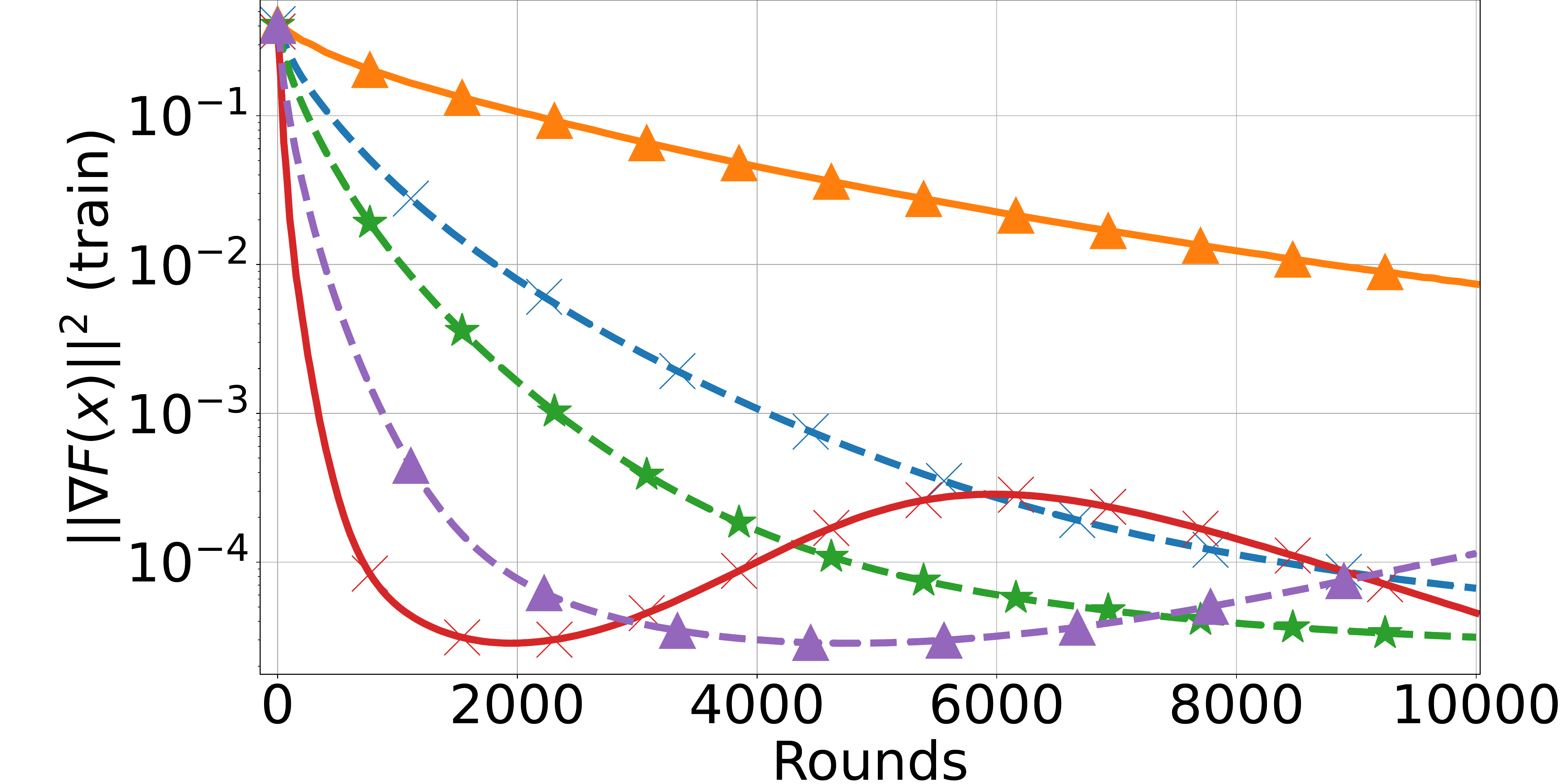}  \caption*{\hspace{20pt}$(a_3)$\, \texttt{mushroom}}
	\end{subfigure}
	
	\begin{subfigure}[ht]{0.32\textwidth}
		\includegraphics[width=\textwidth]{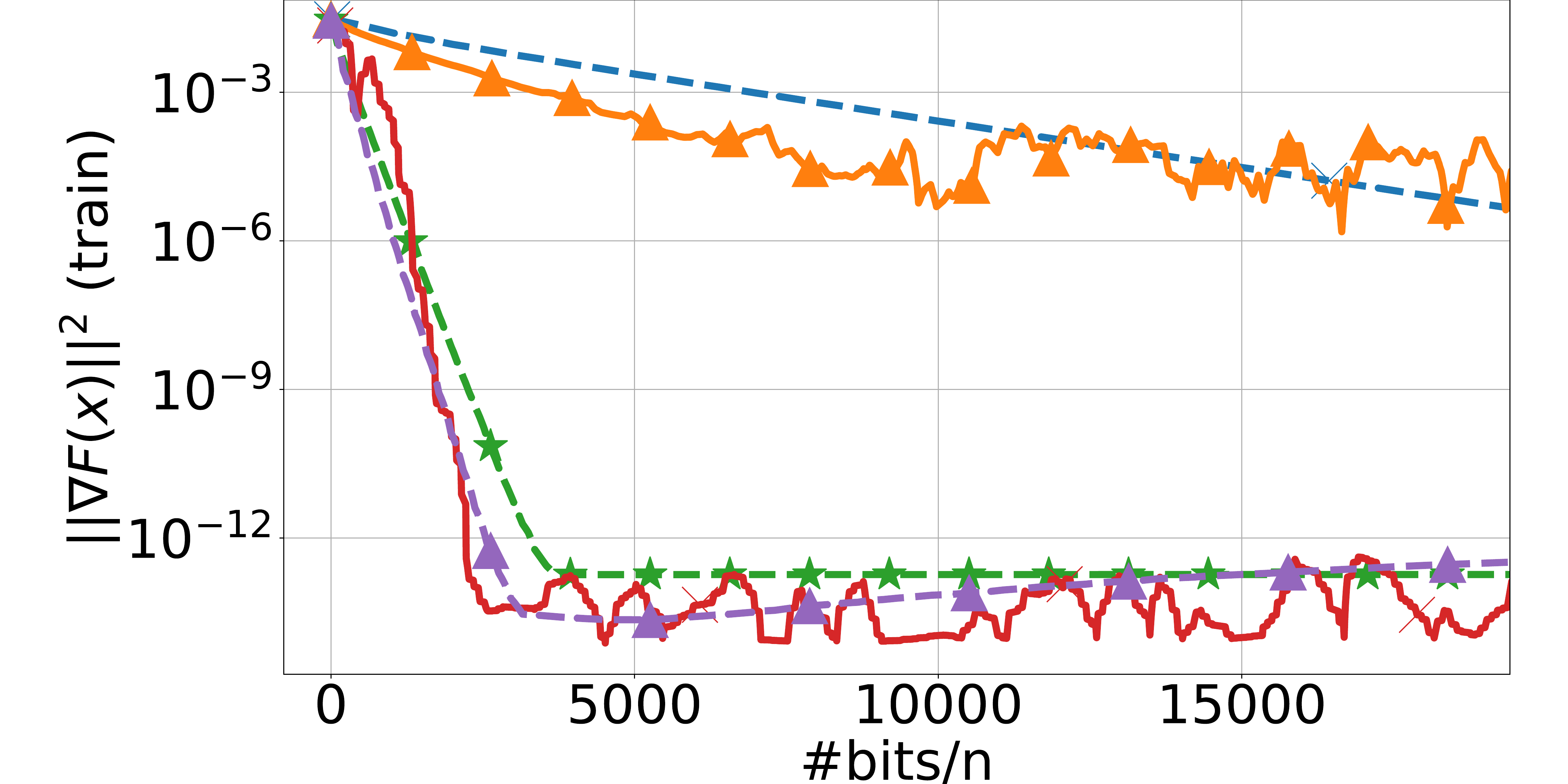}  \caption*{\hspace{20pt}$(b_1)$\, \texttt{phishing}}
	\end{subfigure}
	\begin{subfigure}[ht]{0.32\textwidth}
		\includegraphics[width=\textwidth]{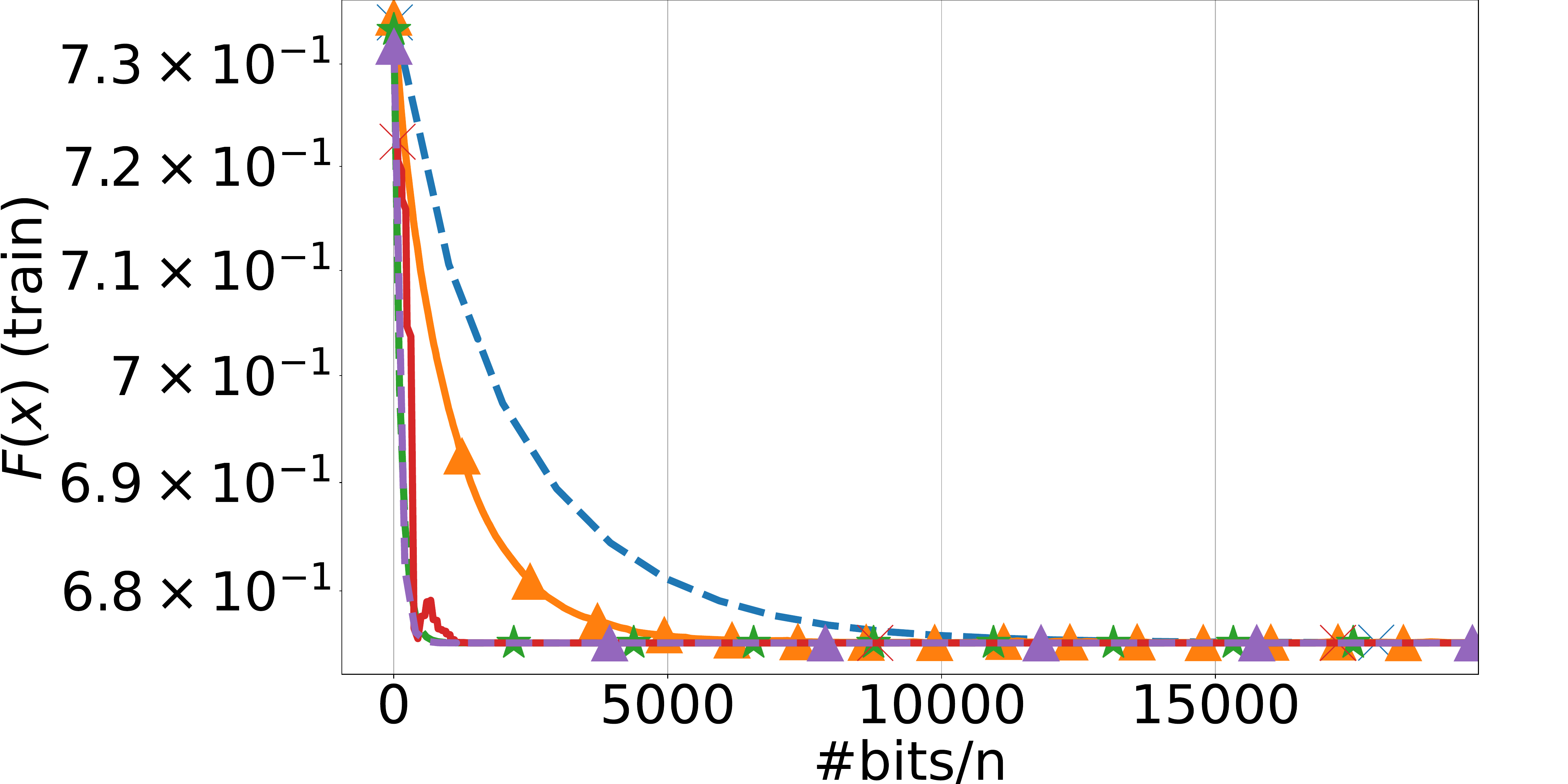}  \caption*{\hspace{20pt}$(b_2)$\, \texttt{phishing}}
	\end{subfigure}
	\begin{subfigure}[ht]{0.32\textwidth}
		\includegraphics[width=\textwidth]{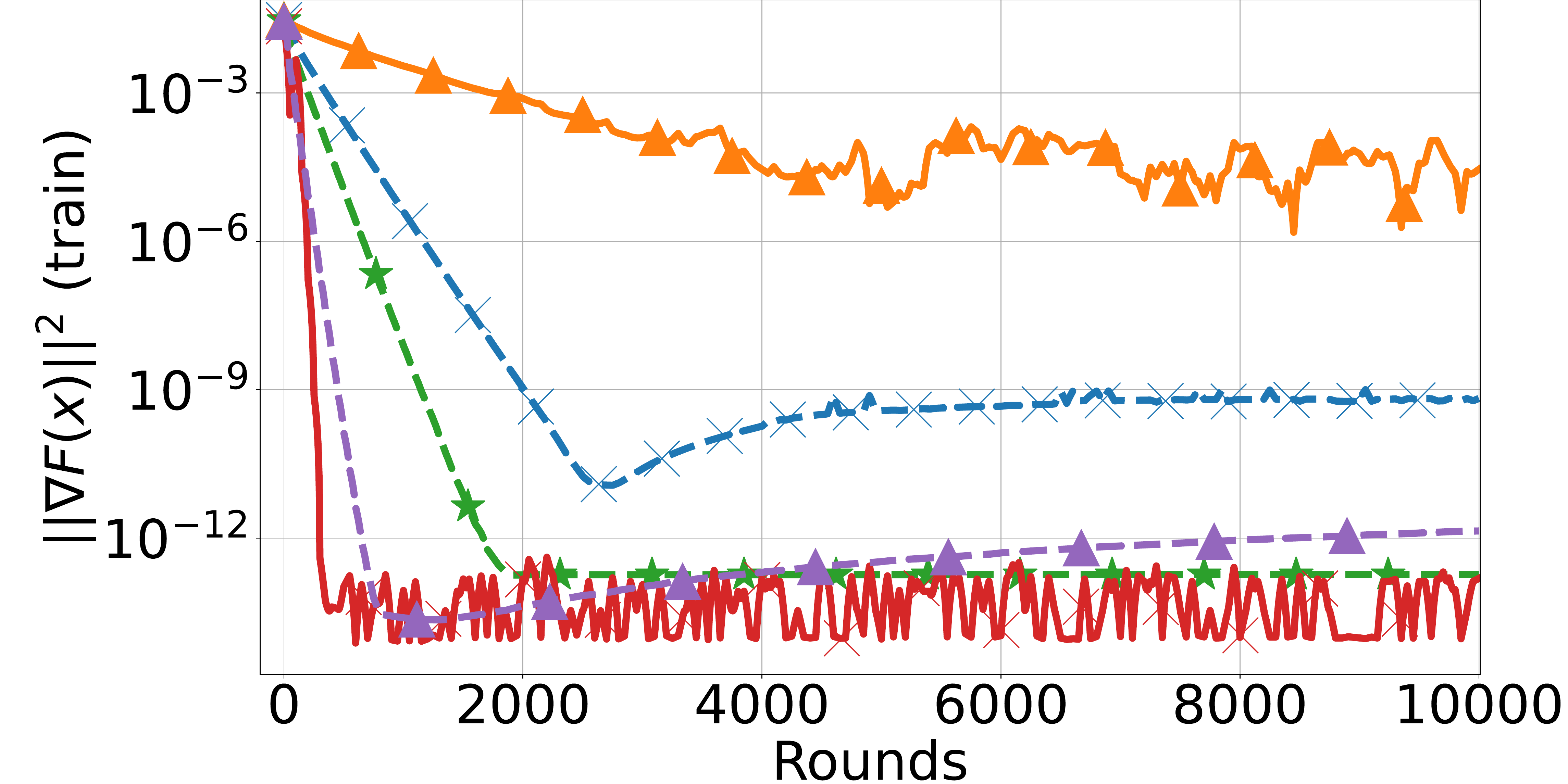}  \caption*{\hspace{20pt}$(b_3)$\, \texttt{phishing}}
	\end{subfigure}
	
	\begin{subfigure}[ht]{0.32\textwidth}
		\includegraphics[width=\textwidth]{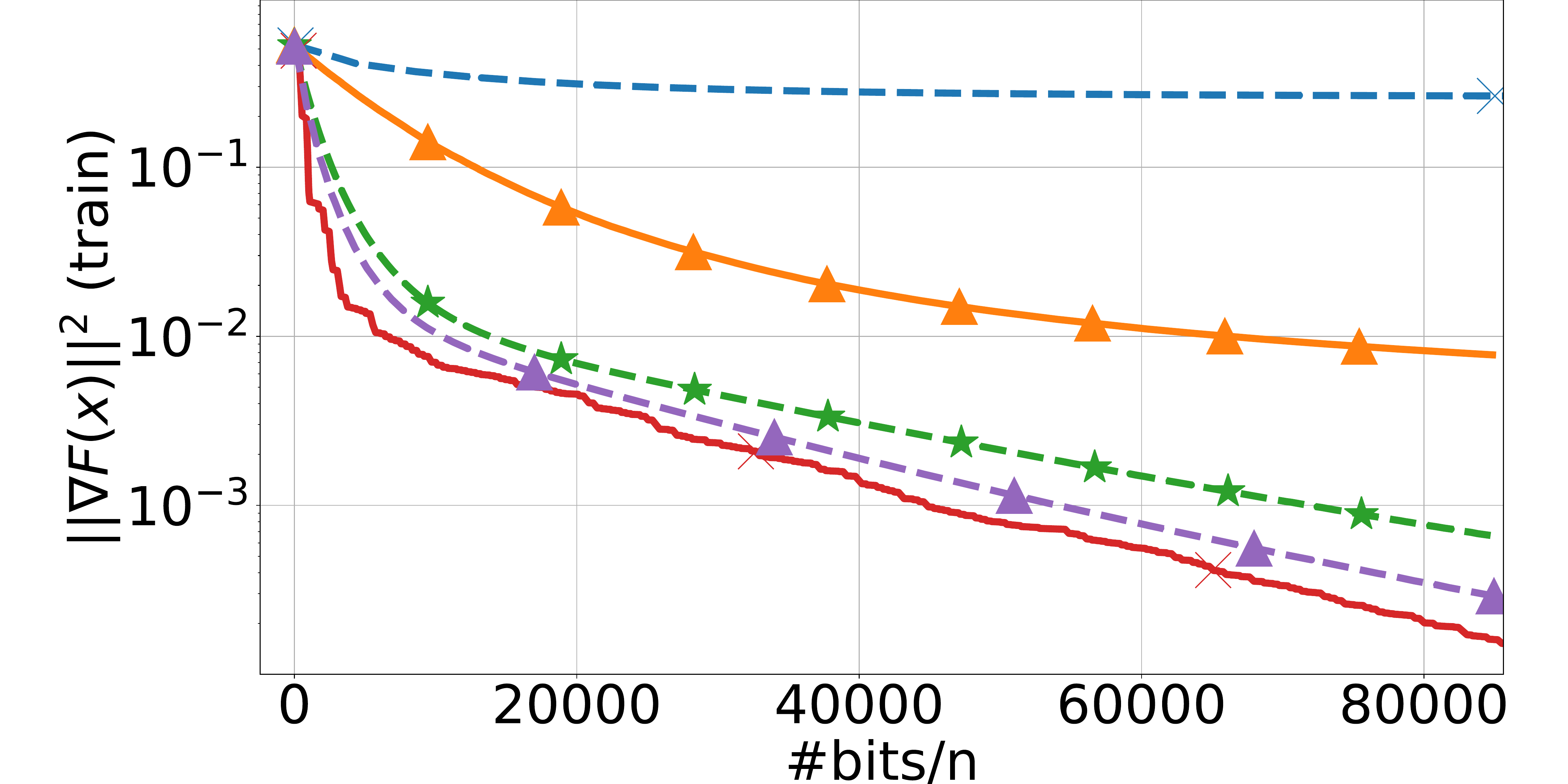}  \caption*{\hspace{20pt}$(c_1)$\, \texttt{w8a}}
	\end{subfigure}
	\begin{subfigure}[ht]{0.32\textwidth}
		\includegraphics[width=\textwidth]{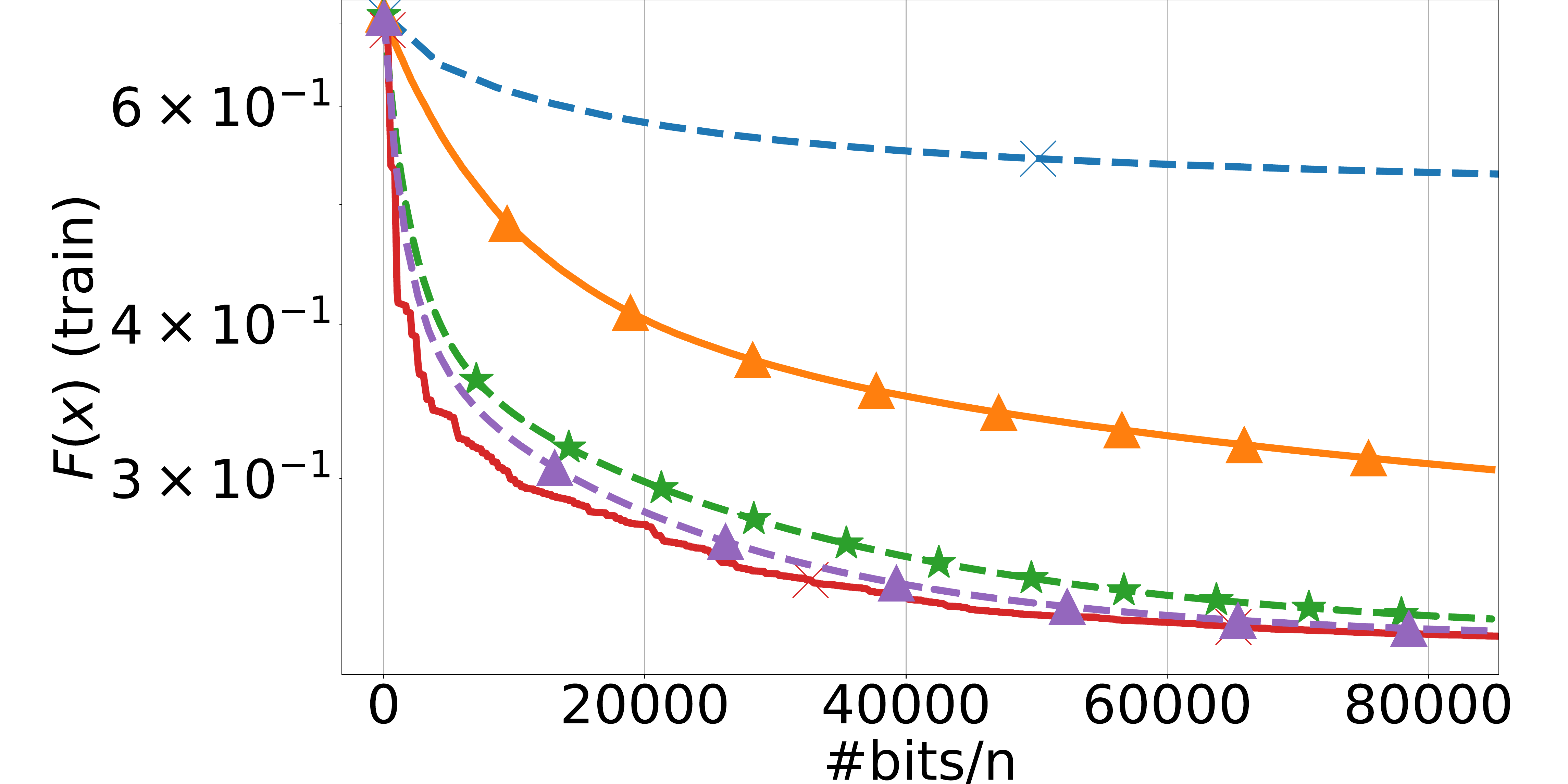}  \caption*{\hspace{20pt}$(c_2)$\, \texttt{w8a}}
	\end{subfigure}
	\begin{subfigure}[ht]{0.32\textwidth}
		\includegraphics[width=\textwidth]{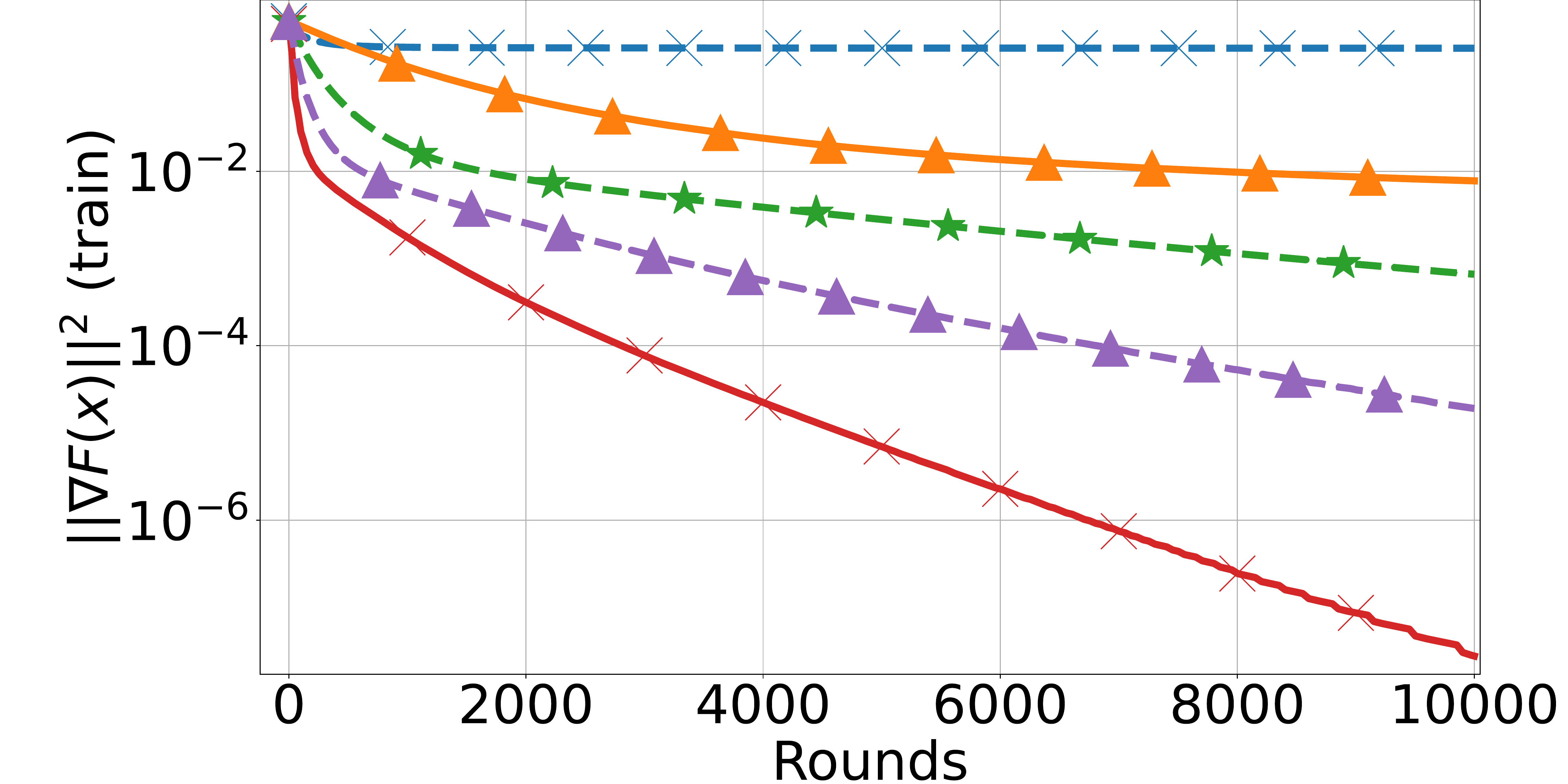}  \caption*{\hspace{20pt}$(c_3)$\, \texttt{w8a}}
	\end{subfigure}
	
	\begin{subfigure}[ht]{0.32\textwidth}
		\includegraphics[width=\textwidth]{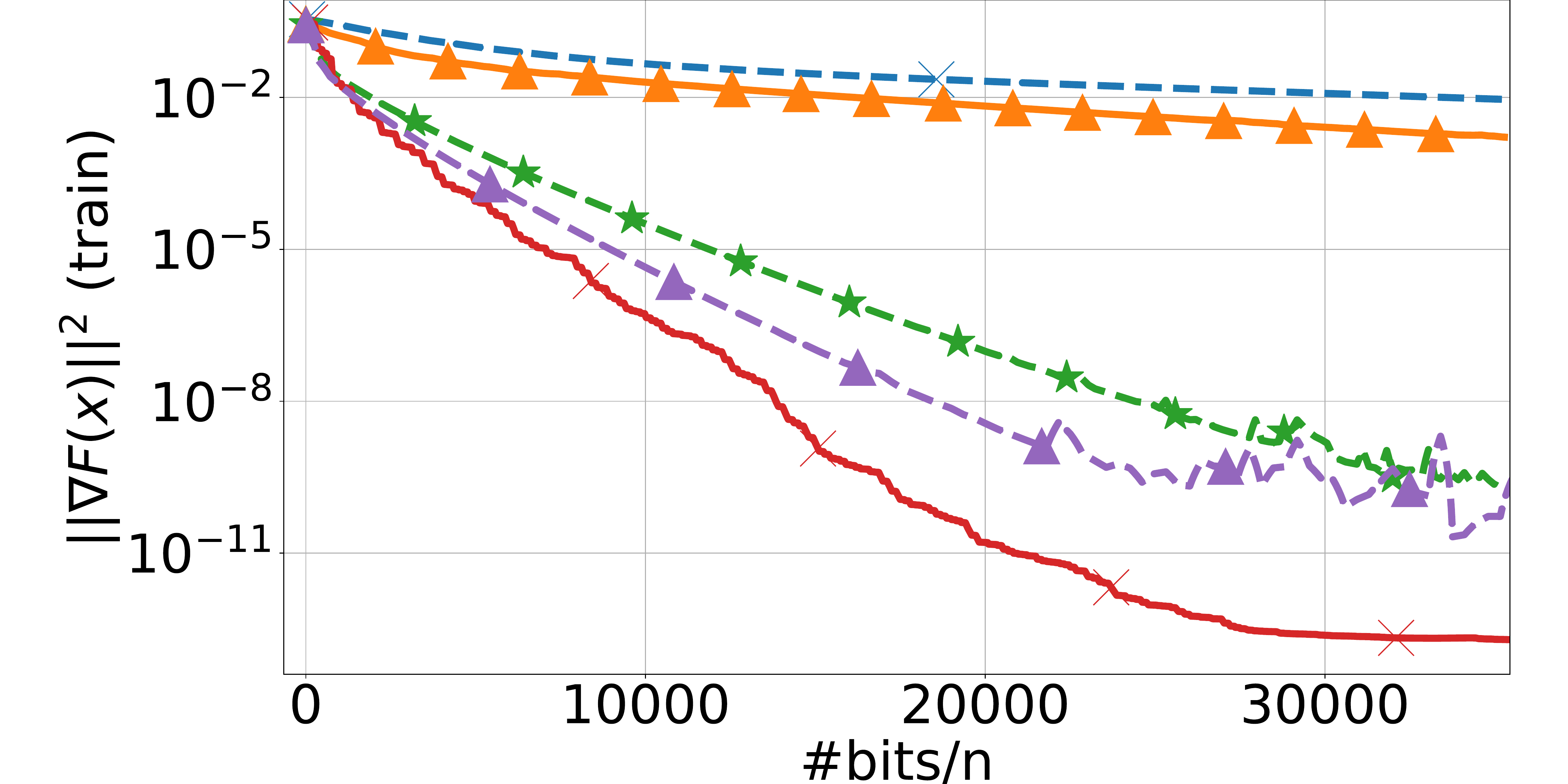}  \caption*{\hspace{20pt}$(d_1)$\, \texttt{a9a}}
	\end{subfigure}
	\begin{subfigure}[ht]{0.32\textwidth}
		\includegraphics[width=\textwidth]{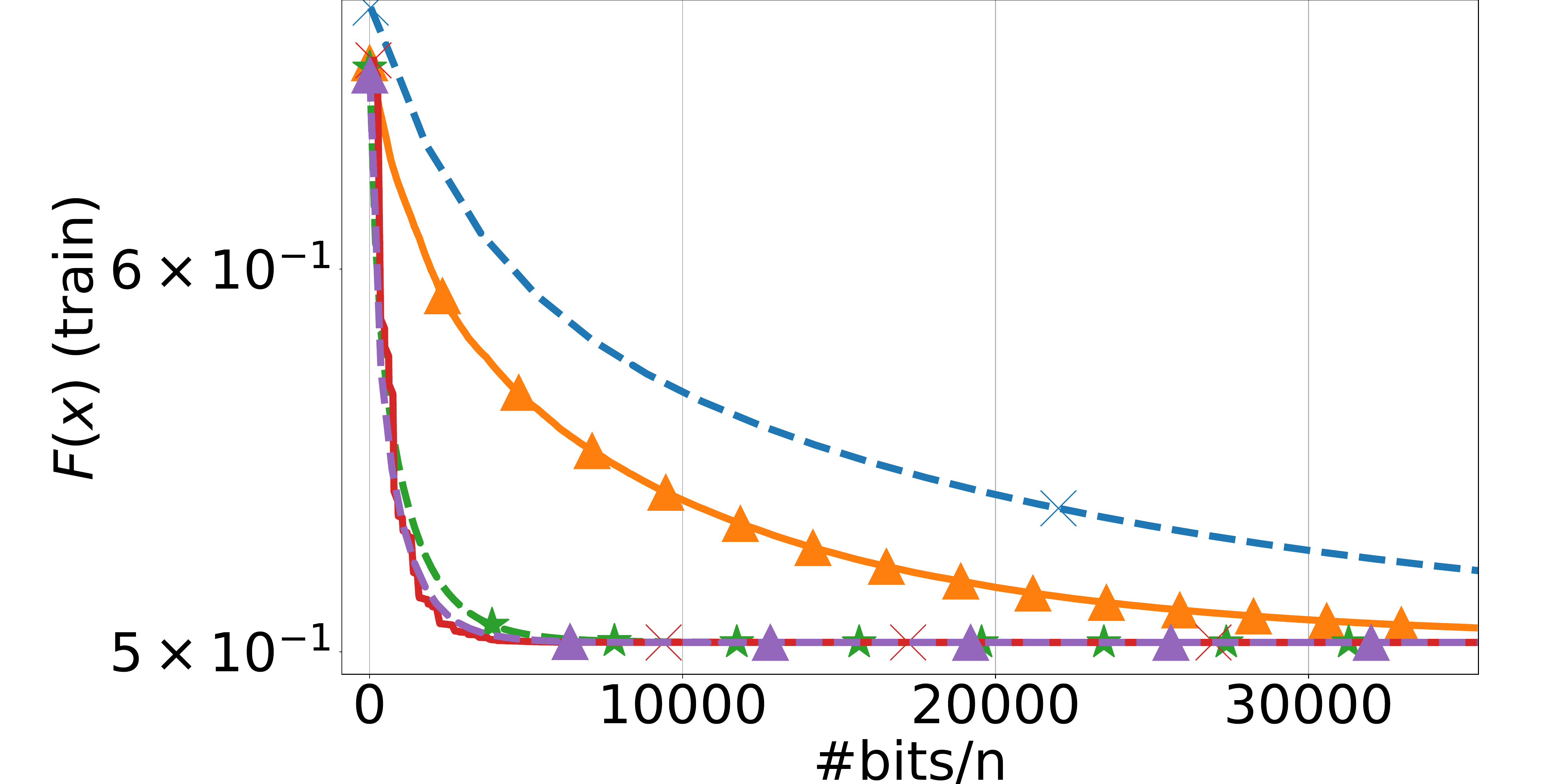}  \caption*{\hspace{20pt}$(d_2)$\, \texttt{a9a}}
	\end{subfigure}
	\begin{subfigure}[ht]{0.32\textwidth}
		\includegraphics[width=\textwidth]{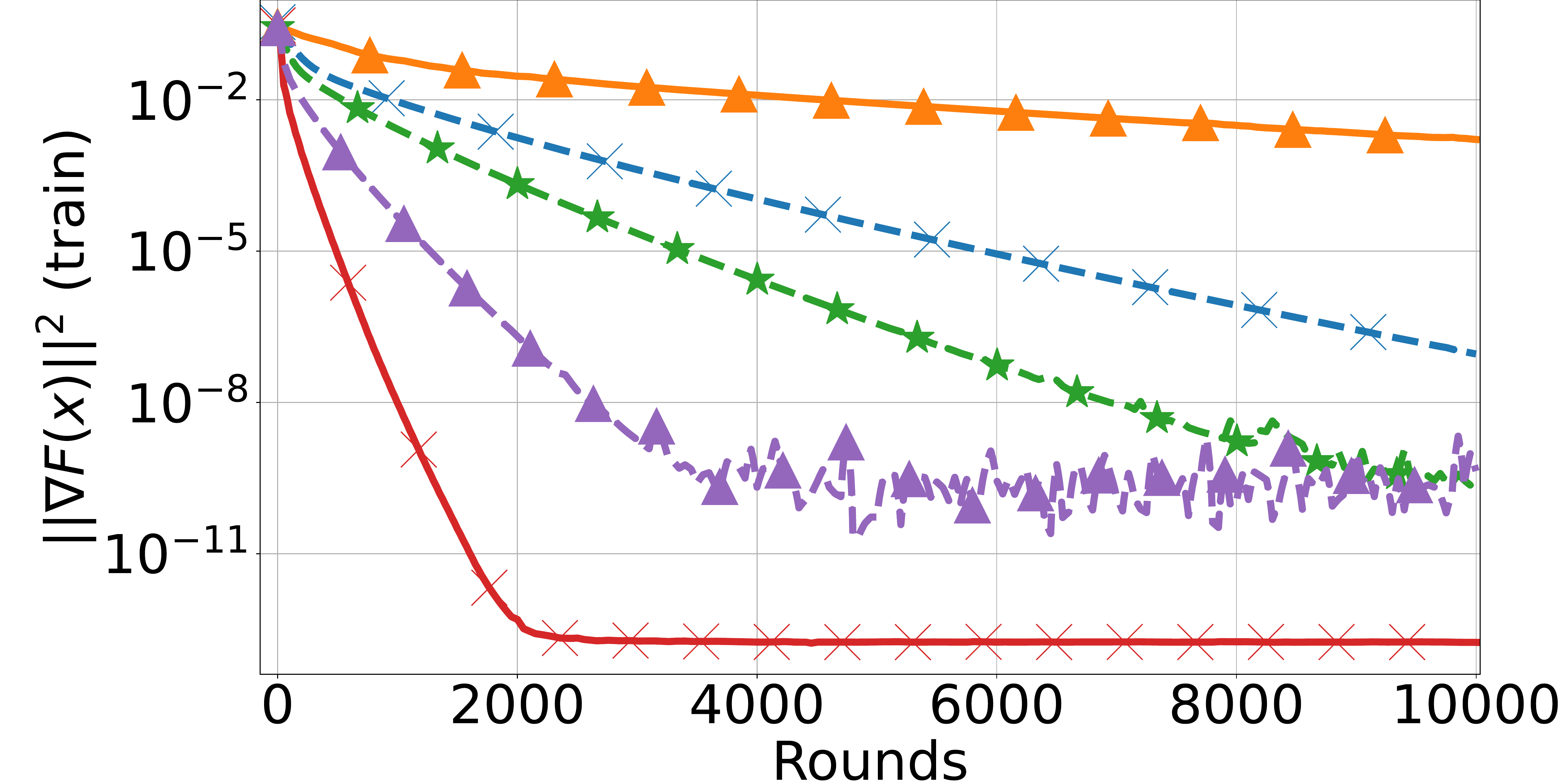}  \caption*{\hspace{20pt}$(d_3)$\, \texttt{a9a}}
	\end{subfigure}	
	
	\vspace{-3pt}
	\caption{\small{Results for logistic regression with nonconvex regularizer on several \texttt{LIBSVM} datasets with $100$ heterogeneous clients. We use \texttt{Natural} compressor on the client-side. The step sizes we choose are the most aggressive step sizes according to the non-convex theory of different algorithms.}}
	
	\label{fig:training_logreg_non_convex_hetero}
\end{figure*}

\begin{figure*}[!ht]
	\centering
	\captionsetup[sub]{font=scriptsize,labelfont={}}	
	\begin{subfigure}[ht]{0.4\textwidth}
		\includegraphics[width=\textwidth]{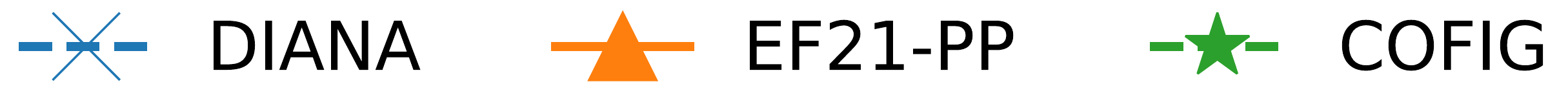}  \caption*{}
	\end{subfigure}
	
	\begin{subfigure}[ht]{0.32\textwidth}
		\includegraphics[width=\textwidth]{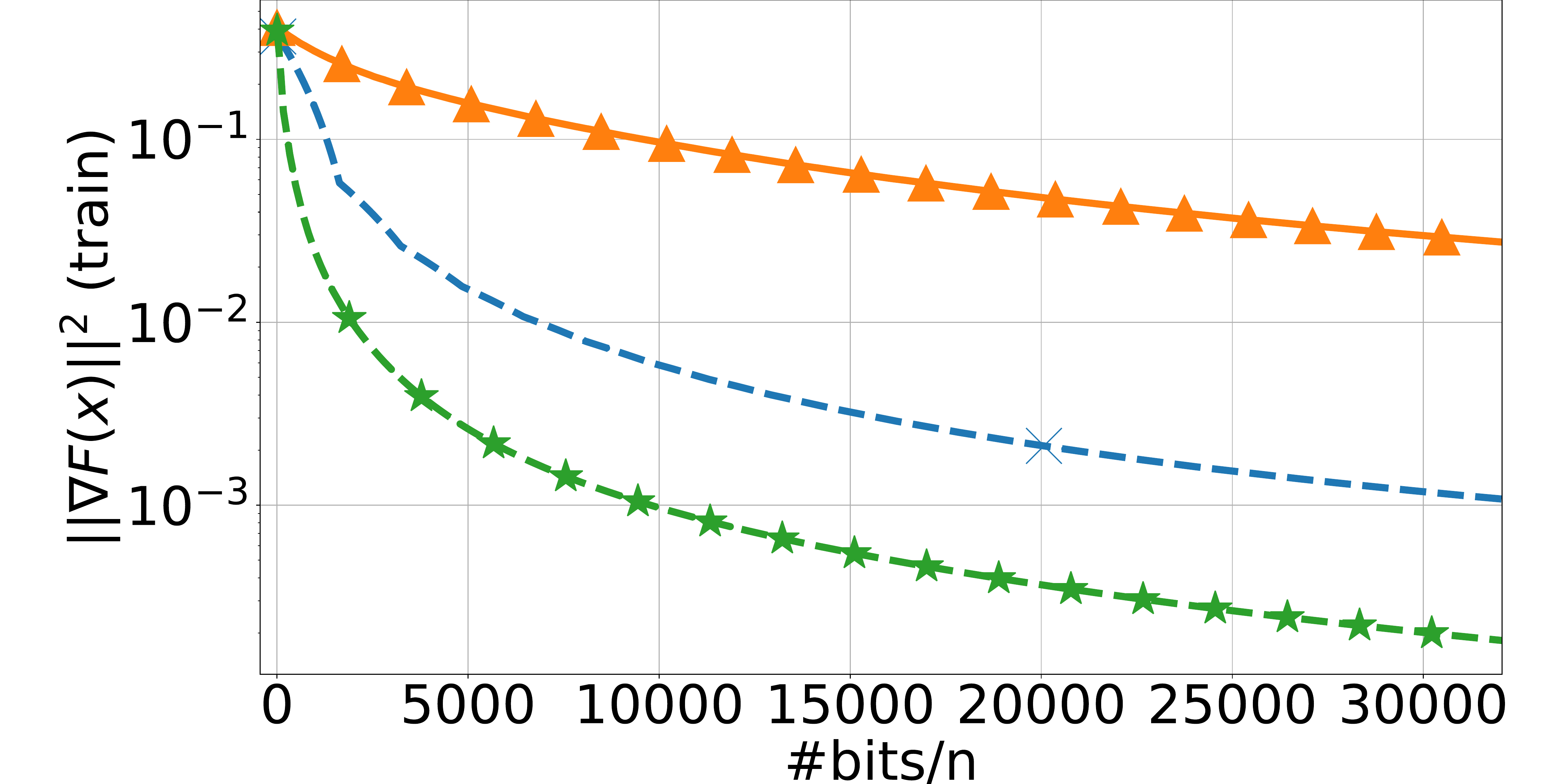}  \caption*{\hspace{20pt}$(a_1)$\, \texttt{mushroom}}
	\end{subfigure}
	\begin{subfigure}[ht]{0.32\textwidth}
		\includegraphics[width=\textwidth]{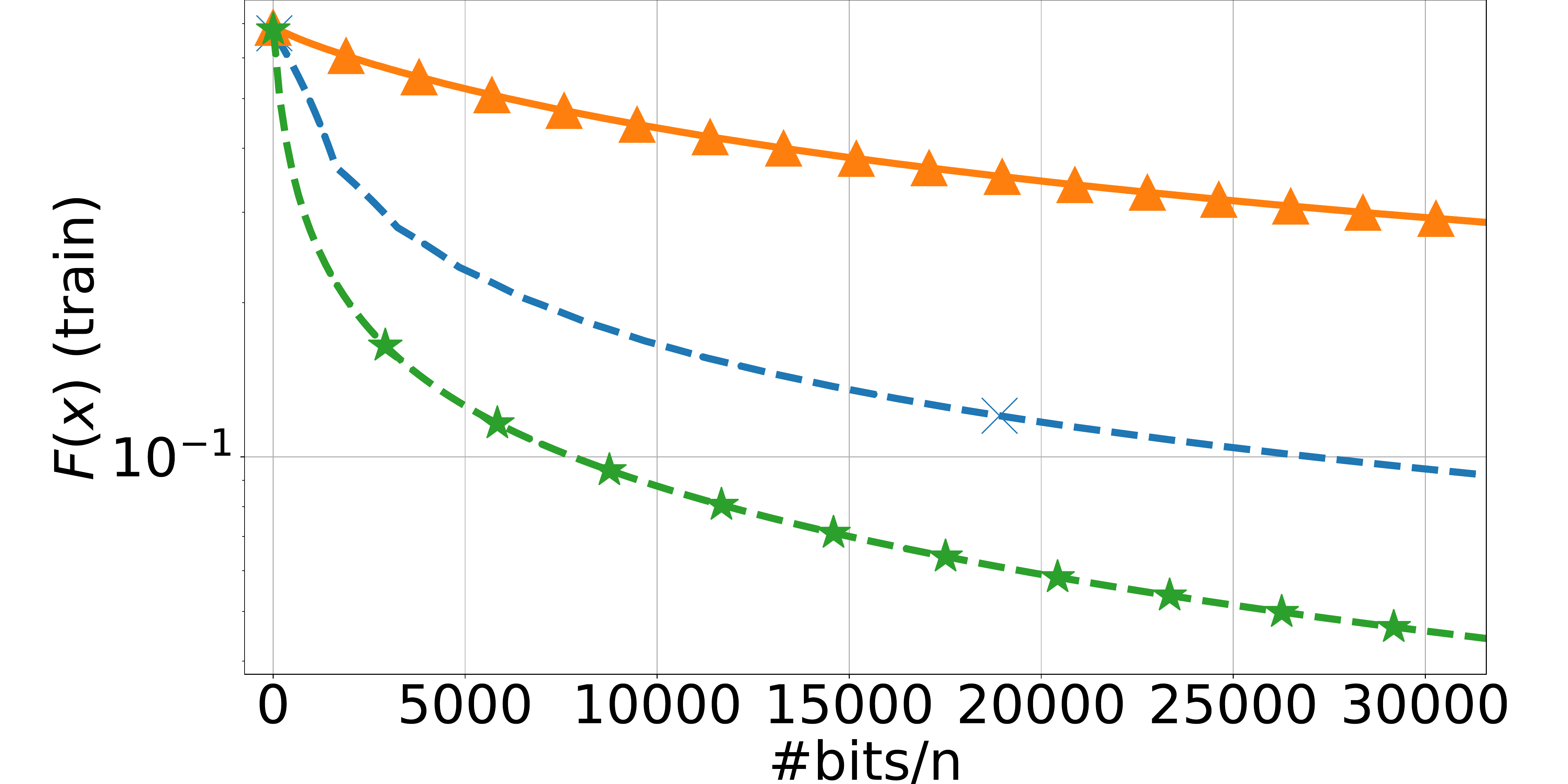}  \caption*{\hspace{20pt}$(a_2)$\, \texttt{mushroom}}
	\end{subfigure}
	\begin{subfigure}[ht]{0.32\textwidth}
		\includegraphics[width=\textwidth]{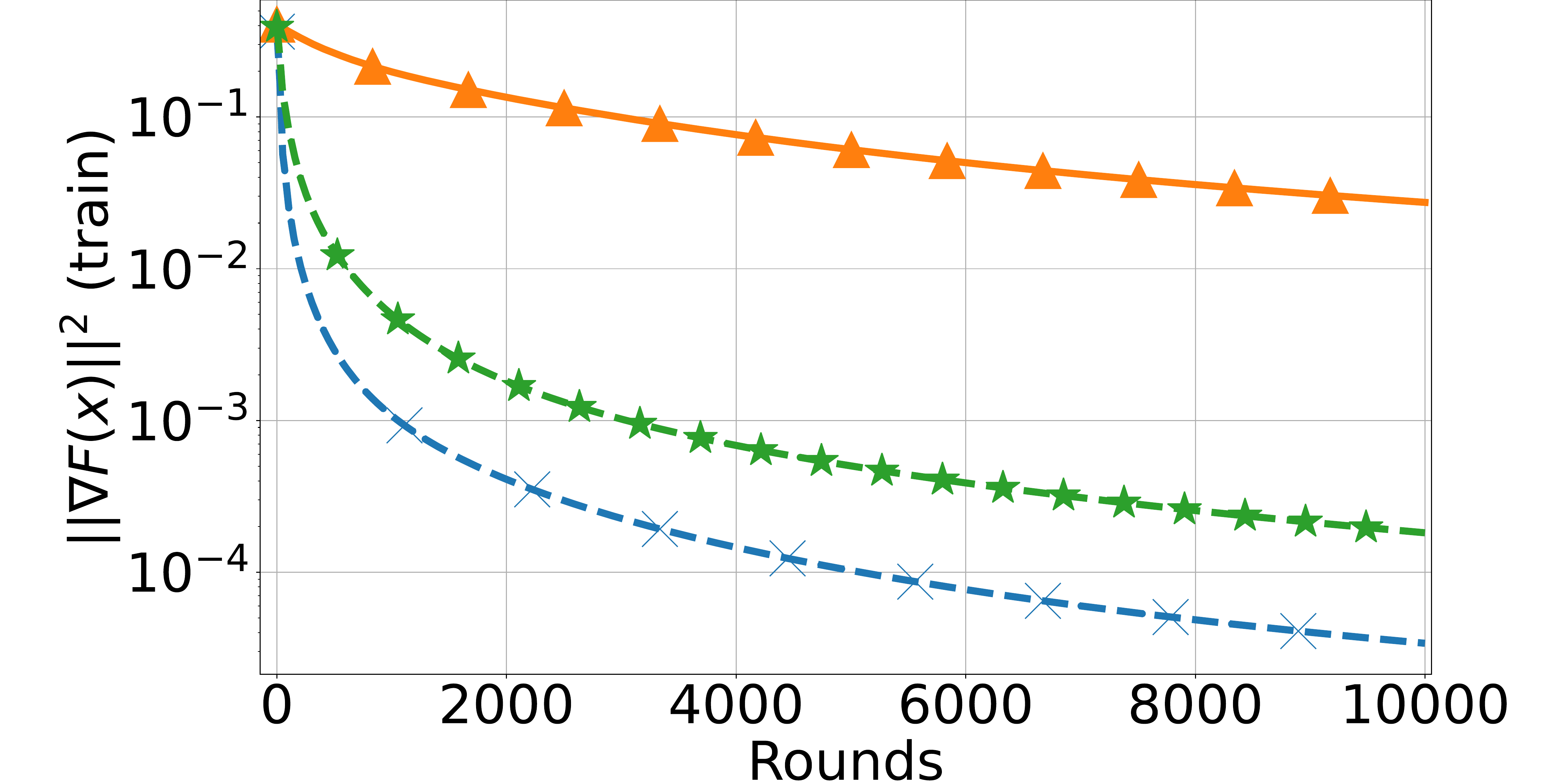}  \caption*{\hspace{20pt}$(a_3)$\, \texttt{mushroom}}
	\end{subfigure}
	
	\begin{subfigure}[ht]{0.32\textwidth}
		\includegraphics[width=\textwidth]{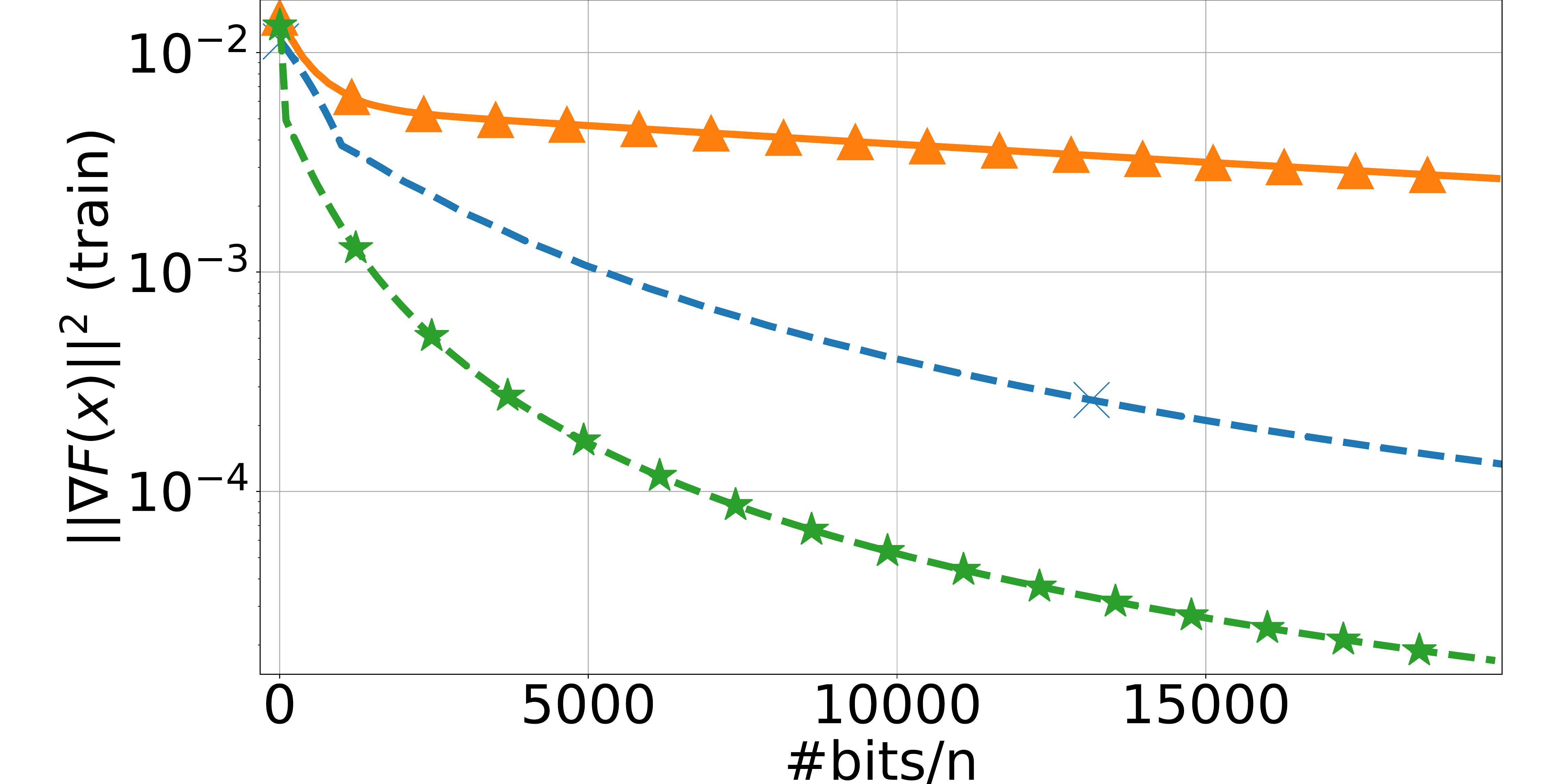}  \caption*{\hspace{20pt}$(b_1)$\, \texttt{phishing}}
	\end{subfigure}
	\begin{subfigure}[ht]{0.32\textwidth}
		\includegraphics[width=\textwidth]{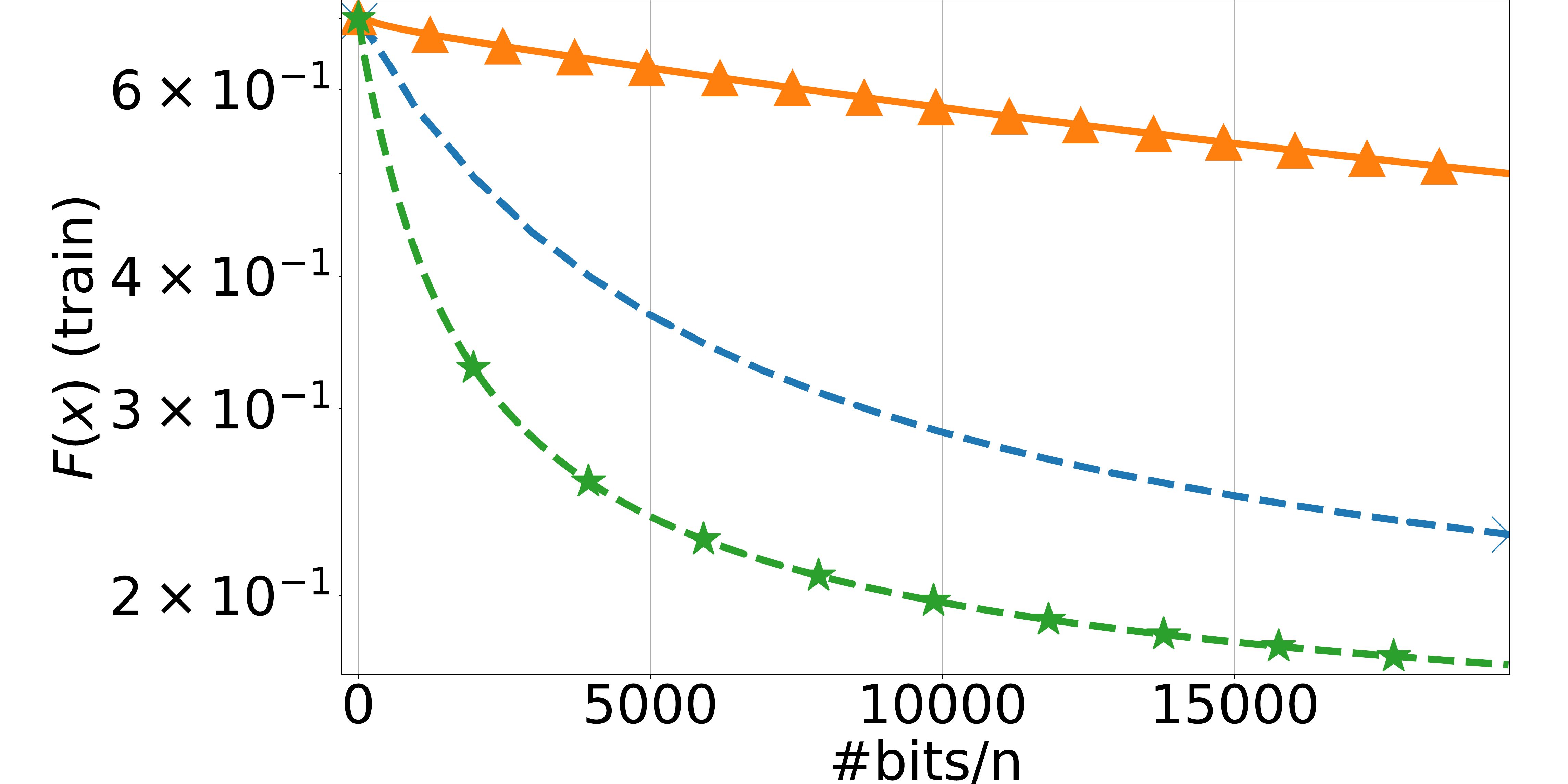}  \caption*{\hspace{20pt}$(b_2)$\, \texttt{phishing}}
	\end{subfigure}
	\begin{subfigure}[ht]{0.32\textwidth}
		\includegraphics[width=\textwidth]{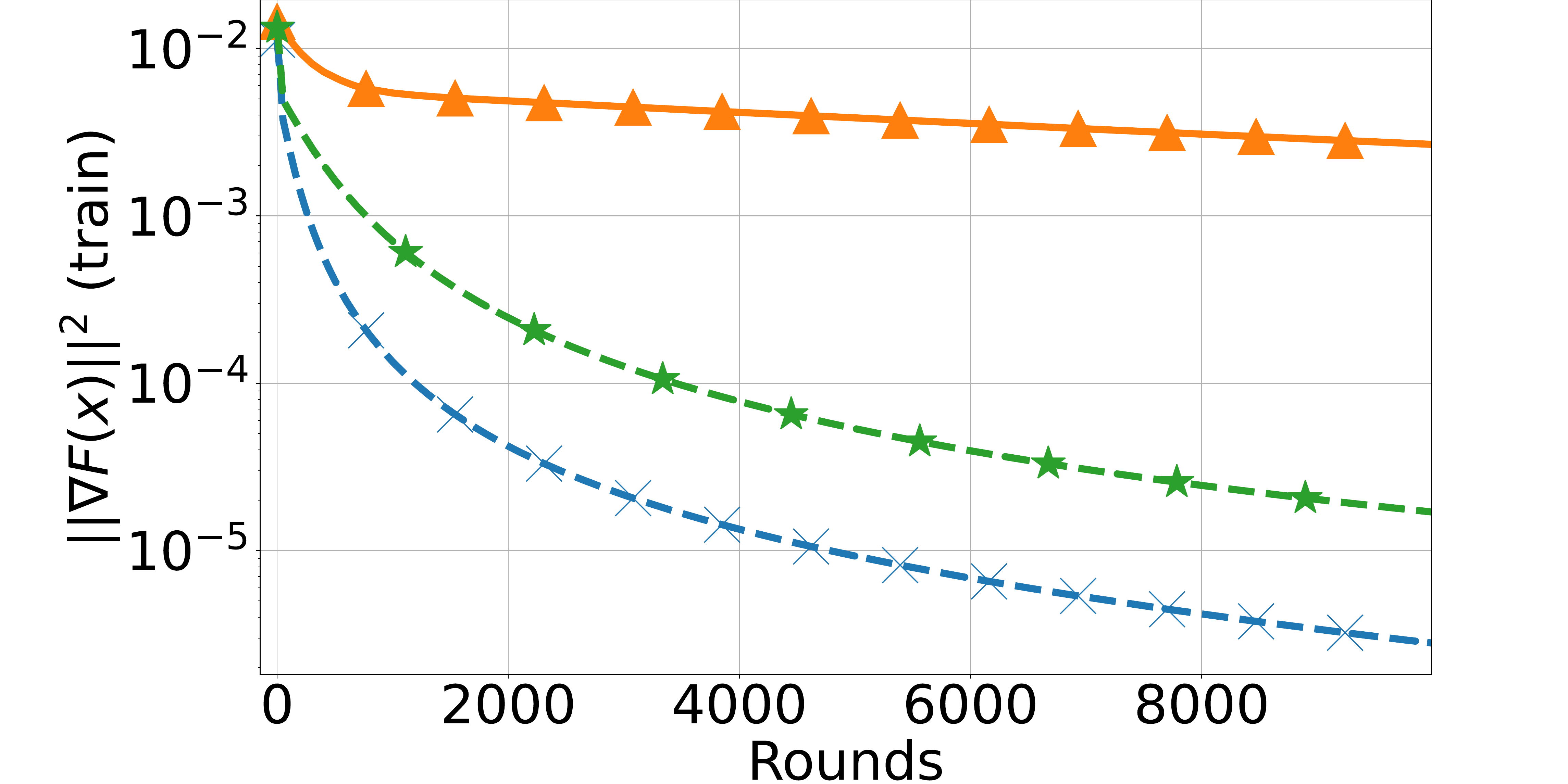}  \caption*{\hspace{20pt}$(b_3)$\, \texttt{phishing}}
	\end{subfigure}
	
	\begin{subfigure}[ht]{0.32\textwidth}
		\includegraphics[width=\textwidth]{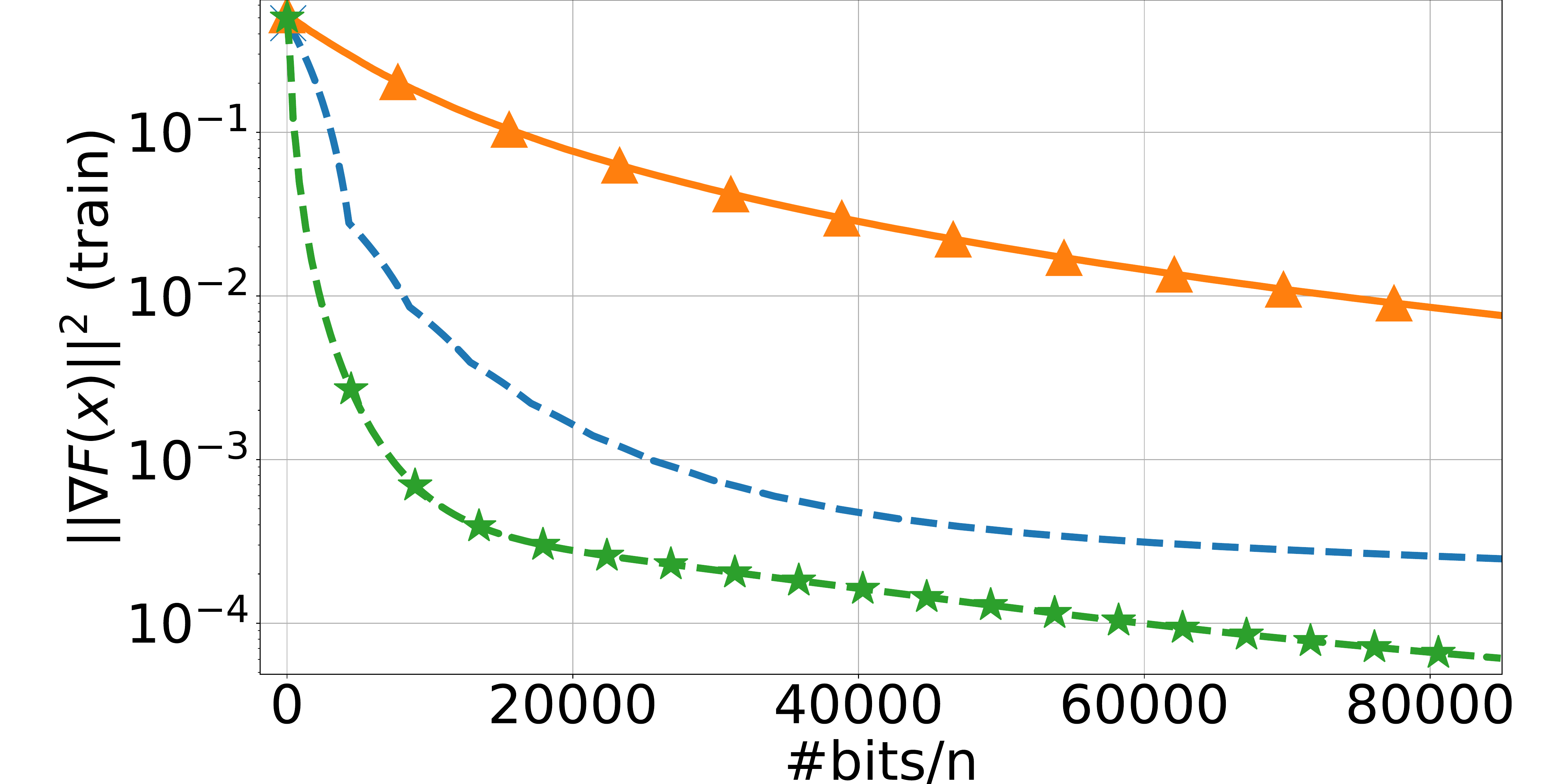}  \caption*{\hspace{20pt}$(c_1)$\, \texttt{w8a}}
	\end{subfigure}
	\begin{subfigure}[ht]{0.32\textwidth}
		\includegraphics[width=\textwidth]{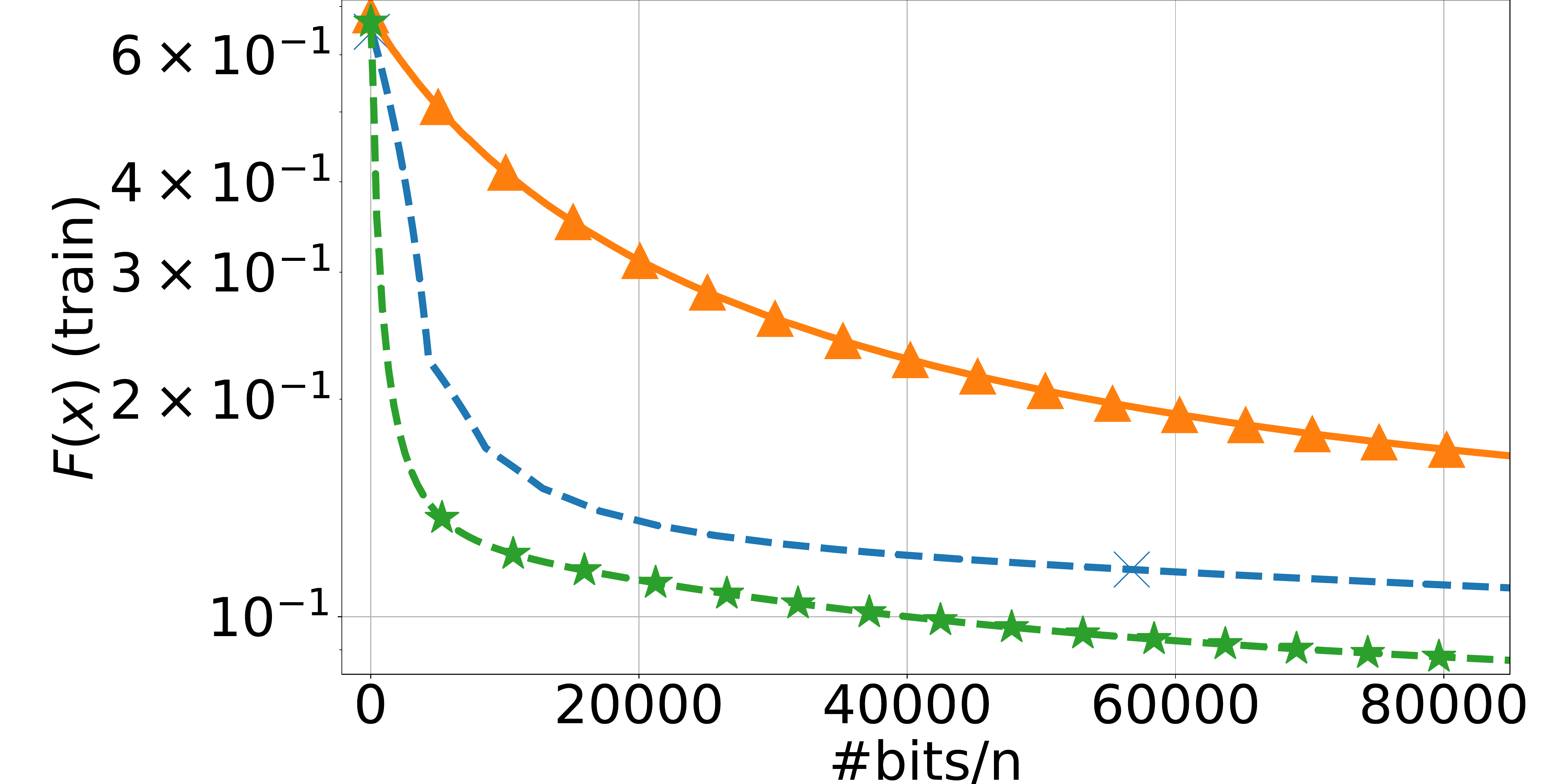}  \caption*{\hspace{20pt}$(c_2)$\, \texttt{w8a}}
	\end{subfigure}
	\begin{subfigure}[ht]{0.32\textwidth}
		\includegraphics[width=\textwidth]{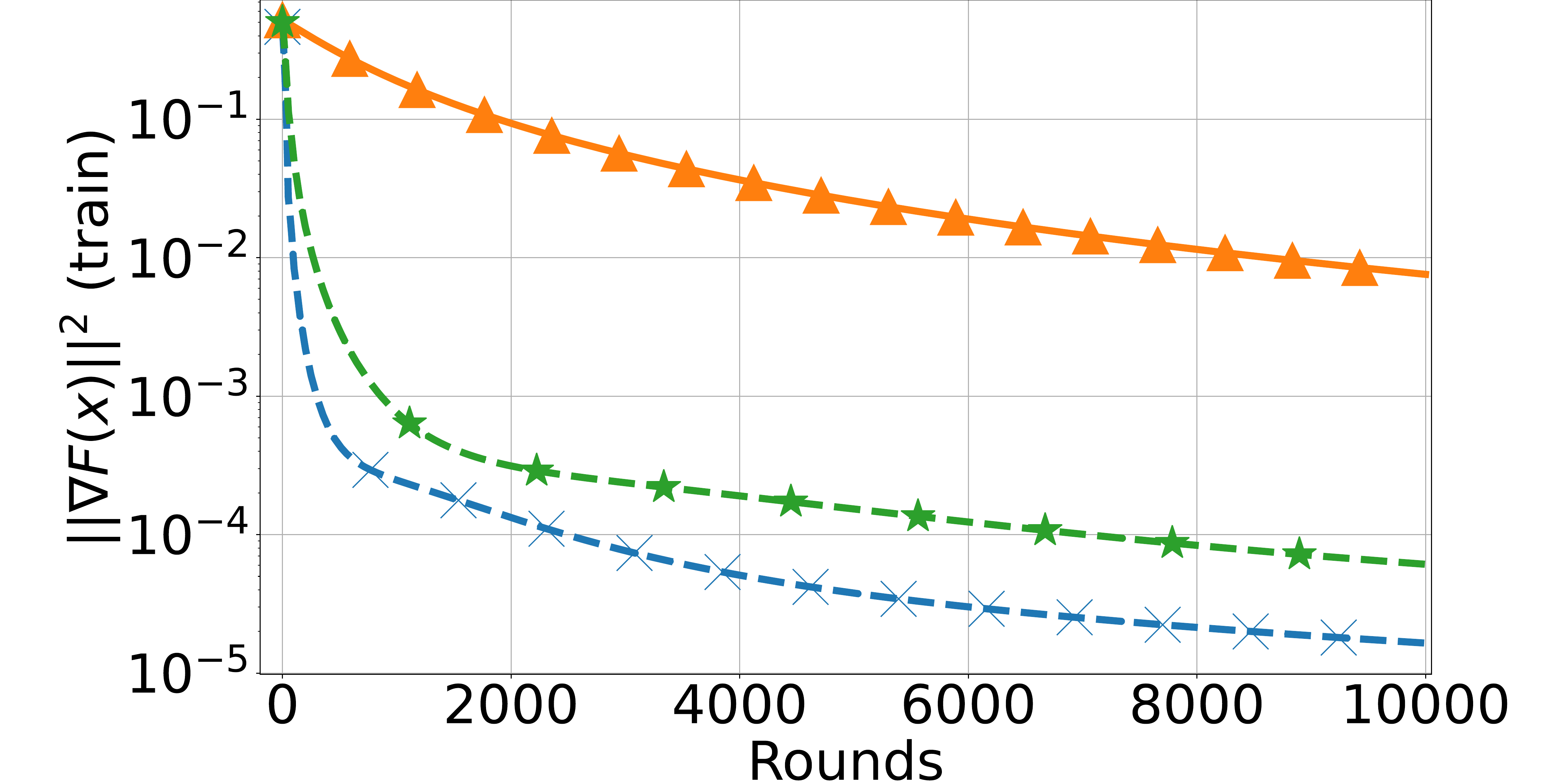}  \caption*{\hspace{20pt}$(c_3)$\, \texttt{w8a}}
	\end{subfigure}
	
	\begin{subfigure}[ht]{0.32\textwidth}
		\includegraphics[width=\textwidth]{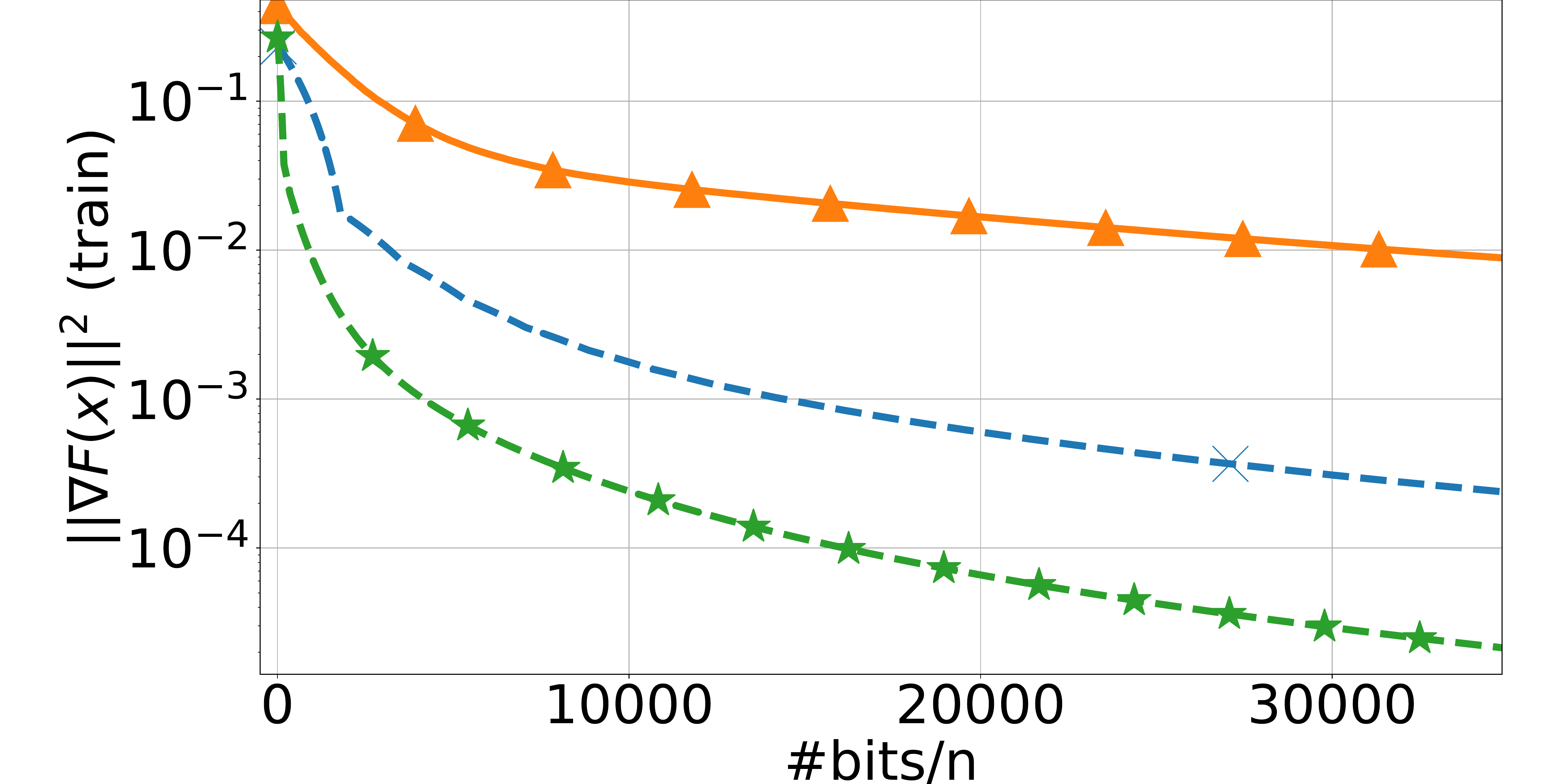}  \caption*{\hspace{20pt}$(d_1)$\, \texttt{a9a}}
	\end{subfigure}
	\begin{subfigure}[ht]{0.32\textwidth}
		\includegraphics[width=\textwidth]{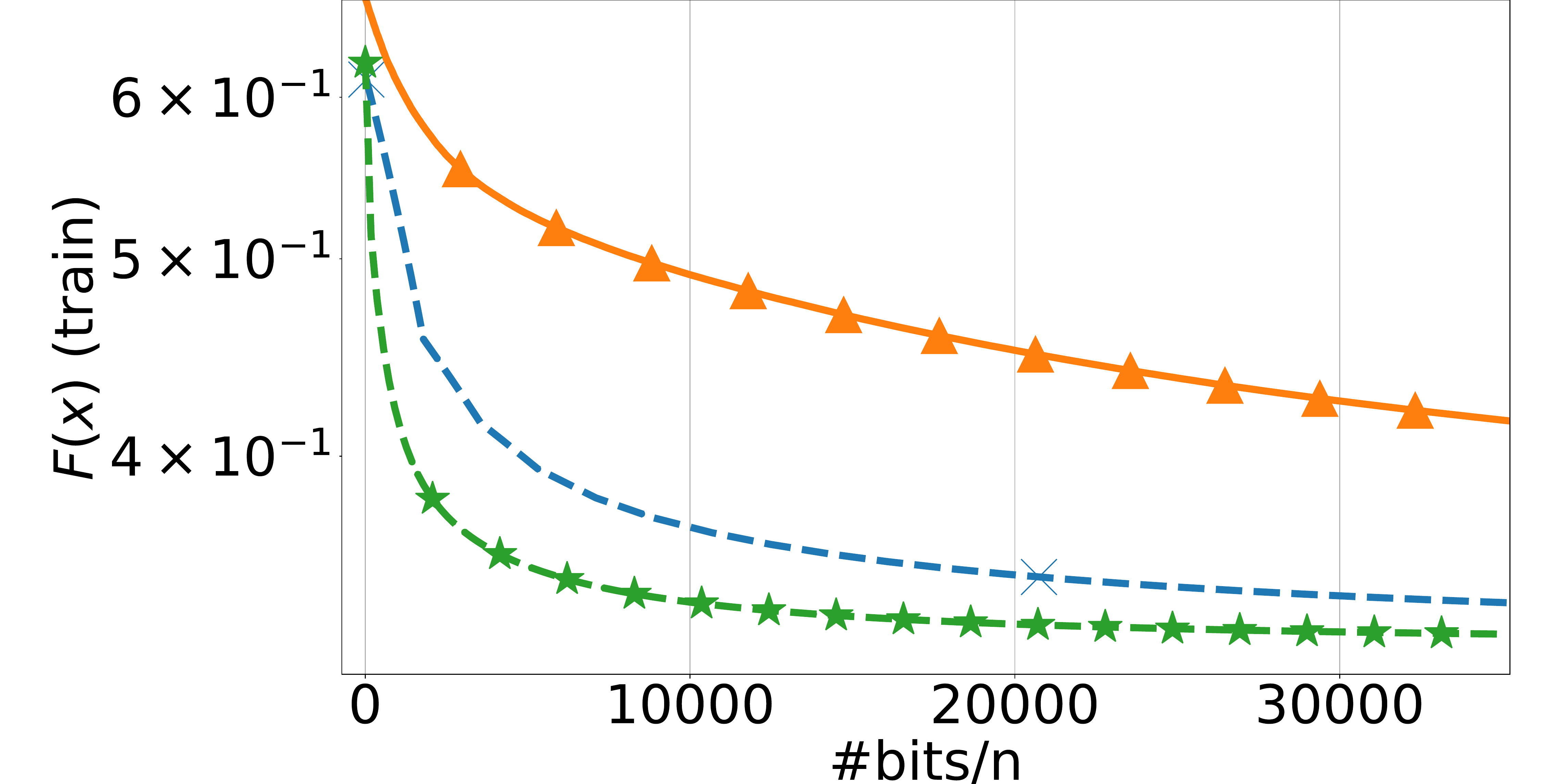}  \caption*{\hspace{20pt}$(d_2)$\, \texttt{a9a}}
	\end{subfigure}
	\begin{subfigure}[ht]{0.32\textwidth}
		\includegraphics[width=\textwidth]{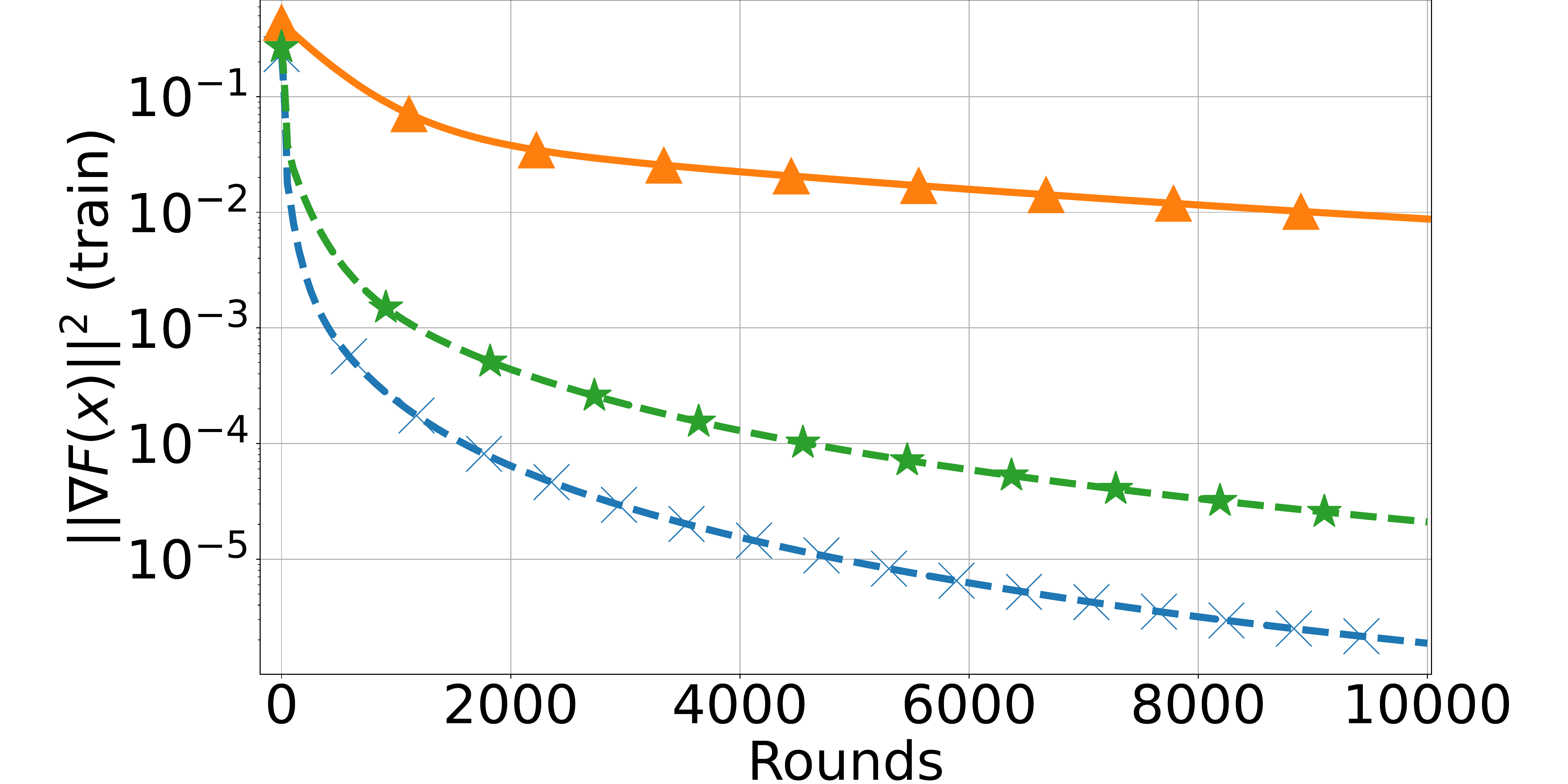}  \caption*{\hspace{20pt}$(d_3)$\, \texttt{a9a}}
	\end{subfigure}	
	
	\vspace{-3pt}
	\caption{\small{Results for logistic regression on several \texttt{LIBSVM} datasets with $100$ homogeneous clients. We use \texttt{Natural} compressor on the client-side. The step sizes we choose are the most aggressive step sizes according to the convex theory of different algorithms.}}
	\label{fig:training_logreg_convex_rr}
\end{figure*}

\begin{figure*}[!ht]
	\centering
	\captionsetup[sub]{font=scriptsize,labelfont={}}
	\begin{subfigure}[ht]{0.4\textwidth}
		\includegraphics[width=\textwidth]{./imgs/logregexp3/legend.pdf}  \caption*{}
	\end{subfigure}
	
	\begin{subfigure}[ht]{0.32\textwidth}
		\includegraphics[width=\textwidth]{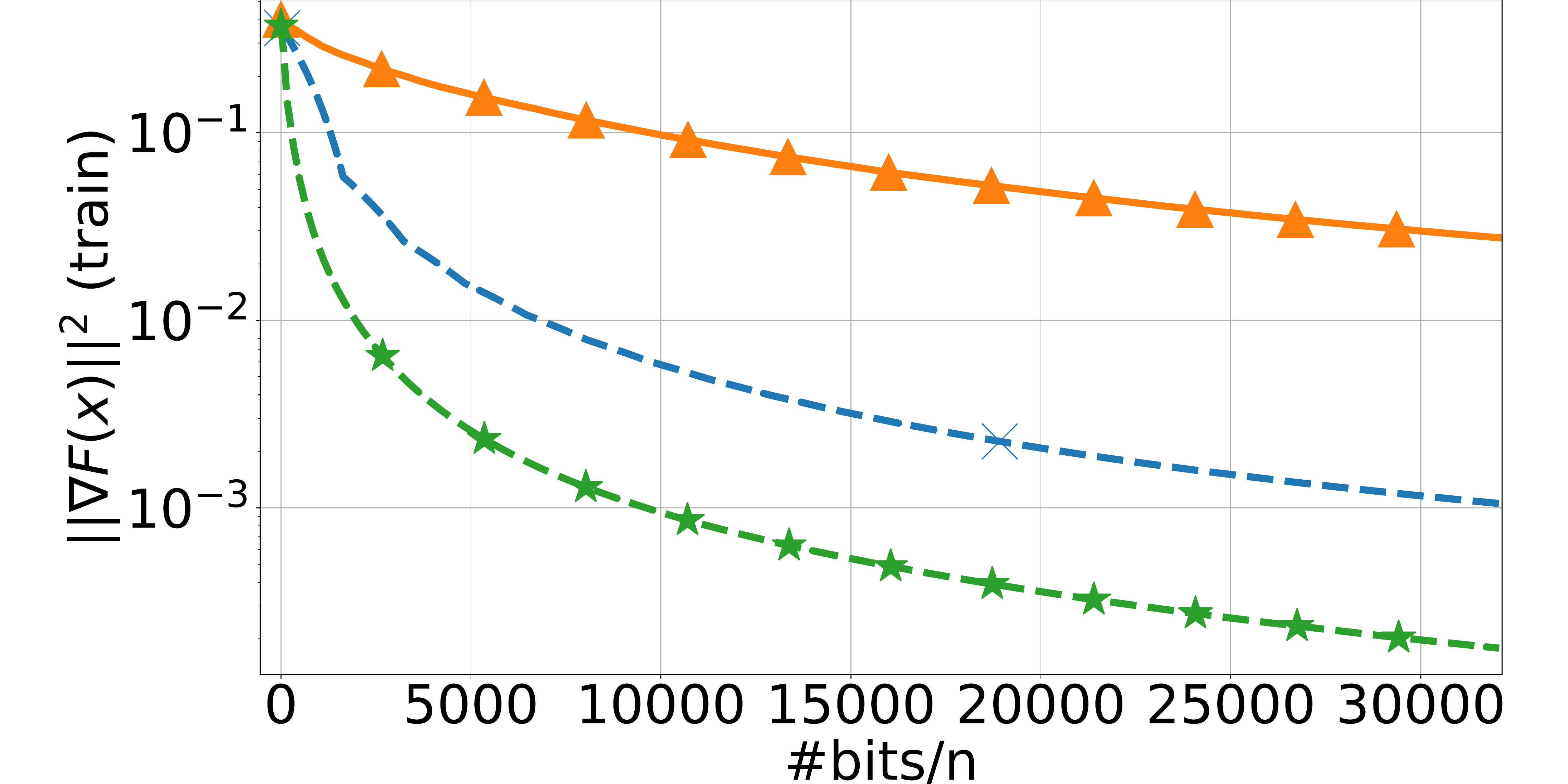}  \caption*{\hspace{20pt}$(a_1)$\, \texttt{mushroom}}
	\end{subfigure}
	\begin{subfigure}[ht]{0.32\textwidth}
		\includegraphics[width=\textwidth]{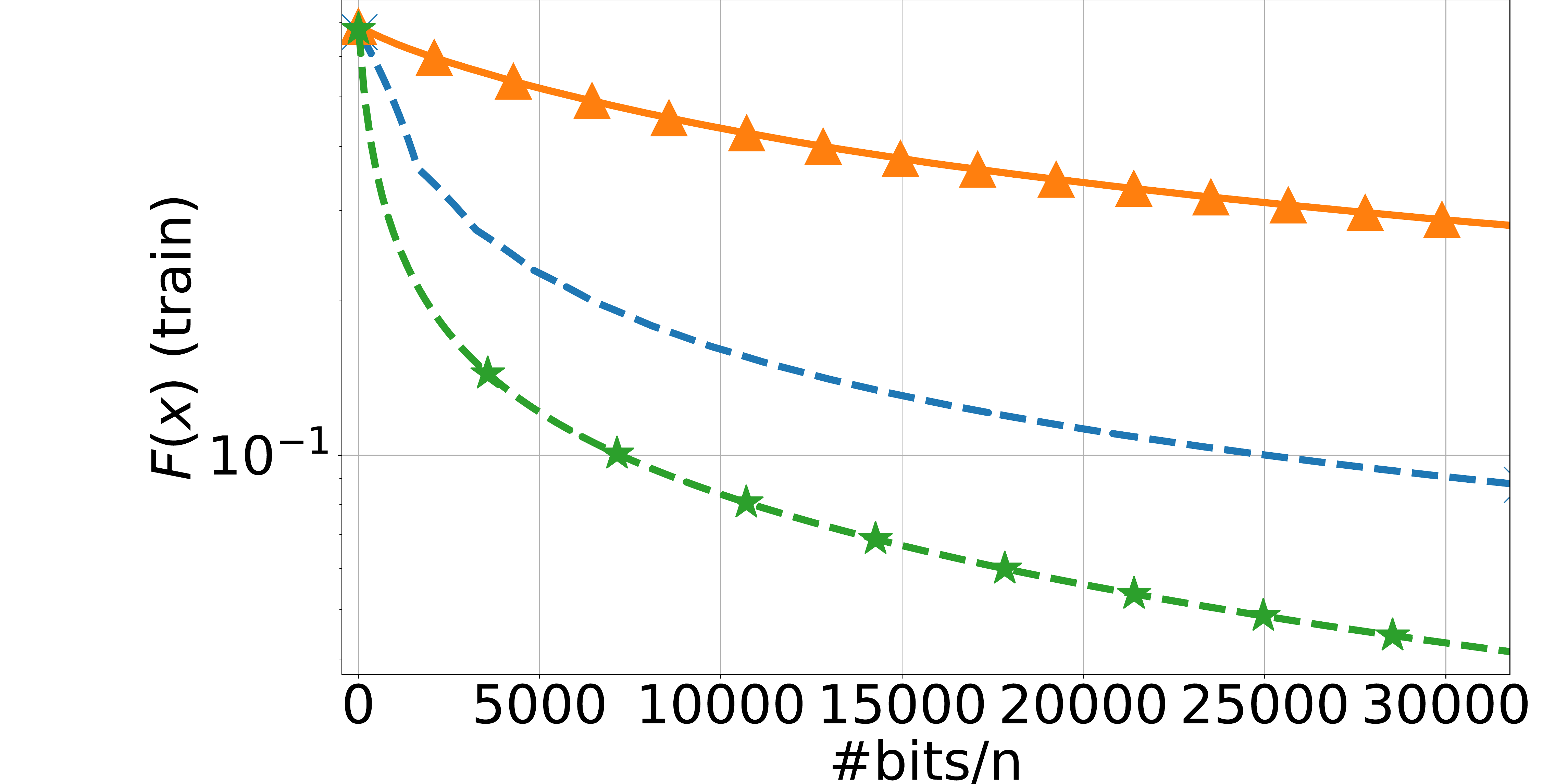}  \caption*{\hspace{20pt}$(a_2)$\, \texttt{mushroom}}
	\end{subfigure}
	\begin{subfigure}[ht]{0.32\textwidth}
		\includegraphics[width=\textwidth]{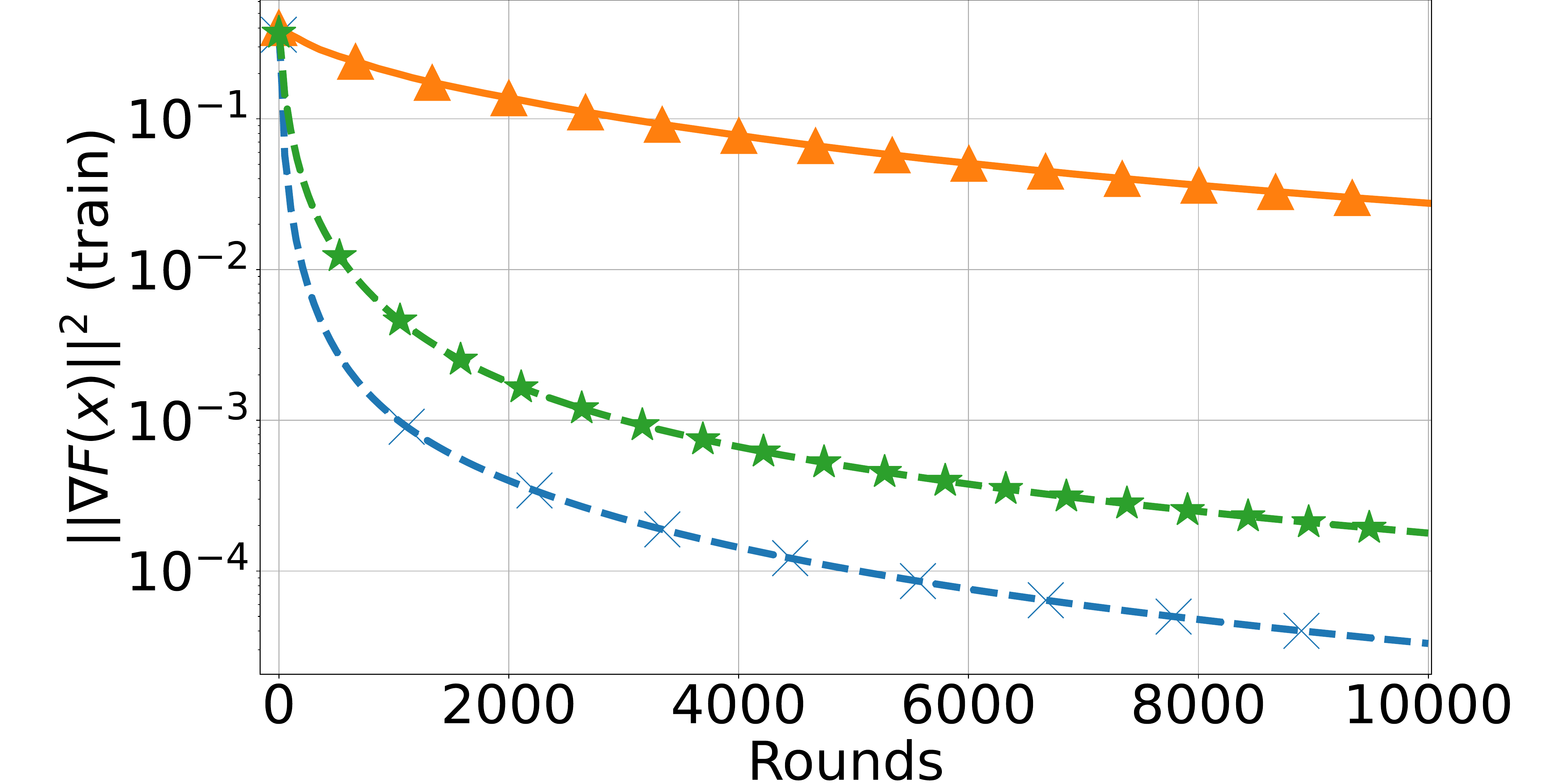}  \caption*{\hspace{20pt}$(a_3)$\, \texttt{mushroom}}
	\end{subfigure}
	
	\begin{subfigure}[ht]{0.32\textwidth}
		\includegraphics[width=\textwidth]{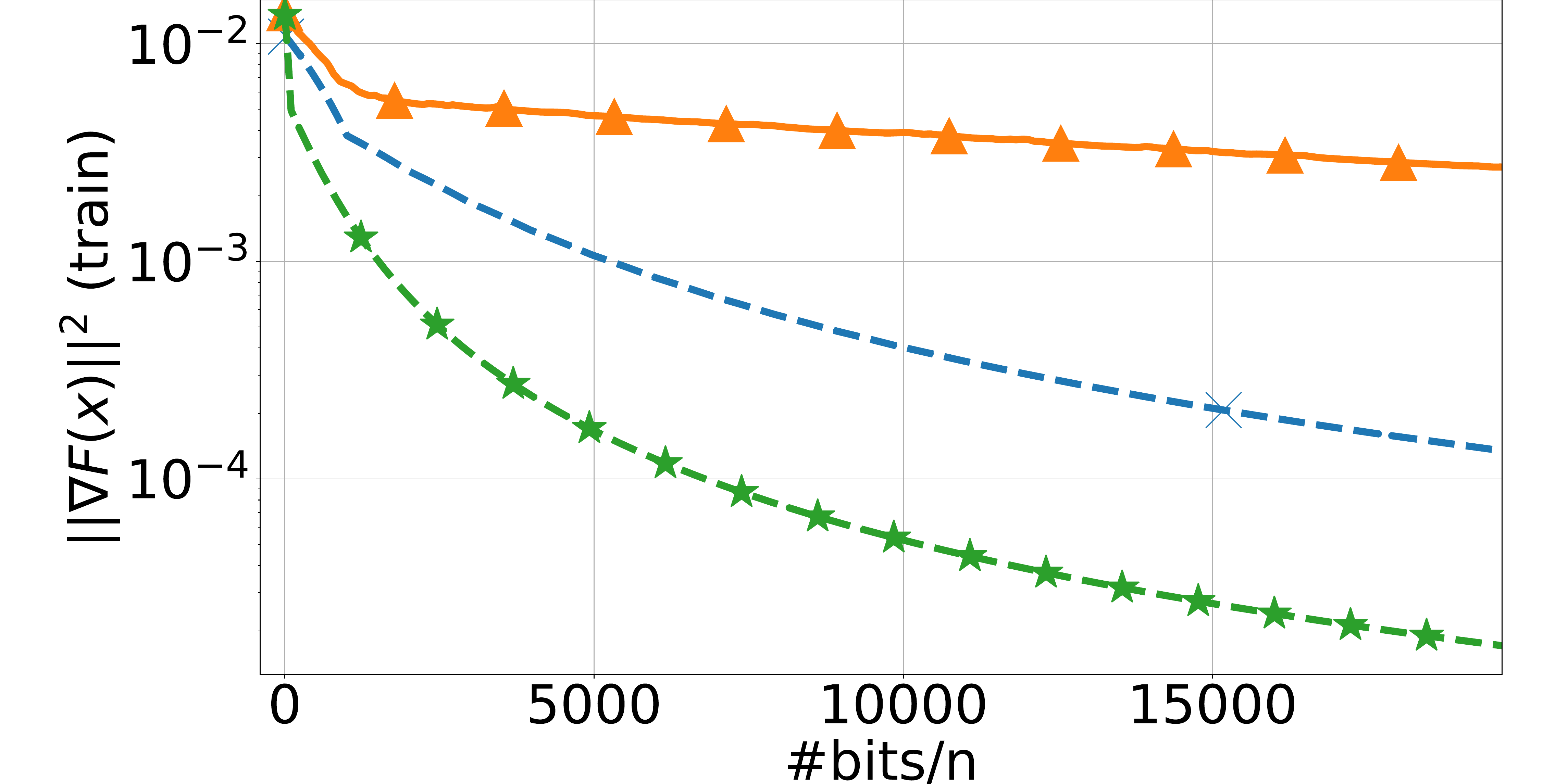}  \caption*{\hspace{20pt}$(b_1)$\, \texttt{phishing}}
	\end{subfigure}
	\begin{subfigure}[ht]{0.32\textwidth}
		\includegraphics[width=\textwidth]{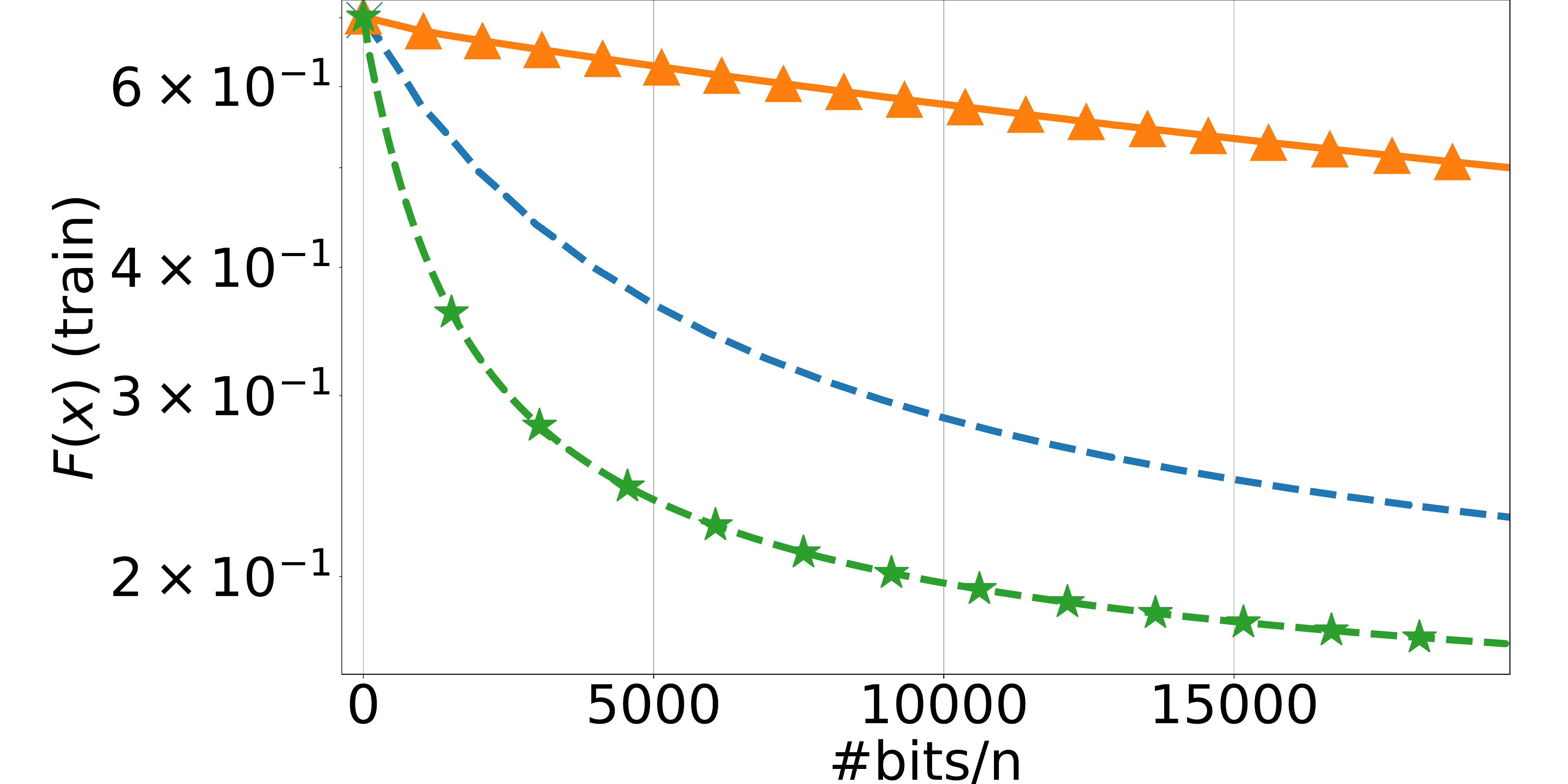}  \caption*{\hspace{20pt}$(b_2)$\, \texttt{phishing}}
	\end{subfigure}
	\begin{subfigure}[ht]{0.32\textwidth}
		\includegraphics[width=\textwidth]{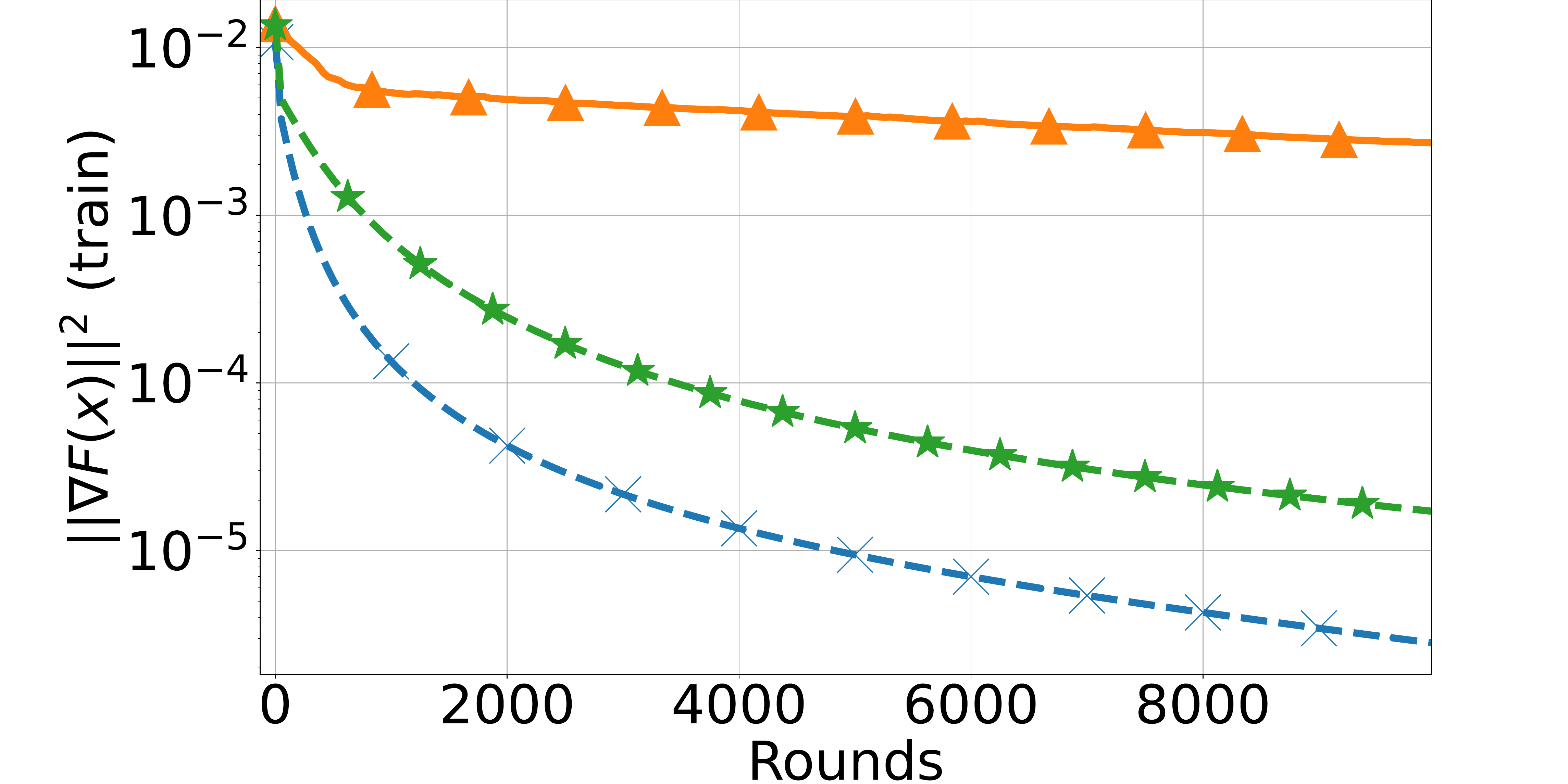}  \caption*{\hspace{20pt}$(b_3)$\, \texttt{phishing}}
	\end{subfigure}
	
	\begin{subfigure}[ht]{0.32\textwidth}
		\includegraphics[width=\textwidth]{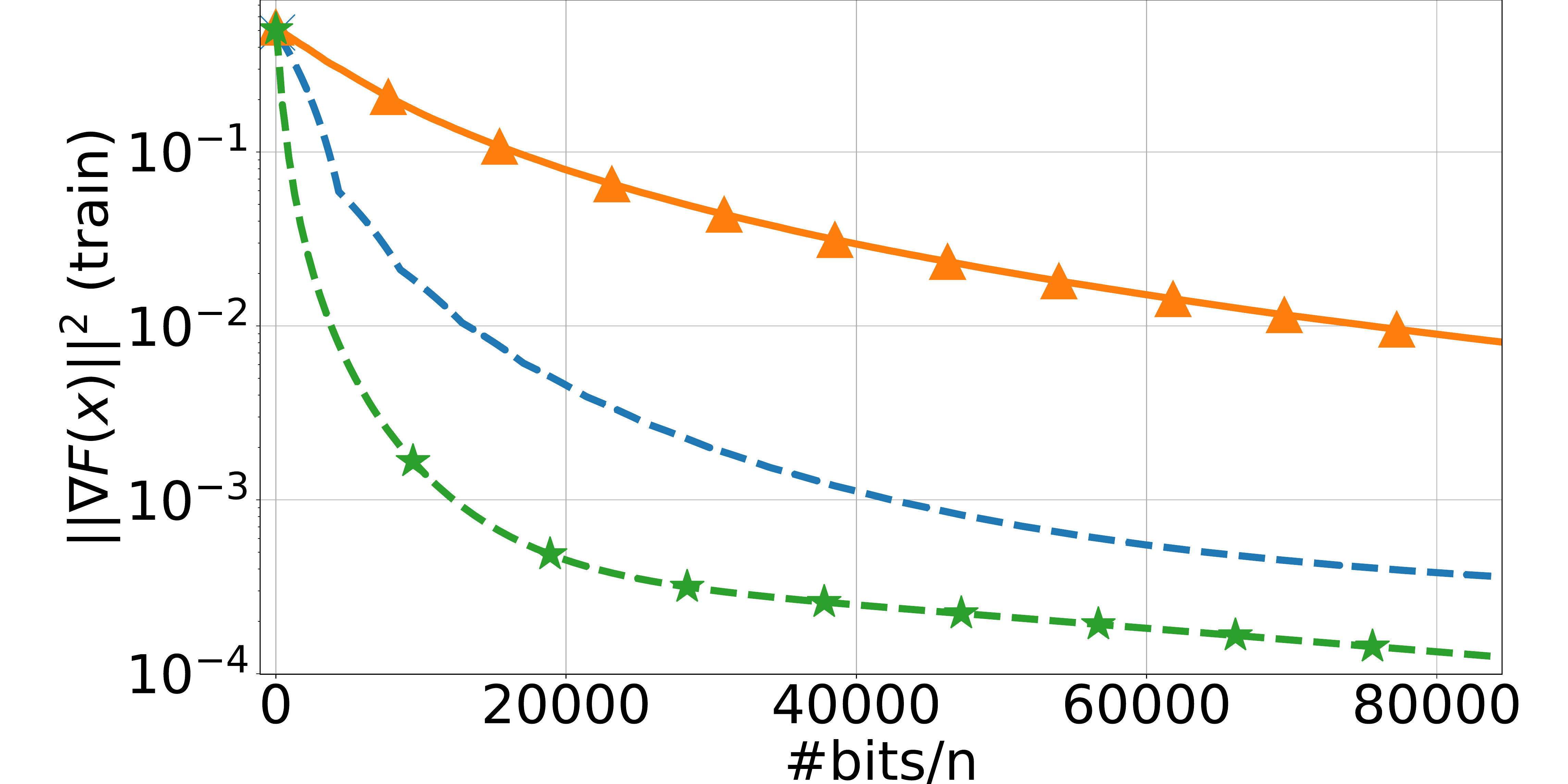}  \caption*{\hspace{20pt}$(c_1)$\, \texttt{w8a}}
	\end{subfigure}
	\begin{subfigure}[ht]{0.32\textwidth}
		\includegraphics[width=\textwidth]{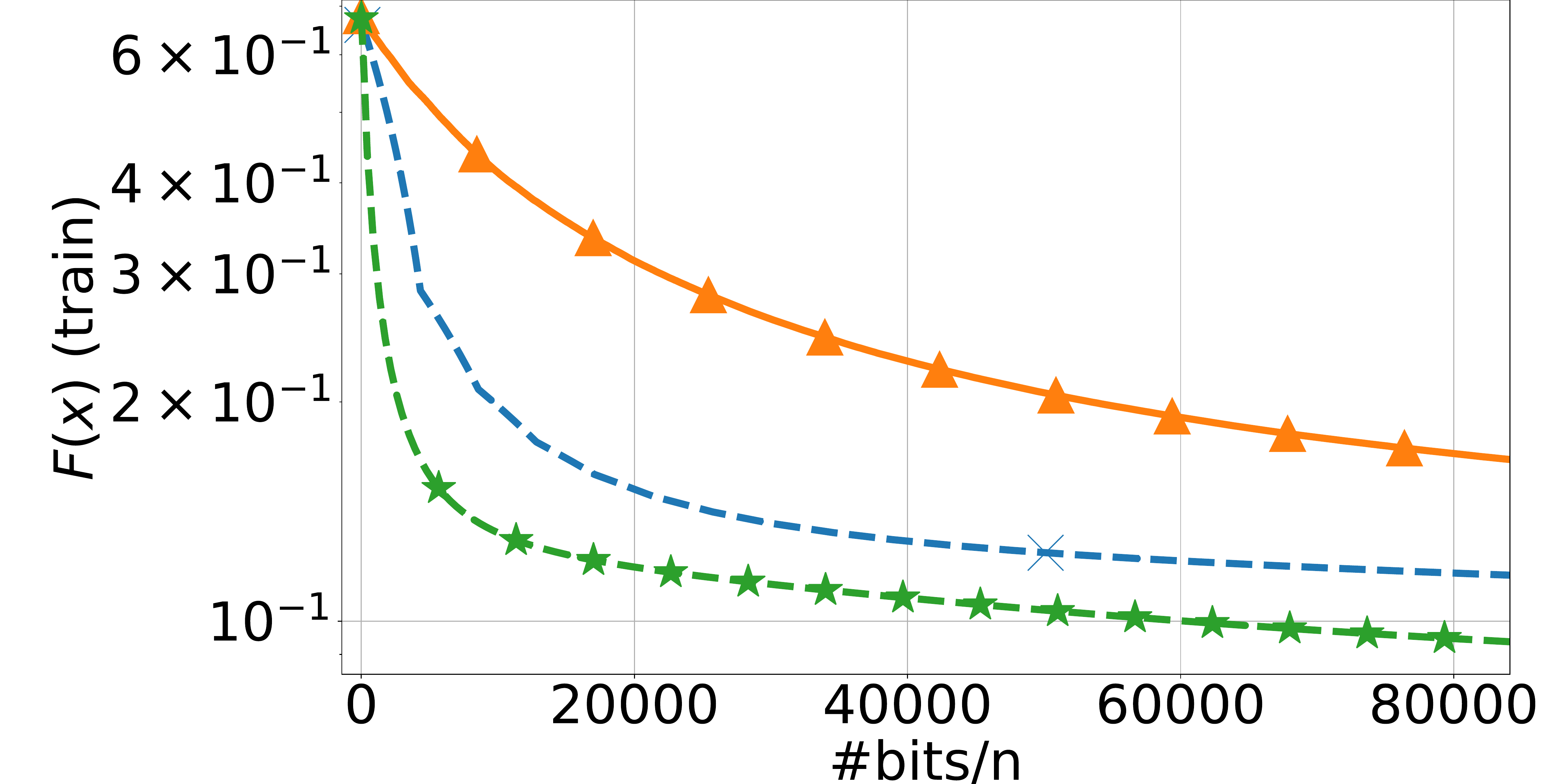}  \caption*{\hspace{20pt}$(c_2)$\, \texttt{w8a}}
	\end{subfigure}
	\begin{subfigure}[ht]{0.32\textwidth}
		\includegraphics[width=\textwidth]{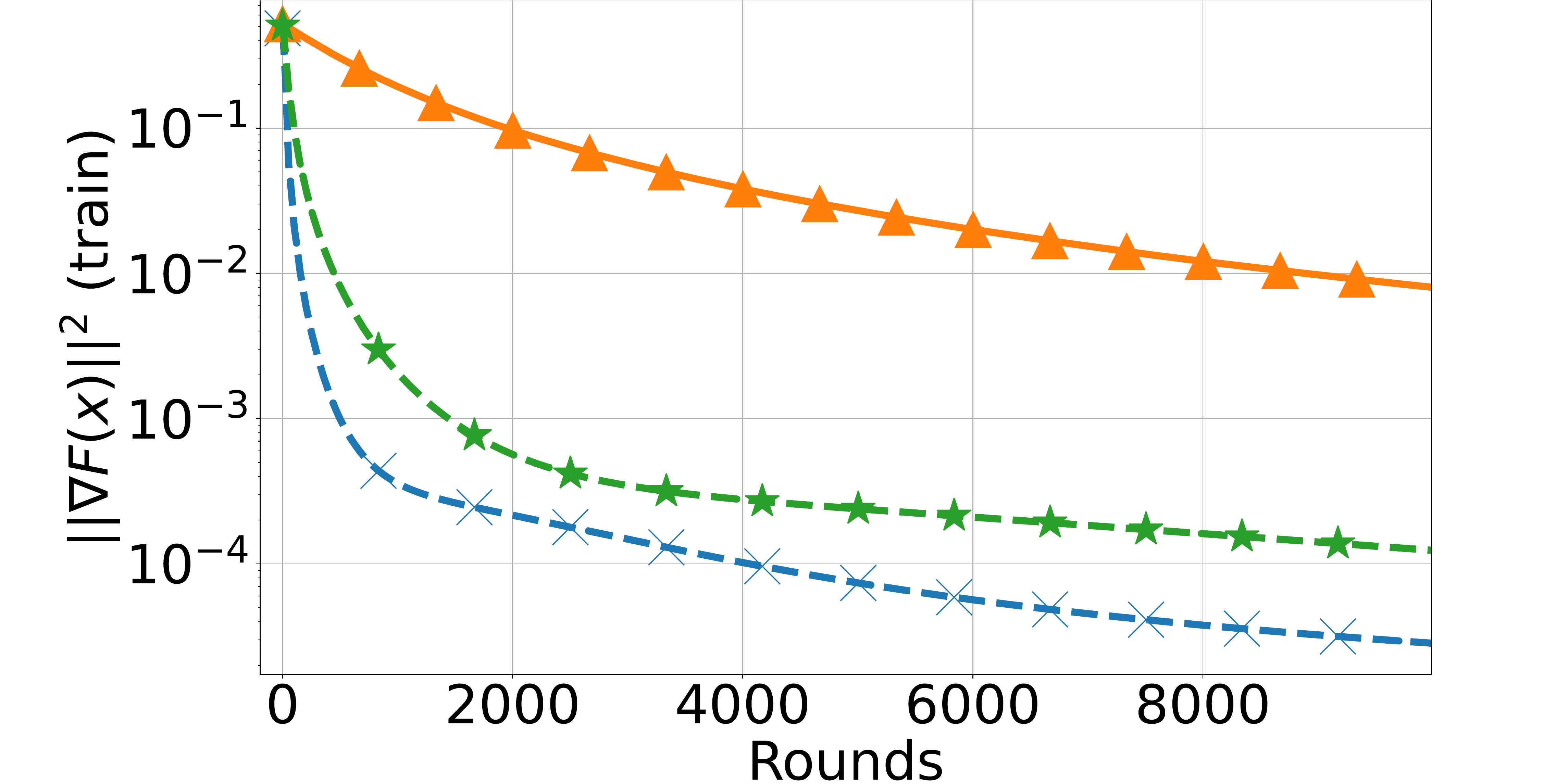}  \caption*{\hspace{20pt}$(c_3)$\, \texttt{w8a}}
	\end{subfigure}
	
	\begin{subfigure}[ht]{0.32\textwidth}
		\includegraphics[width=\textwidth]{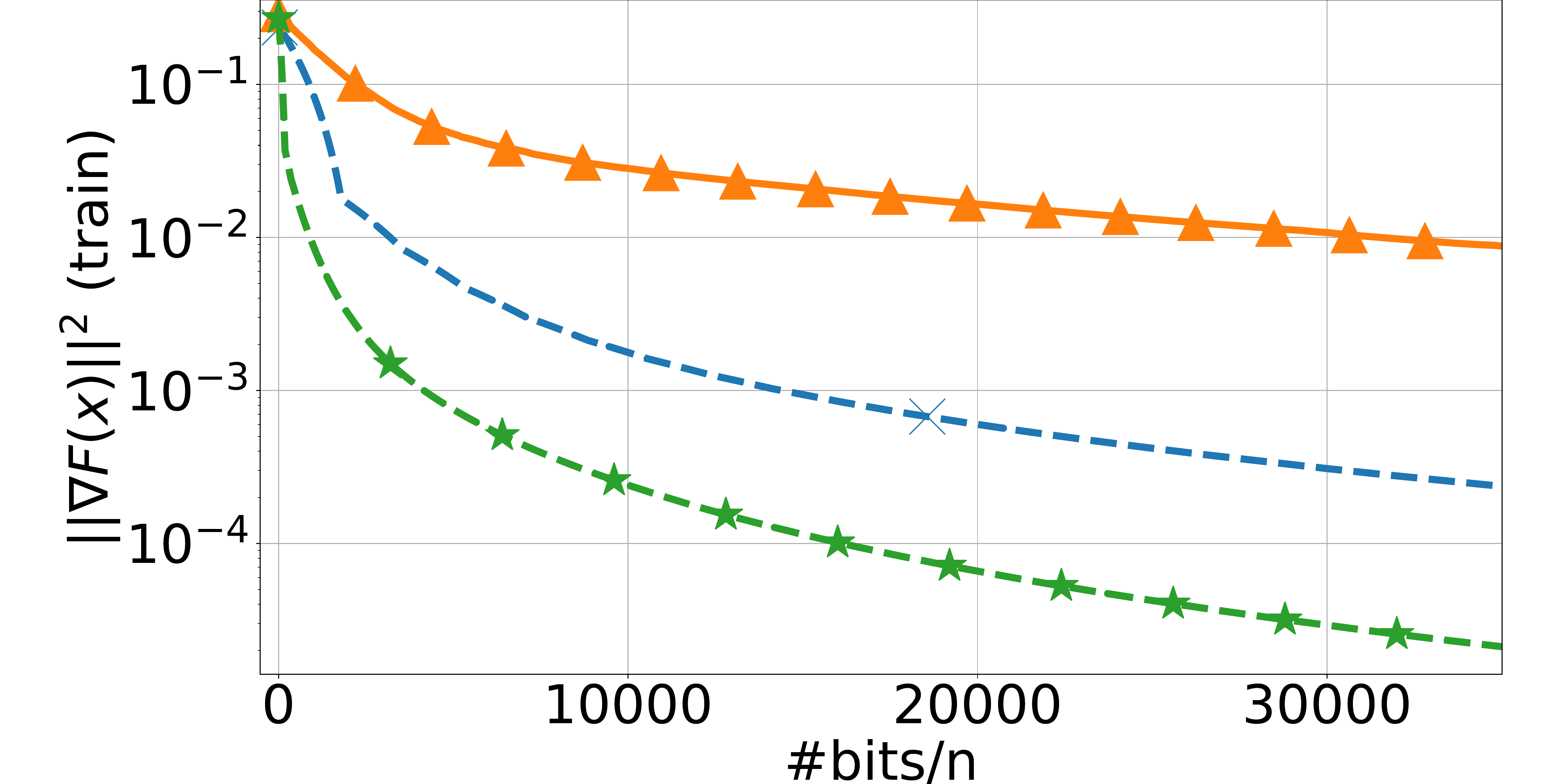}  \caption*{\hspace{20pt}$(d_1)$\, \texttt{a9a}}
	\end{subfigure}
	\begin{subfigure}[ht]{0.32\textwidth}
		\includegraphics[width=\textwidth]{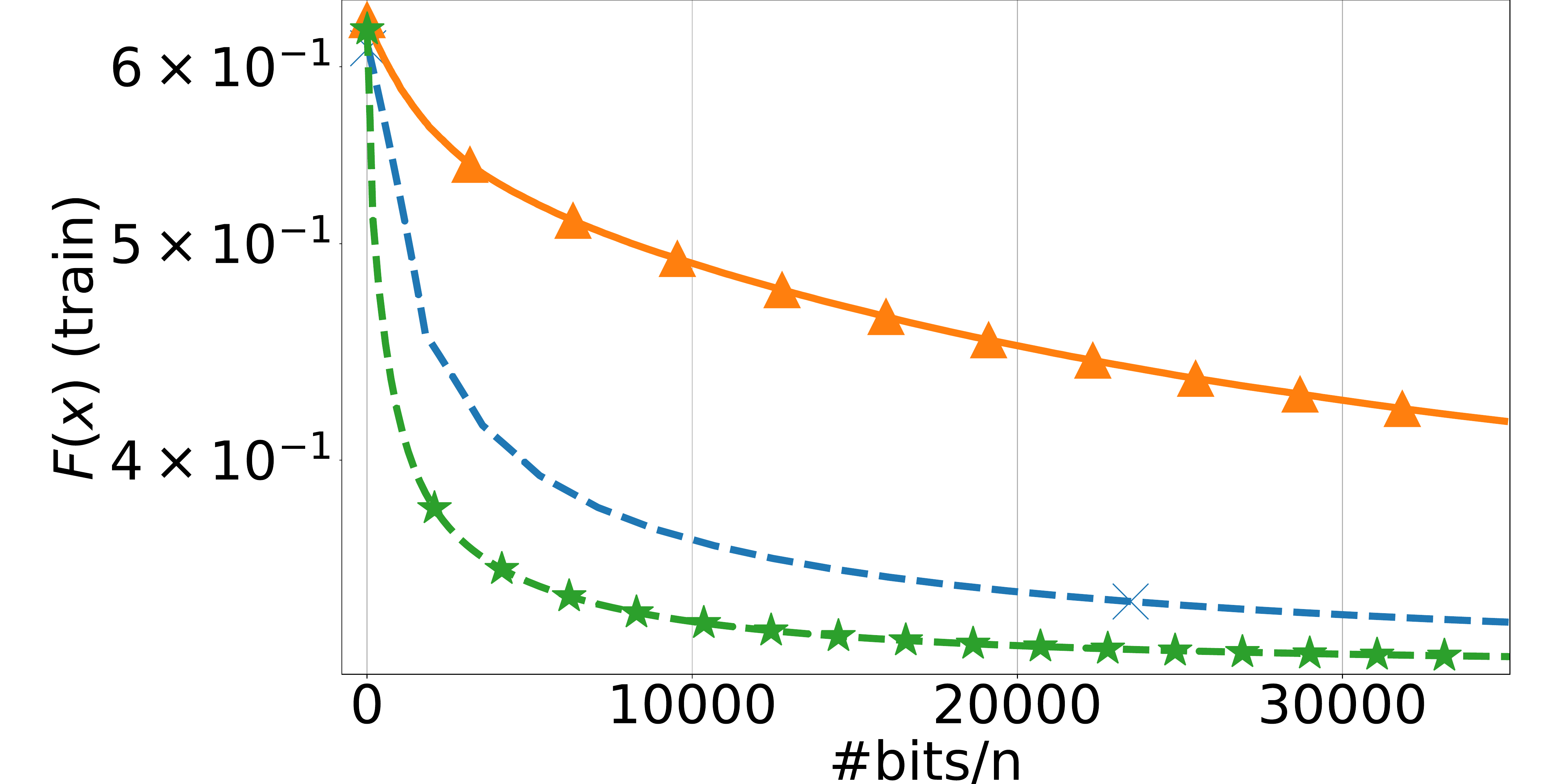}  \caption*{\hspace{20pt}$(d_2)$\, \texttt{a9a}}
	\end{subfigure}
	\begin{subfigure}[ht]{0.32\textwidth}
		\includegraphics[width=\textwidth]{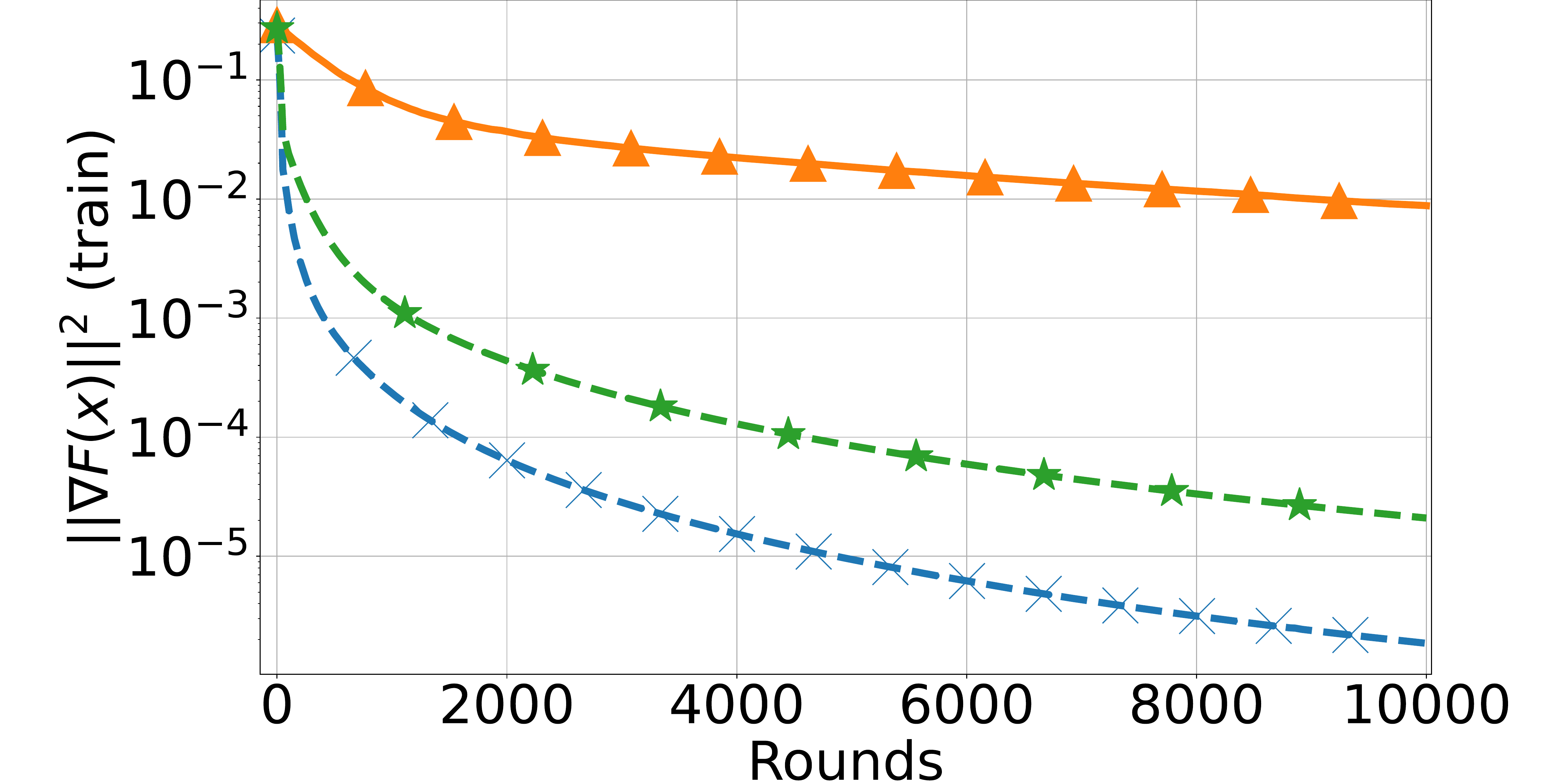}  \caption*{\hspace{20pt}$(d_3)$\, \texttt{a9a}}
	\end{subfigure}	
	
	\vspace{-3pt}
	\caption{\small{Results for logistic regression on several \texttt{LIBSVM} datasets with $100$ heterogeneous clients. We use \texttt{Natural} compressor on the client-side. The step sizes we choose are the most aggressive step sizes according to the convex convergence results of different algorithms.}}
	\label{fig:training_logreg_convex_hetero}
\end{figure*}

\paragraph{Computing environment} The target machine in which we have launched experiments has Intel(R) Xeon(R) Gold 6246 CPU 3.30GHz with 48 logical cores with hyper-threading and also equipped with two GeForce RTX 3090 GPU cards where each GPU has 5248 scalar FPU cores working at 1.71GHz. Each GeForce RTX 3090 GPU card has 24 GBytes of GPU DRAM memory. We conduct our experiments using the simulation framework \texttt{FL\_PYTorch}~\cite{burlachenko2021fl_pytorch} and implement \algname{EF21-PP}, \algname{MARINA-PP}, \algname{COFIG}, and \algname{FRECON}.

\paragraph{Description of datasets} In the following experiments, we analyze algorithms' performance by regularized logistic regression on several LIBSVM~~\cite{chang2011libsvm} datasets: \dataset{a9a}, \dataset{w8a}, \dataset{phishing}, and \dataset{mushroom}. The numbers of optimization variables for these datasets are as follow: $125$ for \dataset{a9a},  $302$ for \dataset{w8a},  $70$ for \dataset{phishing}, and $114$ for \dataset{mushroom}. To simulate the FL settings with different data distribution, we consider two strategies: \textbf{(i) uniform shuffling}, where the considered datasets are randomly shuffled divided into $N=100$ parts ($N=100$ clients); and \textbf{(ii) class-based sorting}, where the datasets are sorted based on the label and divided into $N=100$ parts (that is, 50 clients hold the data with label $1$ and other 50 clients hold the data with label $0$).

\begin{table}[!b]
	\centering
	\caption{Binary classification experimental reference table}
	\label{tbl:log_reg_experiments}
	\begin{tabular}{|c||l|l|l|}
		\hline
		Experiment & Shuffling strategy & Objective type for log. regression  & Results \\
		\hline\hline
		\#1 & uniform shuffling & non-convex ($\alpha=0.1$) & Fig. \ref{fig:training_logreg_non_convex_rr} \\
		\hline
		\#2 & class-based sorting & non-convex ($\alpha=0.1$) & Fig. \ref{fig:training_logreg_non_convex_hetero}  \\
		\hline
		\#3 & uniform shuffling & convex ($\alpha=0$) & Fig. \ref{fig:training_logreg_convex_rr}  \\
		\hline
		\#4 & class-based sorting & convex ($\alpha=0$) & Fig. \ref{fig:training_logreg_convex_hetero}  \\
		\hline
	\end{tabular}
\end{table}

\paragraph{Description of experiments} For non-convex loss experiment we set $\alpha=0.1$ and we have compared \cofig, \frecon with \algname{PP-MARINA}~\cite{gorbunov2021marina}, \algname{DIANA}~\cite{mishchenko2019distributed,horvath2019stochastic}, \algname{EF21-PP}~\cite{fatkhullin2021ef21}. In these experiments we have selected maximum allowable step size according to the theory of these methods in non-convex regimes. For convex loss experiments set $\alpha=0$ in and use convex theory of \algname{DIANA} and \cofig to obtain theoretical step sizes. For \algname{EF21-PP} in the (convex) logistic regression, we use its nonconvex theoretical step size since it does not have convergence analysis under the convex setting. For \algname{EF21-PP}, \cofig, \frecon, and \algname{PP-MARINA} in each round we sampled $10$ clients uniformly from $100$ total clients. For \algname{DIANA}, the number of clients in each round is $100$ since it requires full participation. All algorithms run for $10\,000$ rounds. The amount of communication bits includes the amount of bits from $N$ clients to the master. We use \texttt{Natural} compressor~\cite{horvath2019natural} as a compression mechanism used for the communication from clients to server. In these logistic regression (and its variants) experiments, we let each client to compute the local full gradient ($\sigma=0$). \Cref{tbl:log_reg_experiments} summarizes the experiments settings and their corresponding results.

\paragraph{Experiment results} The results for Experiment \#1 are presented in \Cref{fig:training_logreg_non_convex_rr}. From the results, \cofig and \frecon outperform \algname{EF21-PP} and \algname{DIANA}, and are competitive against \texttt{MARINA-PP} with respect to the communication bits. From these experiments, we claim that \cofig, as a generalization of \algname{DIANA}, behaves better than original \algname{DIANA}. Although \frecon behaves slightly worse than \texttt{PP-MARINA}, but it's important to mention that \algname{PP-MARINA} need to communicate with all clients with uncompressed information periodically, and maybe impossible in practice. Results of Experiment \#2 are presented in \Cref{fig:training_logreg_non_convex_hetero}. Although this setting is harder than the previous Experiment \#1, all observations and claims made for Experiment \#1 almost hold for Experiment \#2. Experiments \#3 and \#4 demonstrate that \cofig in convex setting is practically superior than \algname{EF21-PP}, \algname{DIANA} in all considered LIBSVM datasets. (See \Cref{fig:training_logreg_convex_rr} and \Cref{fig:training_logreg_convex_hetero}). The class-based sorting setting for binary classification if the number of clients is even and each client has the same number of datapoint will essentially mean that each client has training samples corresponding only to one class. Without collaboration with other clients and using only local data points the training process organized through optimization of $f_i(x)$ is meaningless for any client $i$, and we want to highlight that in this extreme case, collaboration between clients is required.

\subsection{Image classification with neural nets}

\begin{figure*}[ht]
	\centering
	\captionsetup[sub]{font=scriptsize,labelfont={}}	
	\begin{subfigure}[ht]{1.0\textwidth}
		\includegraphics[width=\textwidth]{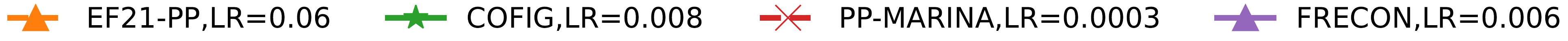}  \caption*{}
	\end{subfigure}
	
	\begin{subfigure}[ht]{0.49\textwidth}
		\includegraphics[width=\textwidth]{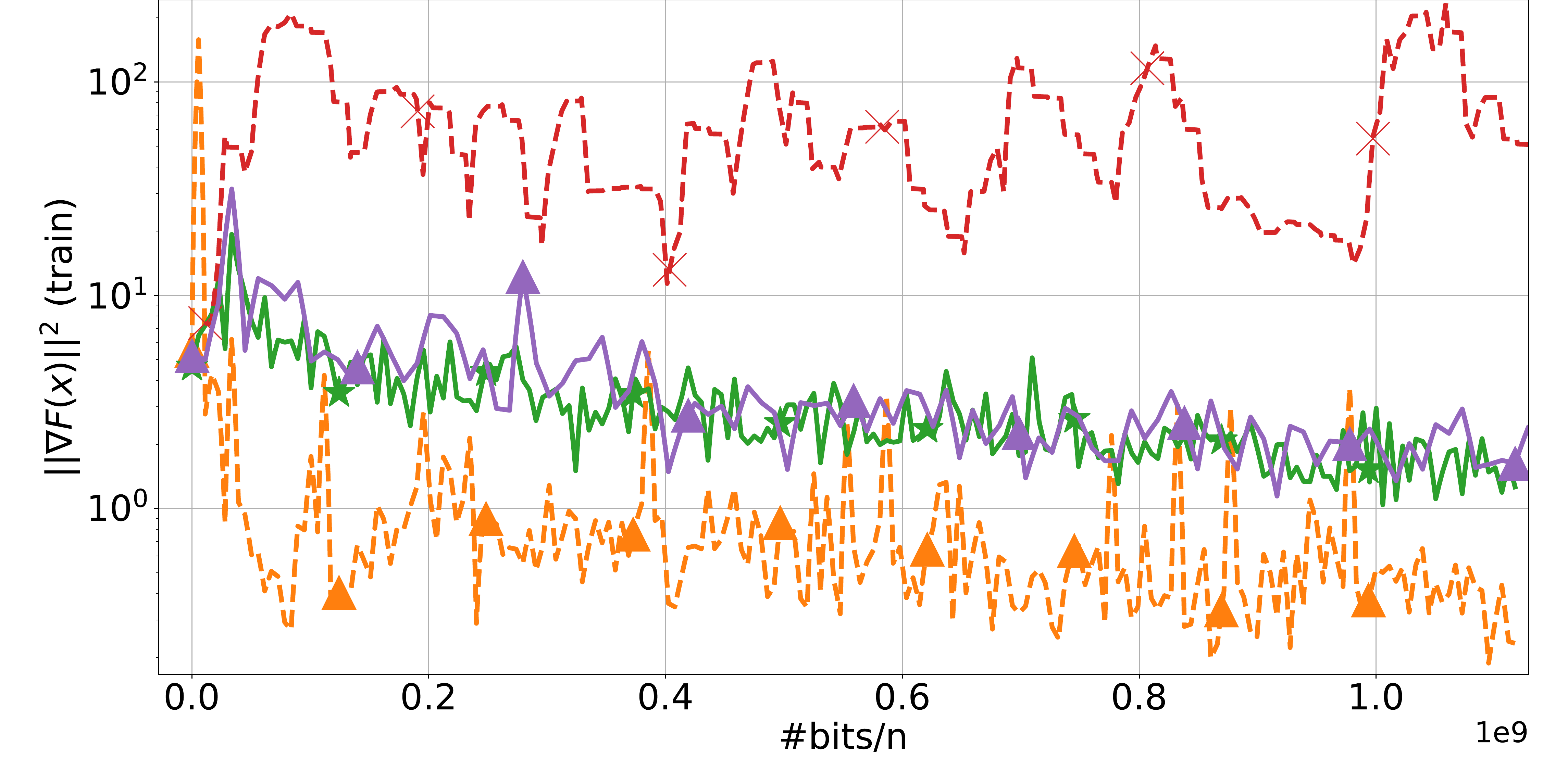}  \caption*{}
	\end{subfigure}
	\begin{subfigure}[ht]{0.49\textwidth}
		\includegraphics[width=\textwidth]{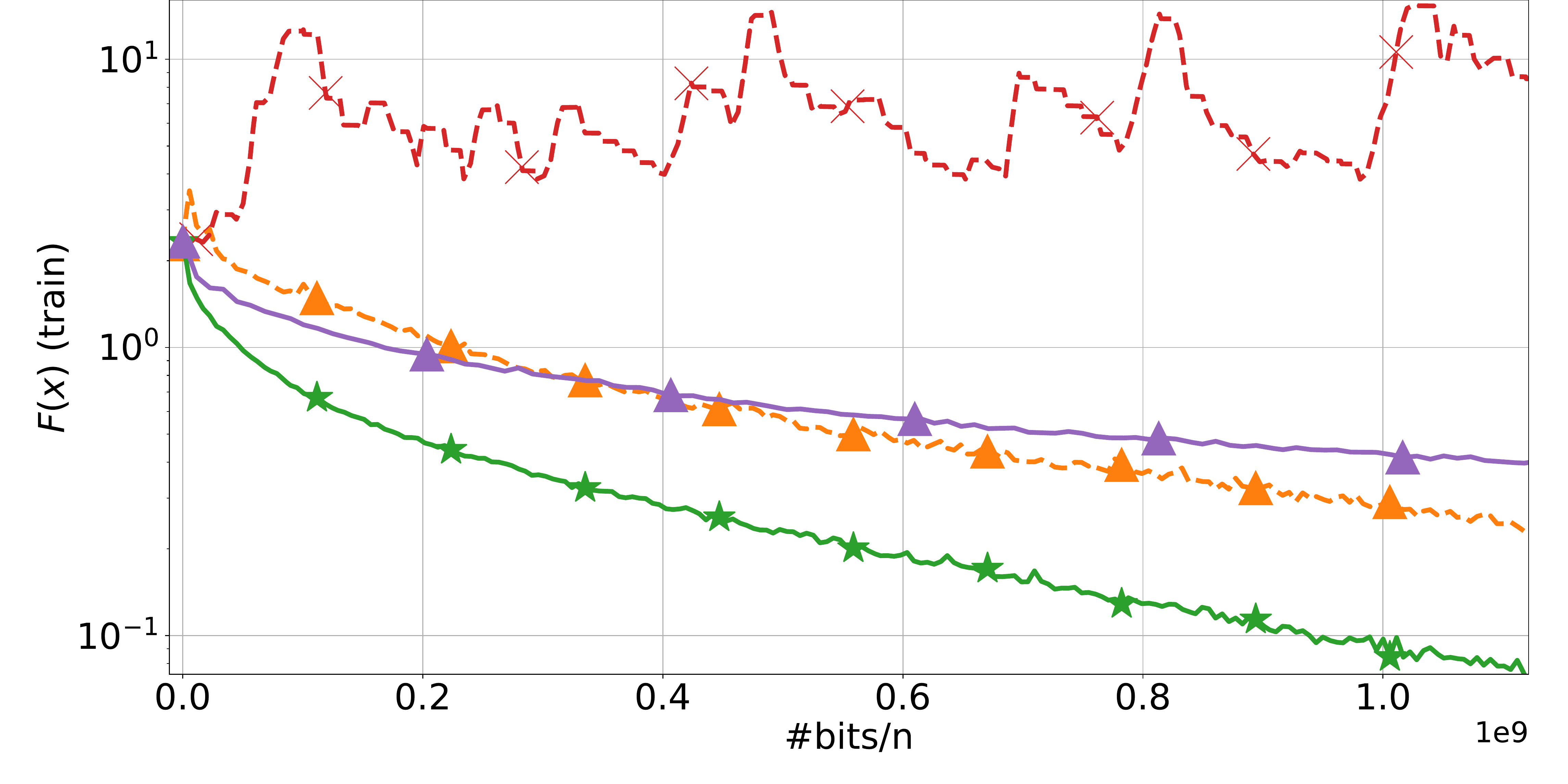}  \caption*{}
	\end{subfigure}
	
	\begin{subfigure}[ht]{0.49\textwidth}
		\includegraphics[width=\textwidth]{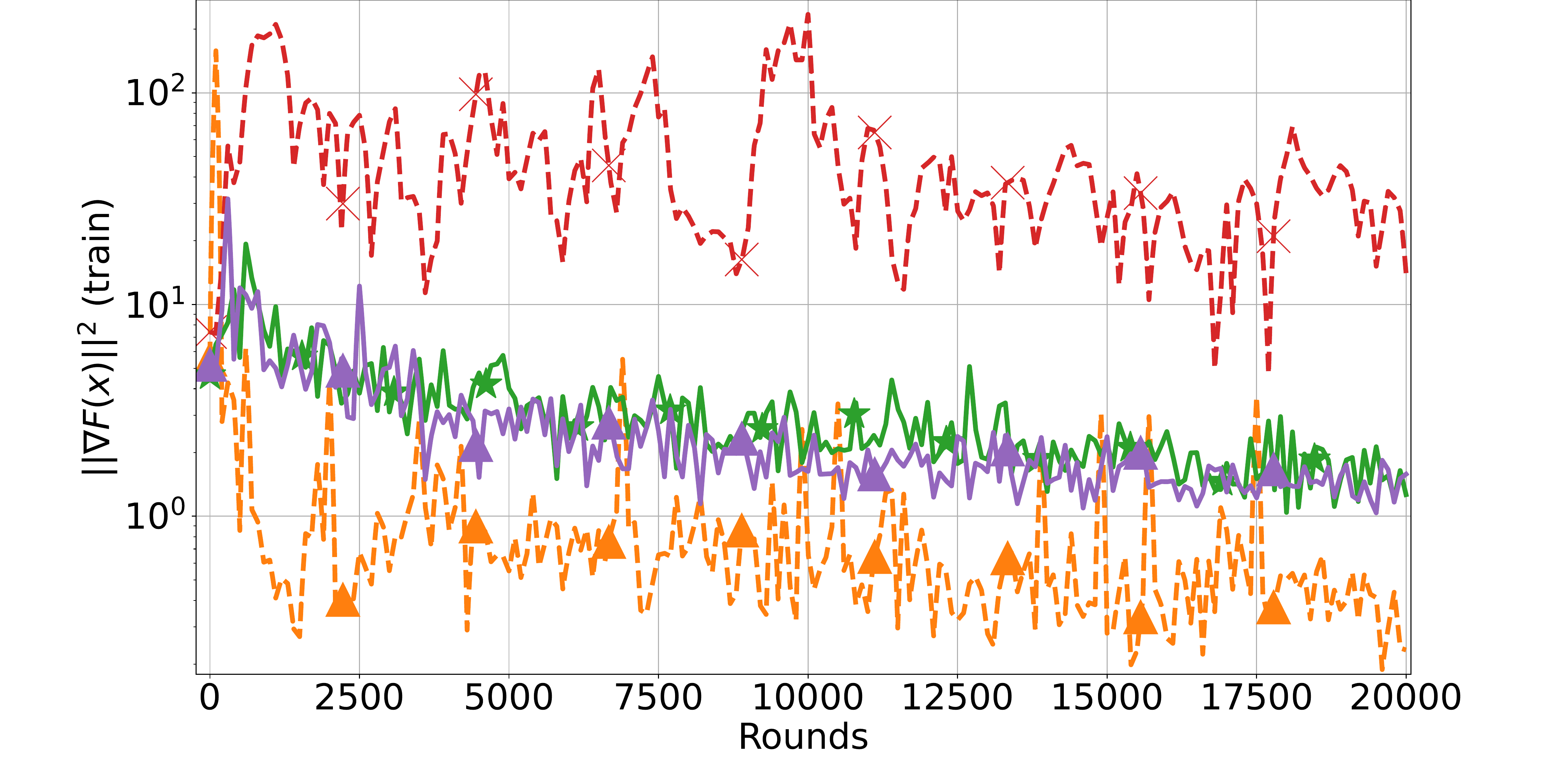}   \caption*{}
	\end{subfigure}
	\begin{subfigure}[ht]{0.49\textwidth}
		\includegraphics[width=\textwidth]{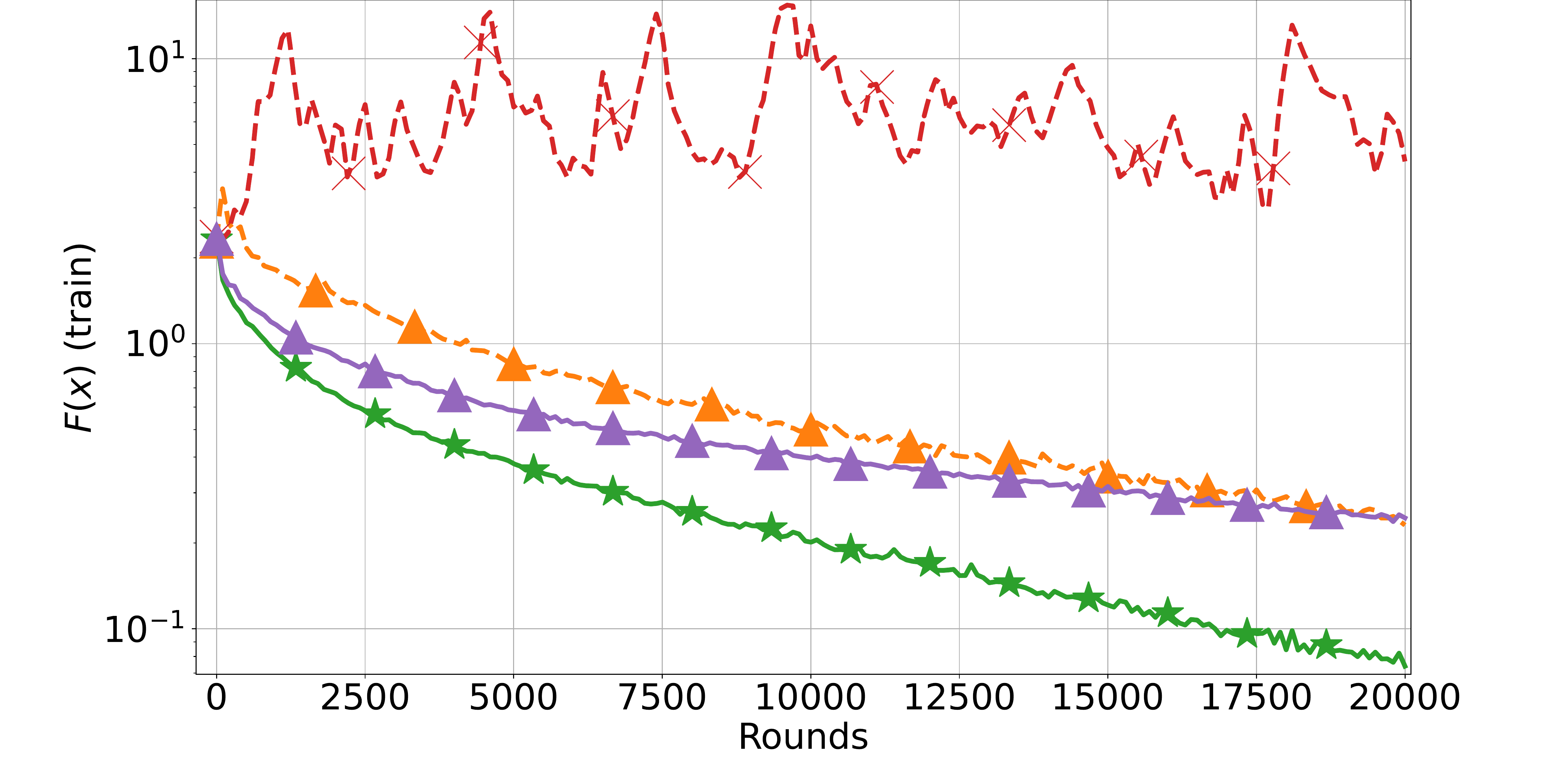}   \caption*{}
	\end{subfigure}
	
	\vspace{-20pt}
	\caption{\small{Comparison of optimization algorithms on training \texttt{ResNet-18} at \texttt{CIFAR10} dataset. Stepsizes for the methods were tuned from ${LR}_{set}$ (Eq. \ref{eq:lrset}). In all cases, we used the \texttt{RandK} sparsification operator. Total number of workers is $N=10$, number of workers per round is $S=1$, number of communication rounds is $20K$.}}
	
	\label{fig:resnet18_experiments_full}
\end{figure*}

\begin{figure*}[ht]
	\centering
	\captionsetup[sub]{font=scriptsize,labelfont={}}	
	\begin{subfigure}[ht]{1.0\textwidth}
		\includegraphics[width=\textwidth]{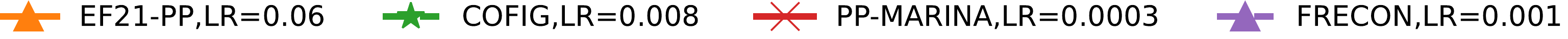}  \caption*{}
	\end{subfigure}
	
	\begin{subfigure}[ht]{0.49\textwidth}
		\includegraphics[width=\textwidth]{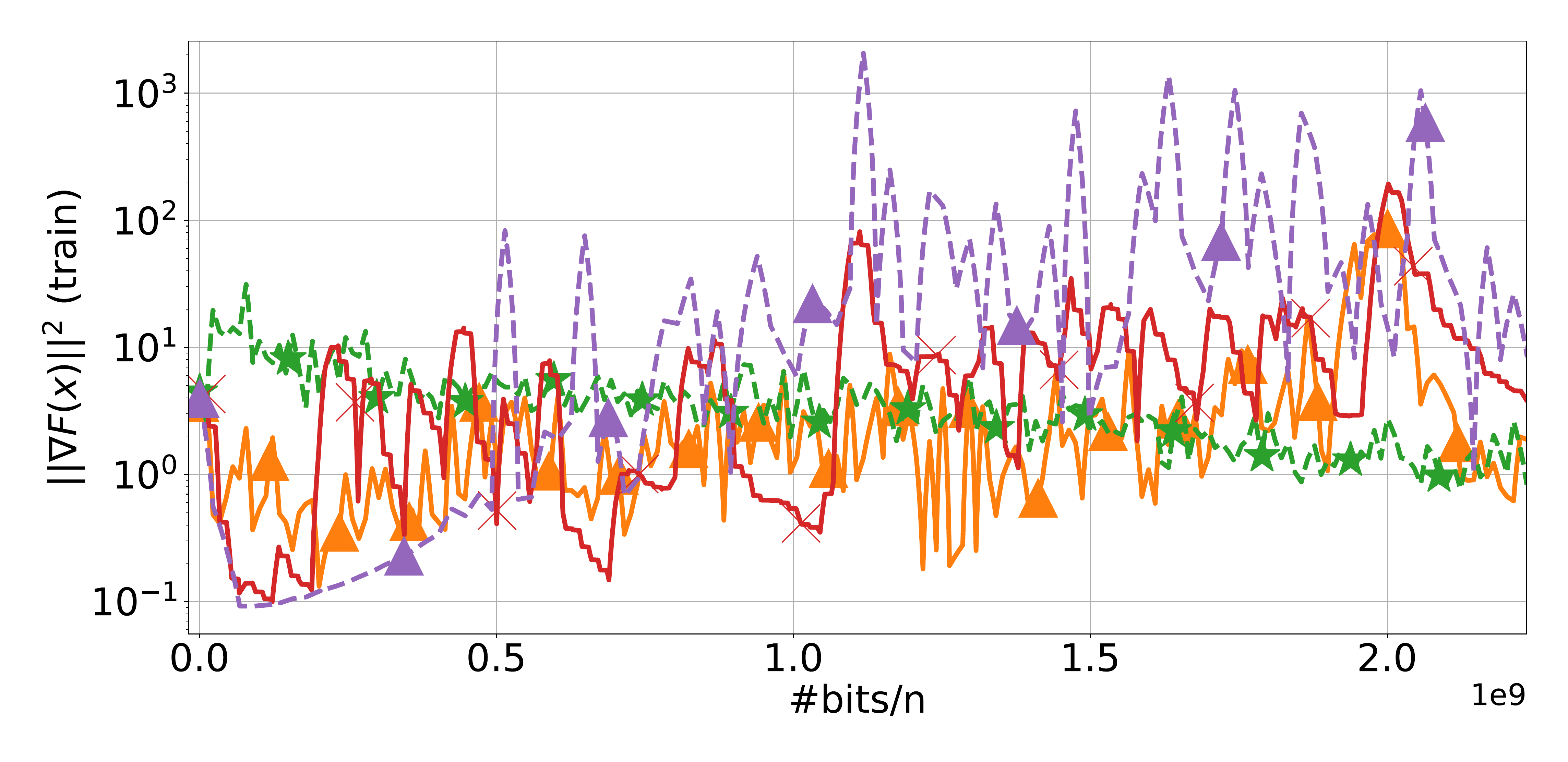}  \caption*{}
	\end{subfigure}
	\begin{subfigure}[ht]{0.49\textwidth}
		\includegraphics[width=\textwidth]{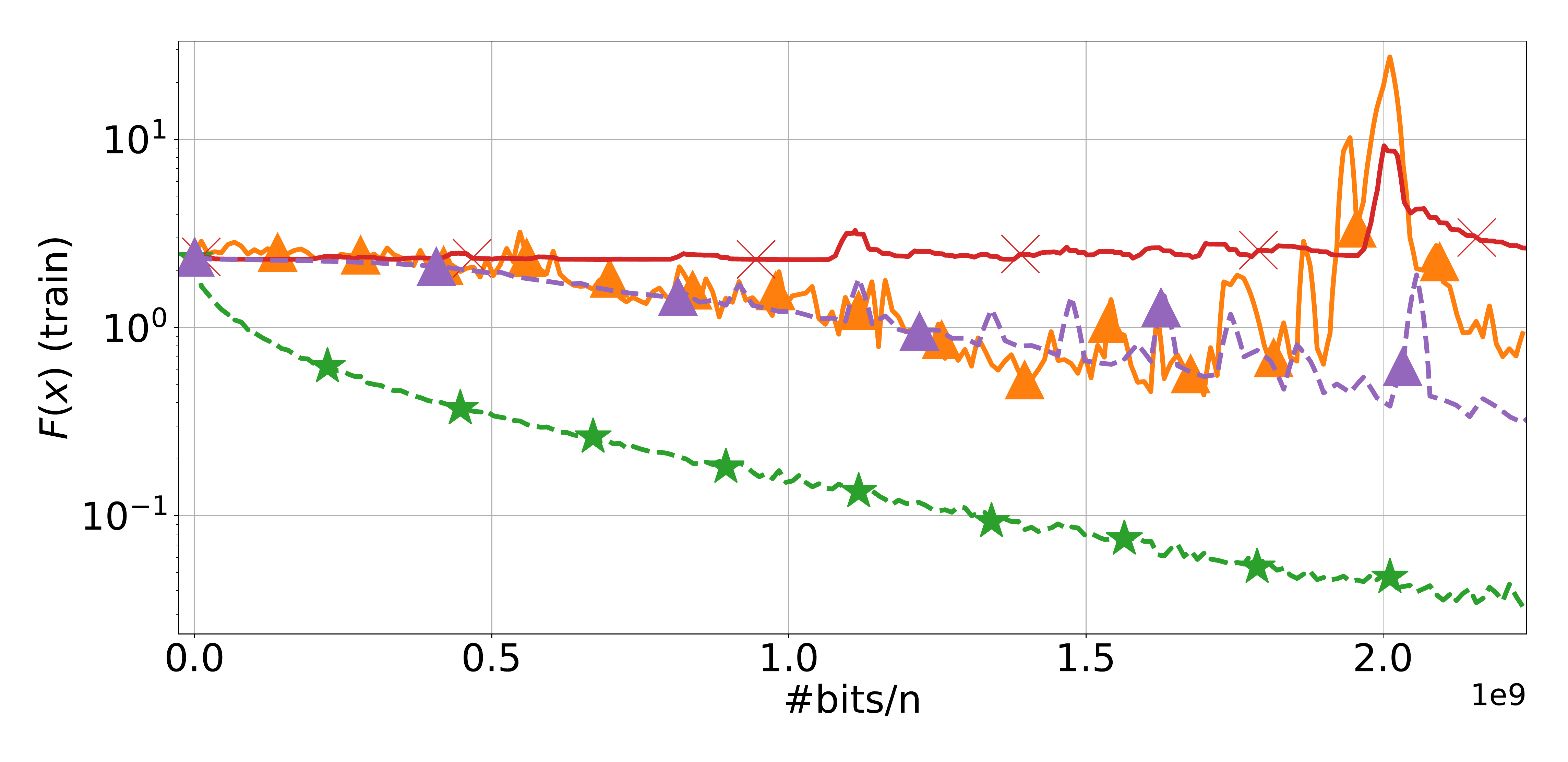}  \caption*{}
	\end{subfigure}
	
	\begin{subfigure}[ht]{0.49\textwidth}
		\includegraphics[width=\textwidth]{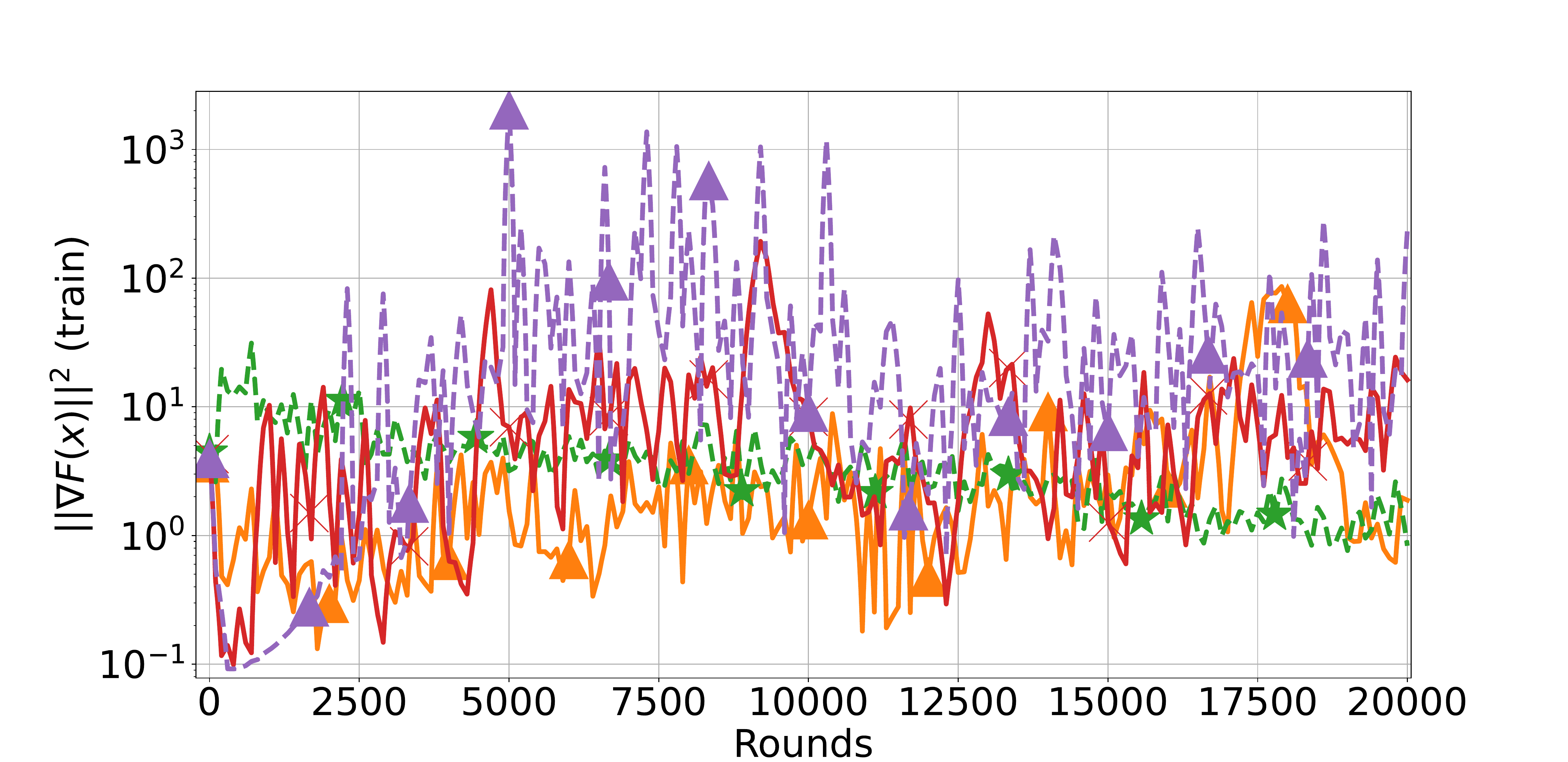}   \caption*{}
	\end{subfigure}
	\begin{subfigure}[ht]{0.49\textwidth}
		\includegraphics[width=\textwidth]{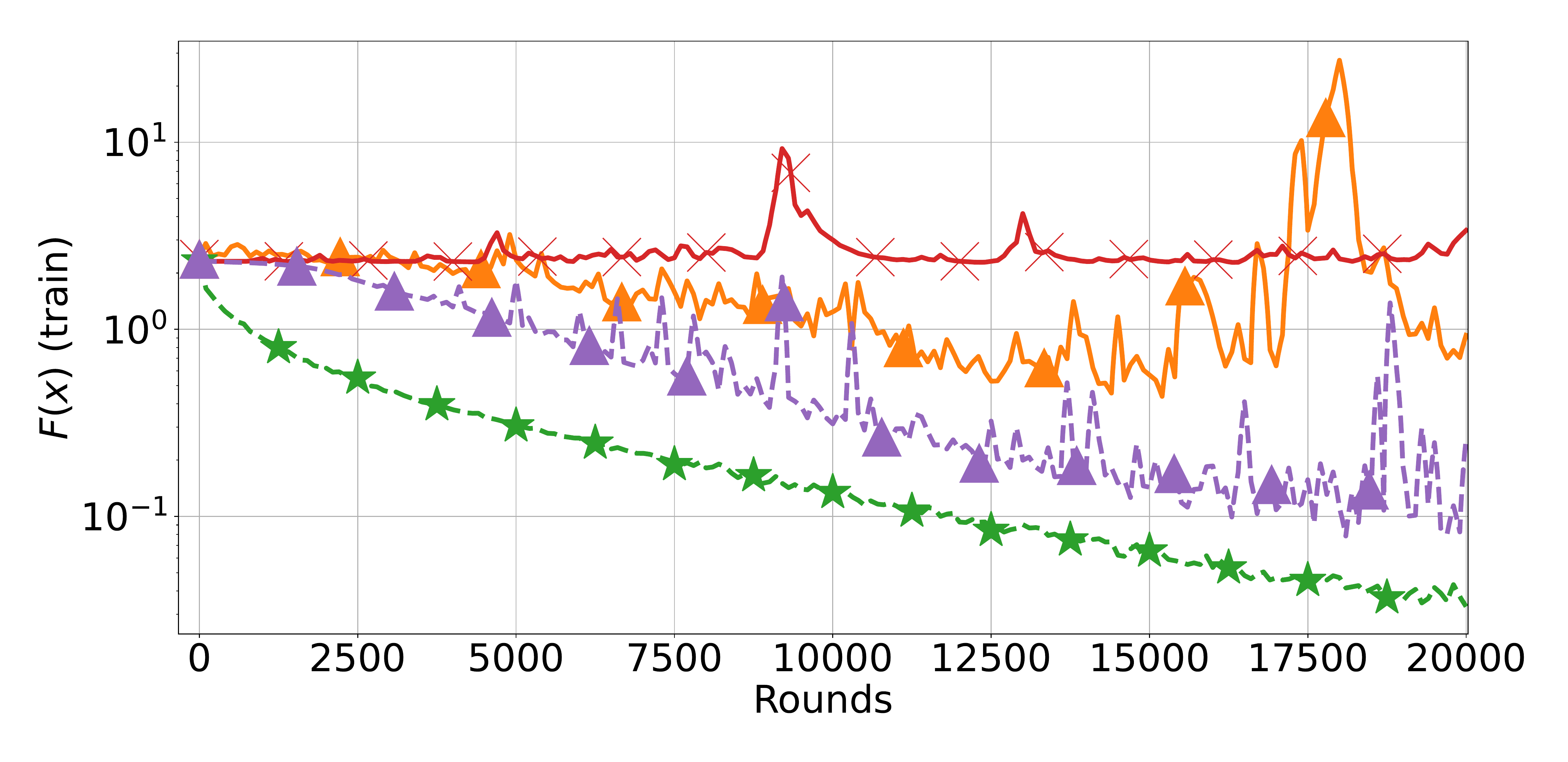}   \caption*{}
	\end{subfigure}
	
	\vspace{-20pt}
	\caption{\small{Comparison of optimization algorithms on training \texttt{ResNet-18} at \texttt{CIFAR10} dataset. Stepsizes for the methods were tuned from ${LR}_{set}$ (Eq. \ref{eq:lrset}). In all cases, we used the \texttt{RandK} sparsification operator. Total number of workers is $N=10$, number of workers per round is $S=2$, number of communication rounds is $20K$.}}
	
	\label{fig:resnet18_s2_experiments_full}
\end{figure*}

\begin{figure*}[ht]
	\centering
	\captionsetup[sub]{font=scriptsize,labelfont={}}	
	\begin{subfigure}[ht]{1.0\textwidth}
		\includegraphics[width=\textwidth]{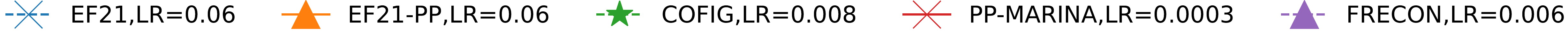}  \caption*{}
	\end{subfigure}
	
	\begin{subfigure}[ht]{0.49\textwidth}
		\includegraphics[width=\textwidth]{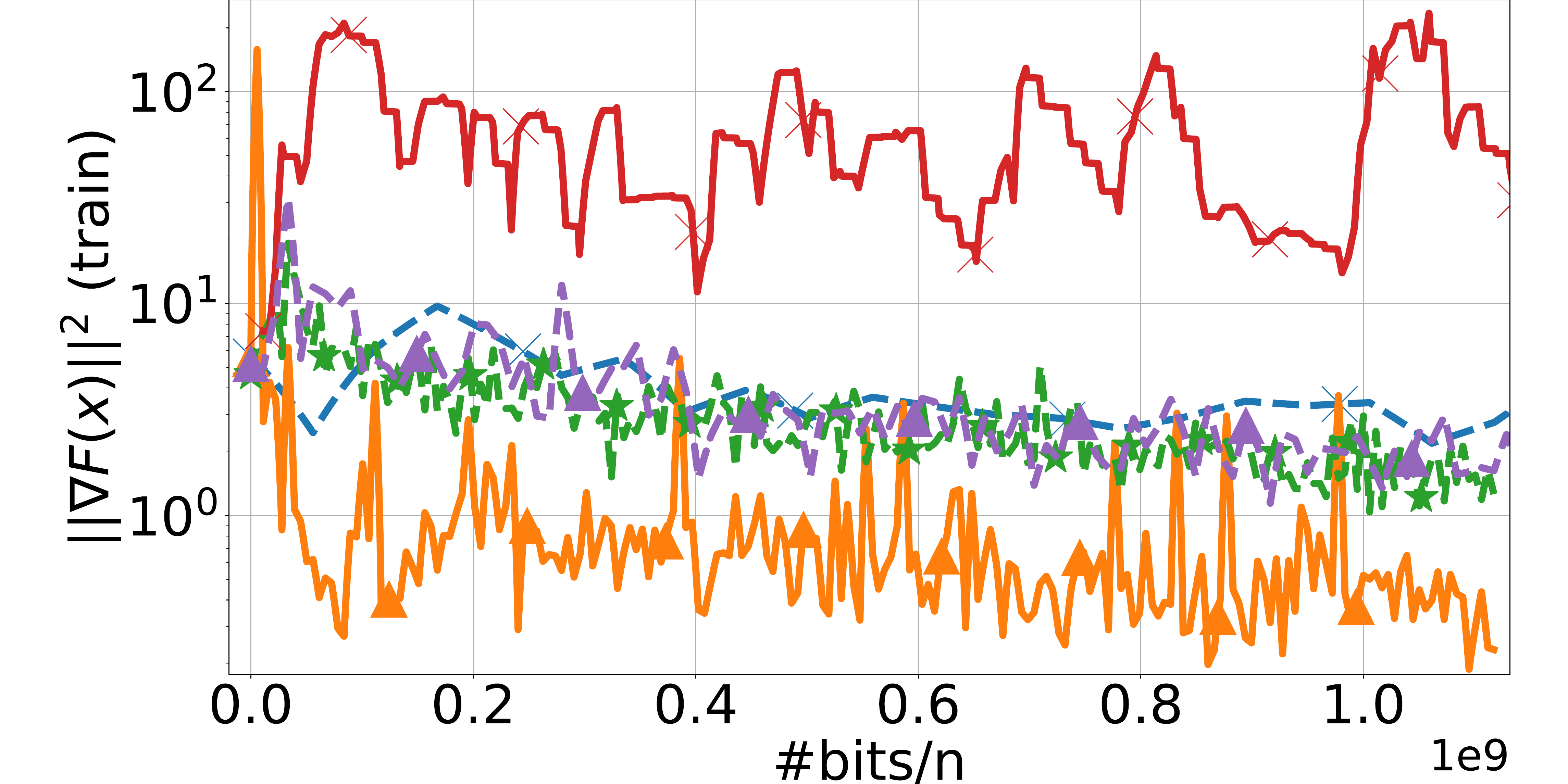}  \caption*{}
	\end{subfigure}
	\begin{subfigure}[ht]{0.49\textwidth}
		\includegraphics[width=\textwidth]{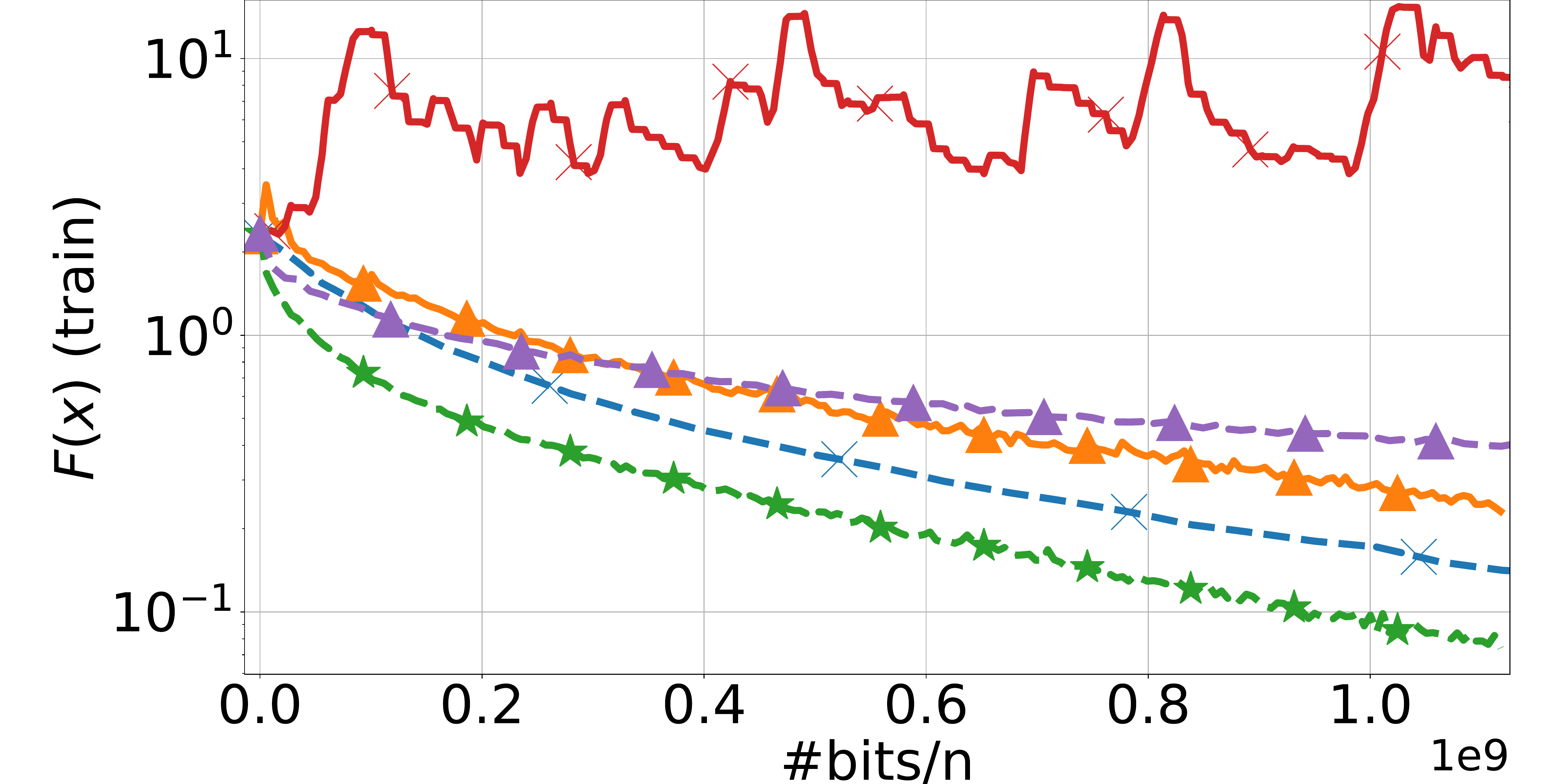}  \caption*{}
	\end{subfigure}
	
	\begin{subfigure}[ht]{0.49\textwidth}
		\includegraphics[width=\textwidth]{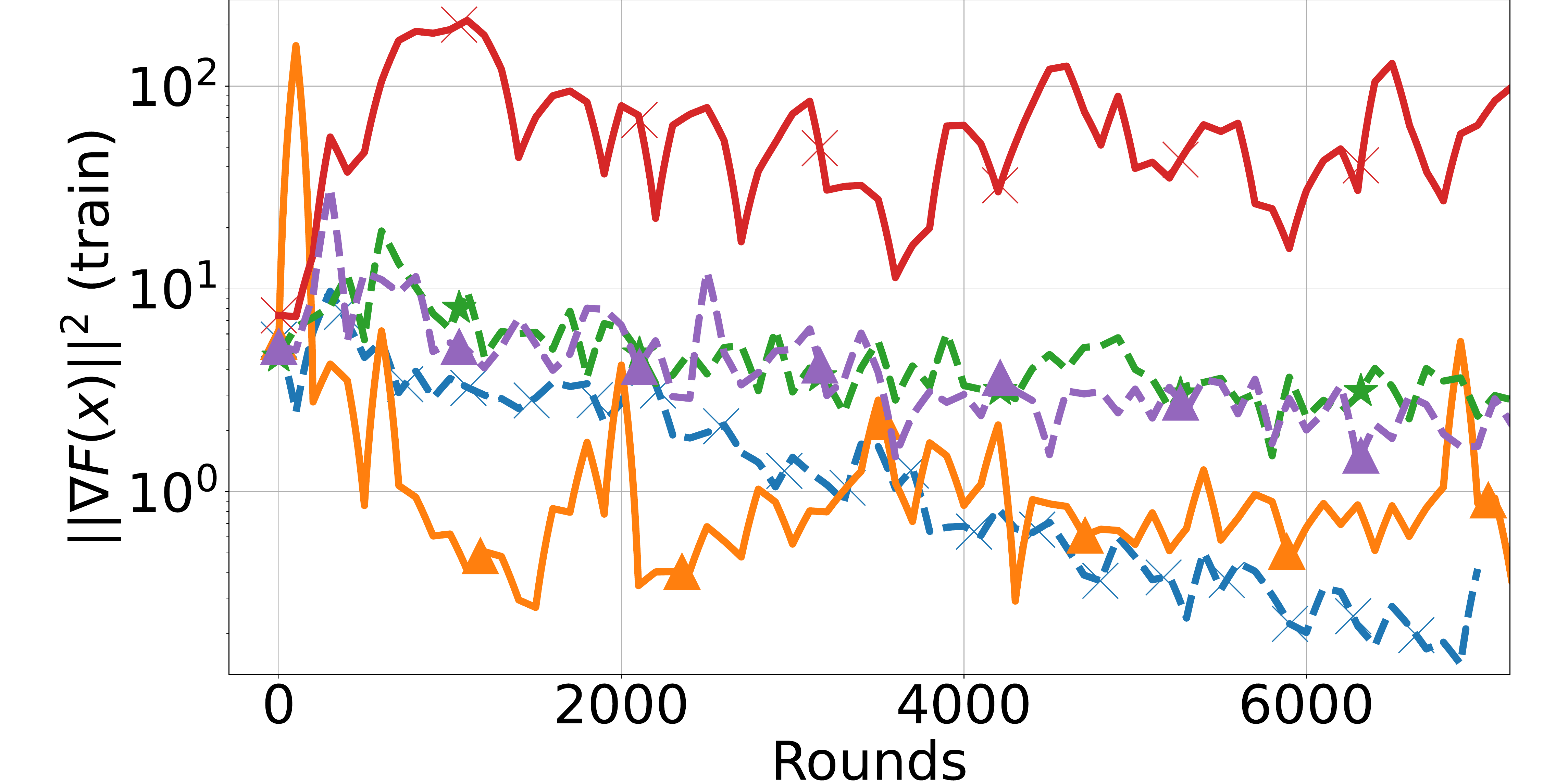}   \caption*{}
	\end{subfigure}
	\begin{subfigure}[ht]{0.49\textwidth}
		\includegraphics[width=\textwidth]{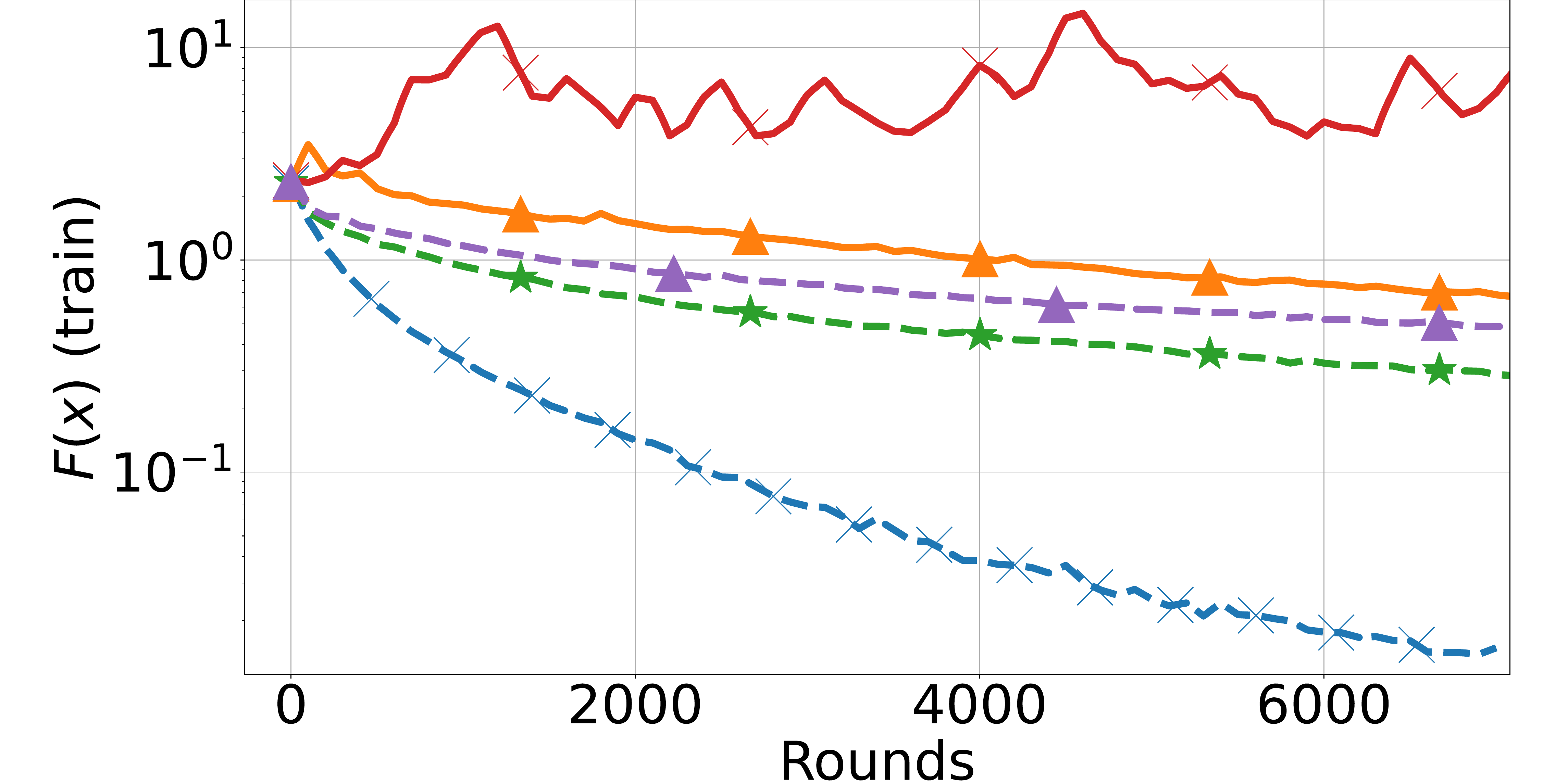}   \caption*{}
	\end{subfigure}
	
	\vspace{-20pt}
	\caption{\small{Comparison of optimization algorithms on training \texttt{ResNet-18} at \texttt{CIFAR10} dataset. Stepsizes for the methods were tuned from ${LR}_{set}$ (Eq. \ref{eq:lrset}). In all cases, we used the \texttt{RandK} sparsification operator. Total number of workers is $N=10$, number of workers per round is $S=1$ for all algorithms, except for \algname{EF21} where the number of workers per round is $S=N=10$.}}
	
	\label{fig:resnet18_experiments_full_fp}
\end{figure*}

In this experiment we train \texttt{ResNet-18}~\cite{he2016deep} on the \dataset{CIFAR10} dataset \cite{krizhevsky2009learning}. The dimension of optimization problem $d=11\,173\,962$, from which $11\,159\,232$ optimization variables come from training parameters of cross correlation operators \texttt{Conv2d}, $9\,600$ optimization variables comes from training \texttt{Batch Normalization} layers, and finally, $5\,130$ optimization variables come from training last linear \texttt{FC} layer.

The sizes of the training and validation set are $50K$ and $10K$ records. 
Similar to the heterogeneous binary classification experiments, the \dataset{CIFAR10} dataset has been partitioned heterogeneously across $10$ clients, where each client contains the data of a single class and there are 10 classes in total (i.e., \textbf{class-based sorting}).

\paragraph{Problem formulation}
We define $f_i(\vx)$ to be local empirical risk associated with the local data, $\gD_i$ stored on client $i$. The loss, $f_i(\vx)$ is a standard unweighted cross-entropy loss constructed with using output from neural network:
\begin{align}
	f(\vx) & \eqdef \dfrac{1}{N} \sum_{i=1}^{N} f_i(\vx) \\
	f_i(\vx) & = \dfrac{1}{N_i} \sum_{j=1}^{N_i} \text{CrossEntropyLoss}(b^{(j)}, g(a^{(j)},\vx)).
\end{align}

Here $a^{(j)} \in \mathbb{R}^{3 \times 28 \times 28}$ is input image, $b^{(j)} \in \{e_1, \dots, e_{10}\}$ is one one-hot encoding of class label, $N_i=|\gD_i|$, and $g$ is a neural networks parameterized by unknown variable $\vx$. Neural network $g(\cdot , \vx)$ obtains input image $a^{(j)}$ and produces vector $\mathbb{R}^{10}$ in probability simplex. 

\paragraph{Description of experiment} In this experiment, we random shuffle the \dataset{CIFAR10} dataset to avoid the problematic behavior of Batch Normalization in \texttt{ResNet-18}. The step sizes for different algorithms are tuned from the following set ${LR}_{set}$ to get fastest convergence:
\begin{align}
	\label{eq:lrset}
	{LR}_{set} = & \{0.1, 0.06, 0.03, 0.01, 0.006, 0.003, 0.001, 0.0006, \\
	& \quad 0.0003, 0.0001,  0.00006, 0.00003, 0.00001\}.\nonumber
\end{align}
In this experiment, we compare \algname{EF21}, \algname{EF21-PP}, \cofig, \frecon, and \algname{PP-MARINA}. We divide the dataset into $10$ parts ($N=10$ clients), and each client holds $5000$ samples. For \algname{EF21-PP}, \cofig, \frecon, and \algname{PP-MARINA}, we test both cases for $S=1$ and $S=2$. We use the \texttt{RandK}~\cite{wangni2018gradient} compressor with $K=0.05D$ for all algorithms, where $D$ is the number of parameters for \texttt{ResNet-18}. Unlike the previous binary classification experiments, in this deep learning experiments, different algorithms cannot access the local full gradients from the clients ($\sigma > 0$). When the algorithms need the local full gradients, each client estimates the gradient using a minibatch with size $b=500$ ($10\%$ of the local data).

\paragraph{Experiment results} {The results for this image classification deep learning experiments are presented in \Cref{fig:resnet18_experiments_full} and \Cref{fig:resnet18_s2_experiments_full} with $S=1$ and $S=2$, respectively.} In our experiments, we observe that \algname{PP-MARINA} is practically impossible in those circumstances. Besides, the performance of \frecon is similar to \algname{EF21-PP}, and our \cofig converges fastest among all these algorithms. For \algname{FRECON} we have tuned parameter $\lambda$ from trying several values in $[0,1]$ we have found that for training this NN the value $\lambda=0.75$ leads to more fast convergence. In \Cref{fig:resnet18_experiments_full_fp} there is a comparison with full participated version \algname{EF21}, and our \cofig converges faster as well in terms of the communication bits.

We attribute the results to the fact that \cofig works better with local variance (\Cref{ass:bounded-variance}) compared with \frecon. {It is also worth noting that based on our hyperparameter search, we observe that for \frecon, $S=1$ (\Cref{fig:resnet18_experiments_full}) can use a larger learning rate than $S=2$ (\Cref{fig:resnet18_s2_experiments_full}), which reflect that \frecon may not work very well or unstable when the local variance $\sigma^2$ (\Cref{ass:bounded-variance}) is somewhat large.} Besides, \algname{EF21-PP}, \frecon, and \cofig all perform better than the full participation algorithm \algname{EF21} in terms of the communicate bits (\Cref{fig:resnet18_experiments_full_fp}), which shows that partial participation may indeed speed up the convergence in terms of the total communication bits.

This experiment shows that when the number of local data is large, and we can not compute the local full gradient, \cofig may behave best. 
Together with the previous experiments, \cofig can be set up as a default optimization algorithm for federated learning applications, and then divert to \frecon if we have further information, e.g., we can compute the local full gradient and/or the number of clients is very large.

\subsection{Remarks for experiments} 
According to both logistic regression (with nonconvex regularizer) and the image classification deep learning experiments, we validate the effectiveness of our proposed algorithms \cofig and \frecon. However, the algorithms still have their own advantages in different settings. In short, we summarize the following messages:
\begin{enumerate}
	\item For problems where we can estimate the local gradients fairly accurately (i.e., the local variance $\sigma^2$ is small), \frecon performs better than \cofig in terms of the communication bits. This trend is also corroborated by the faster theoretical convergence rate of \frecon when $\sigma^2=0$ (Theorem~\ref{thm:frecon-nonconvex}).
	\item When optimizing large-scale neural networks like ResNet that has a relatively large local variance , \cofig performs better than \frecon in terms of the communication bits. This trend is supported by the convergence results of \cofig and \frecon (Theorem \ref{thm:cofig-nonconvex} and \ref{thm:frecon-nonconvex}), where \cofig has a smaller dependency with respect to $\sigma^2$ than \frecon. The large dependency on $\sigma^2$ may incur some unstable behaviors when training large neural networks.
\end{enumerate}

\section{Conclusion}
In this paper, we introduce \cofig and \frecon, which successfully incorporate communication compression with client-variance reduction for dealing with large number of heterogeneous clients in federated learning. 
Our \frecon obtains the best communication complexity for federated nonconvex setting, and our \cofig gives the first theoretical convergence result for federated non-strongly convex setting that supports both communication compression and partial participation, while previous works either obtain worse convergence results or require full clients communication.

\bibliographystyle{siamplain}
\bibliography{ref}

\end{document}